\pdfoutput=1

\documentclass[11pt]{article}

\usepackage[final]{acl}
\usepackage{booktabs}
\usepackage{times}
\usepackage{latexsym}

\usepackage[T1]{fontenc}

\usepackage[utf8]{inputenc}

\usepackage{microtype}

\usepackage{inconsolata}
\usepackage{algorithm}
\usepackage{algorithmic}

\usepackage{graphicx}
\graphicspath{{images/}}
\usepackage{amsmath} 
\usepackage{subfigure}
\usepackage{makecell}
\usepackage{amsfonts} 
\usepackage{pifont}
\usepackage{booktabs}
\usepackage{enumitem}
\usepackage{multirow}
\usepackage{CJKutf8}

\definecolor{deepred}{rgb}{0.698,0.133,0.133}
\definecolor{blue}{rgb}{0,0,1}

\usepackage{tikz}
\usepackage[edges]{forest}
\usepackage{pgfplots}
\usepackage[english]{babel}
\usepackage[framemethod=tikz]{mdframed}
\usepackage{subcaption}
\definecolor{lgreen}{rgb}{0.89,0.94,0.85}
\definecolor{lred}{rgb}{0.98, 0.90, 0.84}
\definecolor{lyellow}{rgb}{1.00, 0.95, 0.80}
\definecolor{lblue}{rgb}{0.85, 0.89, 0.95}
\definecolor{hidden-draw}{RGB}{20,68,106}
\definecolor{hidden-pink}{RGB}{255,245,247}
\definecolor{hidden-blue}{RGB}{240, 248, 255}

\title{From Specific-MLLMs to Omni-MLLMs: \\ A Survey on MLLMs Aligned with Multi-modalities}

\author{Shixin Jiang$^{1}$, Jiafeng Liang$^{1}$, Jiyuan Wang$^{1}$, Xuan Dong$^{1}$, Heng Chang$^{2}$\\ \textbf{Weijiang Yu$^{2}$, Jinhua Du$^{2}$, Ming Liu$^{1,3}$\thanks{Corresponding author}, Bing Qin$^{1,3}$} \\
        $^{1}$Harbin Institute of Technology, Harbin, China\\ 
        $^{2}$Huawei Inc., Shenzhen, China\\
        \textsuperscript{3}Peng Cheng Laboratory, Shenzhen, China\\
   \texttt{\{sxjiang, jfliang, jywang, mliu, qinb\}@ir.hit.edu.cn }
 }

\begin{document}
\maketitle
\begin{abstract}
To tackle complex tasks in real-world scenarios, more researchers are focusing on Omni-MLLMs, which aim to achieve omni-modal understanding and generation. 
Beyond the constraints of any specific non-linguistic modality, Omni-MLLMs map various non-linguistic modalities into the embedding space of LLMs and enable the interaction and understanding of arbitrary combinations of modalities within a single model. 
In this paper, we systematically investigate relevant research and provide a comprehensive survey of Omni-MLLMs. Specifically, we first explain the four core components of Omni-MLLMs for unified multi-modal modeling with a meticulous taxonomy that offers novel perspectives. Then, we introduce the effective integration achieved through two-stage training and discuss the corresponding datasets as well as evaluation. Furthermore, we summarize the main challenges of current Omni-MLLMs and outline future directions. We hope this paper serves as an introduction for beginners and promotes the advancement of related research. 
Resources have been made publicly available at \href{https://github.com/threegold116/Awesome-Omni-MLLMs}{https://github.com/threegold116/Awesome-Omni-MLLMs}.
\end{abstract}
\section{Introduction}

The remarkable performance of continuously evolving Multi-modal Large Language Models (MLLMs) has pointed to a possible direction for achieving general artificial intelligence~\citep{spark_gpt4,gpt4}. MLLMs extend Large Language Models (LLMs) by integrating them with pre-trained models tailored to specific modalities, such as Vision-MLLMs~\citep{llava,qwen2vl,emu}, Audio-MLLMs~\citep{speechgpt,qwenaudio}, and 3D-MLLMs~\citep{pointllm}. However, these modality-specific MLLMs (Specific-MLLMs) are insufficient to tackle complex tasks in real-world scenarios that simultaneously involve multiple modalities. Therefore, efforts are being made to expand the range of modalities for understanding and generation, giving rise to the omni-modality MLLMs (Omni-MLLMs).\looseness=-1
\begin{figure}[t]
    \centering
    \includegraphics[width=1\linewidth]{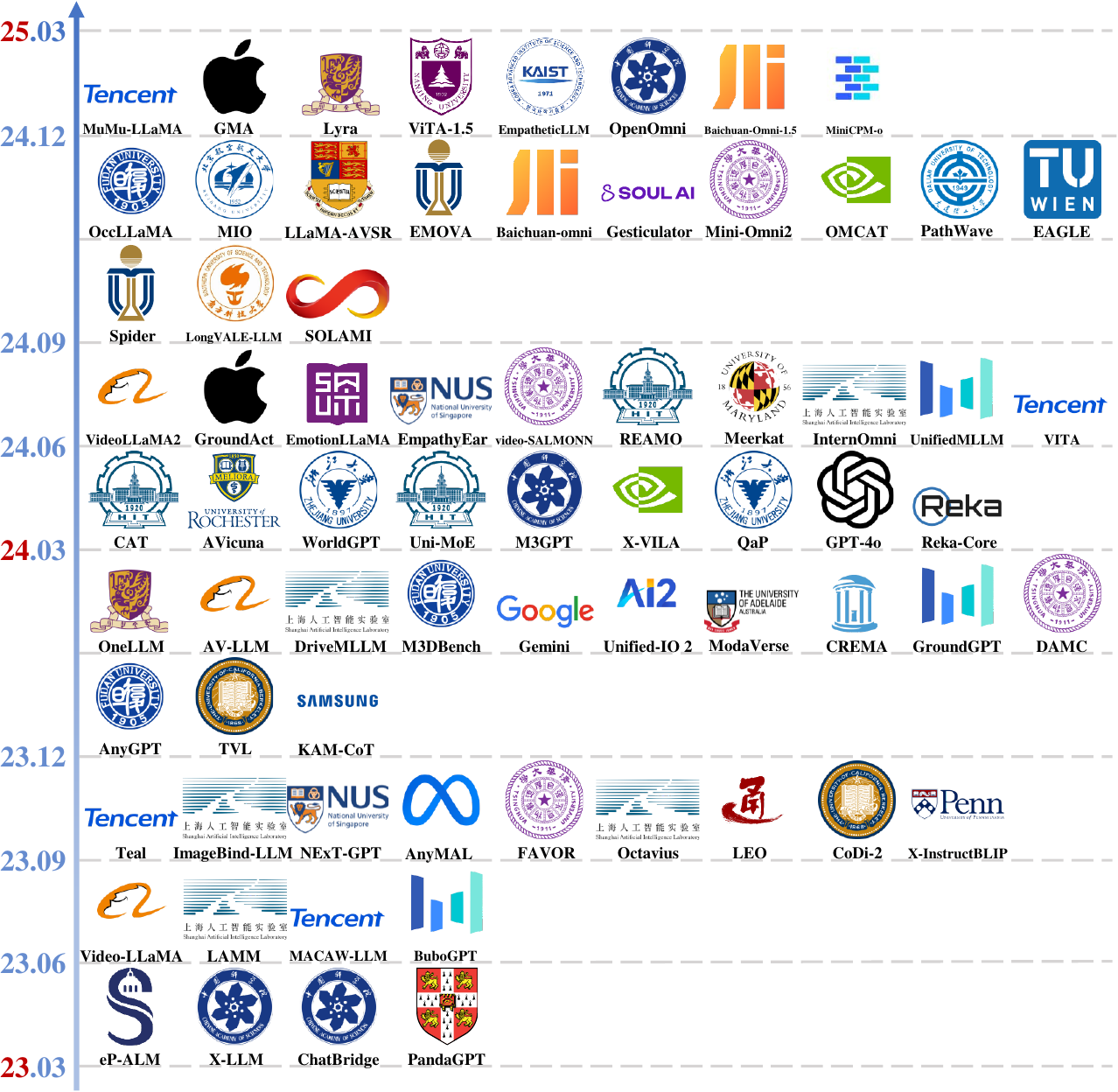}
    \caption{The timeline of representative Omni-MLLMs.}
    \label{fig:time_line}
\end{figure}

By integrating multiple pre-trained models of more non-linguistic modalities~\citep{clip,Whisper,ulip2,stablediffusion,audioldm}, Omni-MLLMs expand the modalities for understanding and even generation based on Specific-MLLMs. Omni-MLLMs leverage the emergent capabilities of LLMs to treat various non-linguistic modalities as different \textit{foreign languages}, enabling the interaction and understanding of information across different modalities within a unified space~\citep{xllm2023chen,xinstruclblip}. Compared to Specific-MLLMs, Omni-MLLMs can perform multiple uni-modal\footnote{ ``Uni'' and  ``Cross'' refer to the number of non-linguistic modalities involved in the interaction, in contrast to “multimodal reasoning,” traditionally reserved for vision-language tasks~\cite{xinstruclblip}.} understanding and generation tasks, as well as cross-modal tasks across two or more non-linguistic modalities, allowing a single model to handle arbitrary combinations of modalities.


A review of the development of Omni-MLLMs reveals that it has been continuously expanding in three directions. On the one hand, the types of modalities processed by Omni-MLLM have been continuously increasing, from X-LLMs that handle vision and audio to X-InstructBLIP~\citep{xinstruclblip} which adds 3D modality capabilities, PandaGPT~\citep{pandagpt} that incorporates IMU modality, and finally One-LLM~\citep{onellm}, which processes eight different modalities simultaneously. On the other hand, the ability to interact across modalities of Omni-MLLMs has also expanded, from the joint 3D-Image and Audio-Image cross-modal reasoning capability in ImageBind-LLM~\citep{imagebindllm} to the cross-modal generation capability of CoDi-2 that leverages interleaved audio and image contexts to generate both audio and images~\citep{codi2}. The Omni-MLLM is thus trending towards an ``Any-to-Any'' model. Besides, the application scenarios of Omni-MLLMs have been broadened, encompassing real-time multimodal speech interaction like Mini-Omni2 and Lyra~\citep{Mini-Omni2,Lyra}, world simulation like WordGPT~\citep{worldgpt}, multi-sensor autonomous driving like DriveMLM~\citep{drivemlm}, etc. In addition to the open-source models, there are also some closed-source Omni-MLLMs such as GPT-4o~\citep{gpt4o}, Gemini~\citep{Gemini1.5}, and Reka~\citep{reka}. The timeline of Omni-MLLMs is shown in Figure~\ref{fig:time_line}. Despite the emergence of numerous Omni-MLLMs, there is still a lack of systematic evaluation and analysis.



To fill the gap, we propose this work to conduct a comprehensive and detailed analysis of Omni-MLLMs. We first review the architecture of Omni-MLLMs in four parts~(\S\ref{section:omni_mllm}). Next, we summarize how Omni-MLLMs expand across multiple modalities through the two-stage training process~(\S\ref{section:omni_mllm_training}); then present the training data construction and performance evaluation~(\S\ref{section:omni_mllm_resource}). Furthermore, we highlight some key challenges and future directions~(\S\ref{section:omni_mllm_future}). Finally, we provide a brief summary~(\S\ref{section:omni_mllm_conclusion}) and discuss related surveys in the Appendix~\ref{section:appendix_related}.

Our contributions can be summarized as follows:
(1) \textbf{\textit{Comprehensive Survey}}: This is the first comprehensive survey dedicated for Omni-MLLMs;
(2) \textbf{\textit{Meticulous taxonomy}}: We introduce a meticulous taxonomy (shown in Figure~\ref{fig:taxonomy});
(3) \textbf{\textit{Challenges and Future}}: We outline the challenges of Omni-MLLMs and shed light on future research.

\section{Omni-MLLM Architecture}\label{section:omni_mllm}

\tikzset{%
    parent/.style =          {align=center,text width=0.7cm, rounded corners=2pt, line width=0.8mm, fill=white!0, draw=white!90},
    child/.style =           {align=center,text width=1.4cm,rounded corners=2pt, fill=blue!10,draw=blue!90,line width=0.3mm},
}

\tikzstyle{my-box}=[
    rectangle,
    draw=hidden-draw,
    rounded corners,
    text opacity=1,
    minimum height=1.5em,
    minimum width=5em,
    inner sep=2pt,
    align=center,
    fill opacity=.5,
    line width=0.8pt,
]
\tikzstyle{leaf}=[my-box, minimum height=1.5em,
    fill=hidden-blue!100, text=black, align=left,font=\normalsize,
    inner xsep=2pt,
    inner ysep=4pt,
    line width=0.8pt,
]
\begin{figure*}[t!]
    \centering
    \resizebox{1.0\textwidth}{!}{
        \begin{forest}
            forked edges,
            for tree={
                grow'=0,
                draw,
                reversed=true,
                anchor=base west,
                parent anchor=east,
                child anchor=west,
                base=left,
                font=\large,
                rectangle,
                rounded corners,
                align=left,
                minimum width=4em,
                edge+={darkgray, line width=1pt},
                s sep=3pt,
                inner xsep=2pt,
                inner ysep=3pt,
                line width=0.8pt,
                ver/.style={rotate=90, child anchor=north, parent anchor=south, anchor=center},
            },
            where level=1{text width=5.5em,font=\normalsize,}{},
            where level=2{text width=5.5em,font=\normalsize,}{},
            where level=3{text width=5.5em,font=\normalsize,}{},
            where level=4{text width=5em,font=\normalsize,}{},
			[
			    Omni-MLLMs, ver
			    [
			      Continuous\\Encoding 
                [
                Multi-branch\\Projection
                    [
                    \textcolor{deepred}{\textbf{Cross-modal Understanding:}} eP-ALM~\citep{ePalm}{,}X-LLM~\citep{xllm2023chen}{, }ChatBridge\\~\citep{chatbridge}{, }Video-LLaMA~\citep{videollama}{, }LAMM~\citep{LAMM}{, }Macaw-LLM\\~\citep{macawllm}{, }BuboGPT~\citep{bubogpt}{, }AnyMAL~\citep{anymal}{, }FAVOR~\citep{favor}{, }\\Octavius~\citep{Octavius}{, }X-InstructBLIP~\citep{xinstruclblip}{, }AV-LLM~\citep{avllm}{, }\\DriveMLM~\citep{drivemlm}{, }Omni-3D~\citep{Omni-3D}{, }MultiPLY~\citep{MultiPLY}{, }GroundingGPT\\~\citep{groundinggpt}{, }DAMC~\citep{ModelComposition}{, }CAT~\citep{cat}{, }AVicuna~\citep{avicuna}{, }\\QaP~\citep{qap}{, }Uni-Moe~\citep{unimoe}{, }Video-LLaMA2~\citep{videollama2}{, }Emotion-LLaMA\\~\citep{emotionllama}{, }video-SALMONN~\citep{video-SALMONN}{, }InternOmni~\citep{interomni}{, }EAGLE\\~\citep{eagel}{, }Meerkat~\citep{meerkat}{, }Dolphin~\citep{Dolphin}{, }Med-2E3\\~\citep{Med-2E3}{, }Llama-AVSR~\citep{Llama-AVSR}{, }Baichuan-Omni~\citep{baichuan-omni}{, }OMCAT\\~\citep{omcat}{, }LongVALE-LLM~\citep{LongVALE-LLM}{, }Baichuan-Omni1.5~\citep{baichuanomni1.5}{, }\\OpenOmni~\citep{openomni}
                    \\ \textcolor{deepred}{\textbf{Cross-modal Understanding \& Generation:}} NExT-GPT~\citep{nextgpt}{, }Unified-IO2~\citep{unifiedio2}{, }\\ModaVerse~\citep{modaverse}{, }X-VILA~\citep{xvila}{, }REAMO~\citep{REAMO}{, }UnifiedMLLM\\~\citep{unifiedmllm}{, }VITA~\citep{vita}{, }Mini-Omni2~\citep{Mini-Omni2}{, }CAD-MLLM~\citep{Omni-CAD}{, }\\MuMu-LLaMA~\citep{mumullama}{, }Lyra~\citep{Lyra}{, }VITA-1.5~\citep{vita-1.5}{, }\\EmpatheticLLM~\citep{EmpatheticLLM}{, }MiniCPM-o~\citep{minicpm-o-2.6}
                        , leaf, text width=46em
                    ]
                ]
                [
              Uni-branch\\Projection
                    [  \textcolor{deepred}{\textbf{Cross-modal Understanding:}} PandaGPT~\citep{pandagpt}{, }ImageBind-LLM~\citep{imagebindllm}{, }One-LLM\\~\citep{onellm}{, }CREMA~\citep{CREMA}{, }TVL~\citep{TVL}{, }PathWeave~\citep{PathWeave}
                    \\ \textcolor{deepred}{\textbf{Cross-modal Understanding \& Generation:}} CoDi-2~\citep{codi2}{, }WorldGPT~\citep{worldgpt}{, }\\EmpathyEar~\citep{EmpathyEar}{, }EGMI~\citep{egmi}
                        , leaf, text width=46em
                    ]
                ]
			    ]
			    [
		           Discrete\\Encoding
			        [Embedding 
                    [\textcolor{deepred}{\textbf{Cross-modal Understanding \& Generation:}} Teal~\citep{teal}{, }AnyGPT~\citep{anygpt}{, }M3GPT\\~\citep{M3GPT}{, }OccLLaMA~\citep{occllama}{, }MIO~\citep{MIO}{, }Gesticulator~\citep{Gesticulator}{, }\\SOLAMI~\citep{solami}
			              , leaf, text width=46em
			            ]]
			    ]
                [
		           Hybrid\\Encoding
			         [
                        Projection\\\&Embedding
                     [ \textcolor{deepred}
                     {\textbf{Cross-modal Understanding:}} SynesLM~\citep{SynesLM}
                     \\\textcolor{deepred}{\textbf{Cross-modal Understanding \& Generation:}} LEO~\citep{LEO}{, }Ground-Action~\citep{groundaction}{, }\\EMOVA~\citep{emova}{, }GMA~\citep{GMA}, leaf, text width=46em
			            ]]
			    ]
			]
            \end{forest}
            
    }
    \caption{Taxonomy for Omni-MLLMs based on their encoding and alignment methods.}
    \label{fig:taxonomy}
\end{figure*}
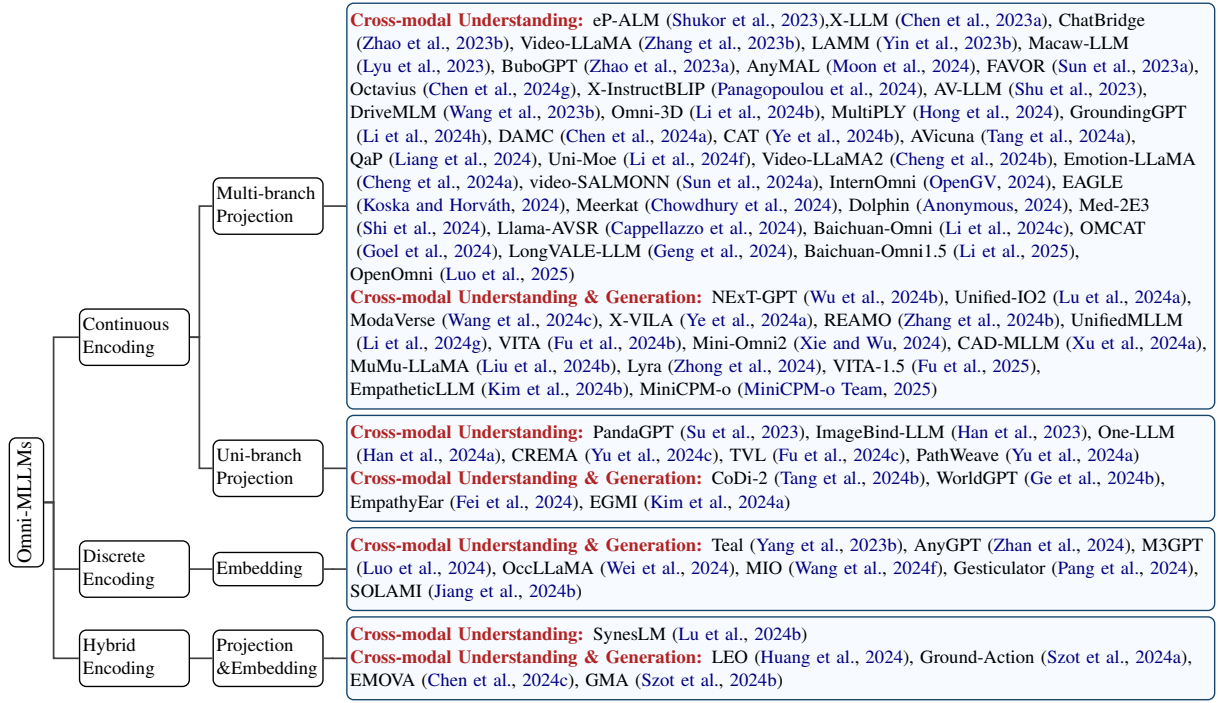

As the extension of Specific-MLLMs, Omni-MLLMs inherit the architecture of \textit{encoding, alignment, interaction, and generation} and broaden the types of non-linguistic modalities involved. This section introduces the implementation methods and functions of the four components in Omni-MLLM: Multi-modalities Encoding~(\S\ref{section:omni_mllm_encoding}), Multi-modalities Alignment~(\S\ref{section:omni_mllm_connector}), Multi-modalities Interaction~(\S\ref{section:omni_mllm_interaction}), and Multi-modalities Generation~(\S\ref{section:omni_mllm_generation}). More details about the architecture of Omni-MLLMs are shown in Appendix~\ref{section:omnimllm_architures_details}. 

\subsection{Multi-modalities Encoding}\label{section:omni_mllm_encoding}
Based on the encoding feature spaces of multiple modalities, we categorize the Omni-MLLM encoding methods into three types: 1) continuous encoding, 2) discrete encoding, and 3) hybrid encoding.
\subsubsection{Continuous Encoding}\label{section:omni_mllm_encoding_continuous}
Continuous encoding refers to encoding the modality into the continuous feature space. Omni-MLLMs that adopt continuous encoding, such as X-LLM~\citep{xllm2023chen} and ChatBridge~\citep{chatbridge}, often integrate multiple pre-trained uni-modality encoders. These modality-specific encoders encode different modalities $\boldsymbol{X}$ into distinct feature spaces $\mathbb{R}_{x}$ as $\mathbf{F}_x$, formulated as:
\begin{equation}
  \begin{aligned}
    \mathbf{F}_x = \operatorname{SpecificEncoder}\left(
    \boldsymbol{X}
    \right),\ \ 
    \mathbf{F}_x \in \mathbb{R}_x
\label{eq:continous_encoding_specific}
    \end{aligned}
\end{equation}where SpecificEncoder refers to different modality-specific encoders used in Omni-MLLMs, such as InternVit~\citep{internvl} for encoding visual modality, Whisper~\citep{Whisper} for encoding auditory modality, ULIP-2~\citep{ulip2} for encoding 3D modality, IMU2CLIP~\citep{IMU2CLIP} for encoding IMU modality, etc.

Besides using heterogeneous encoders for continuous encoding, some Omni-MLLMs~\citep{onellm,imagebindllm,pandagpt,TVL} employ pre-aligned encoders for multiple modalities, encoding different modalities $\boldsymbol{X}$ into the same feature space $\mathbb{R}_{uni}$, as shown in Equation~\ref{eq:continous_encoding_prealign}.\begin{equation}
  \begin{aligned}
    \mathbf{F}_x = \operatorname{PreAlignEncoder}\left(
    \boldsymbol{X}
    \right),\ \ 
    \mathbf{F}_x \in \mathbb{R}_{uni}
\label{eq:continous_encoding_prealign}
  \end{aligned}
\end{equation}where $\operatorname{PreAlignEncoder} $ refer to encoders that uniformly encode multiple modalities, such as LanguageBind~\citep{languagebind} which uses text as a bridge to align different modalities, and ImageBind~\cite{imagebind} which uses images as a bridge to align different modalities.
\subsubsection{Discrete Encoding}
To better facilitate the seamless integration and generation of new non-linguistic modalities, some Omni-MLLMs, such as AnyGPT~\citep{anygpt} and Teal~\citep{teal}, adopt a discrete encoding approach. This method encodes different raw modalities $\boldsymbol{X}$ into the same discrete token space $\mathbb{V}_{uni}$ as $\mathbf{T}_x$, formulated as follows: \begin{equation}
  \begin{aligned}
    \mathbf{T}_x = \operatorname{SpecificTokenizer}\left(
    \boldsymbol{X}
    \right),\ \ 
    \mathbf{T}_x \in \mathbb{V}_{uni}
\label{eq:diserect_encoding}
  \end{aligned}
\end{equation}where $\operatorname{SpecificTokenizer}$ refers to different modality-specific tokenizers used in Omni-MLLMs, including the SEED tokenizer~\citep{seed_tokenizer} based on Vector Quantized Tokenization (VQ), the SpeechTokenizer~\citep{SpeechTokenizer} based on Residual Vector Quantized Tokenization (RVQ), the AudioTokenizer of Teal~\citep{teal} based on k-means clustering, and so on.

\subsubsection{Hybrid Encoding}
Although discrete encoding facilitates the unified processing of different non-linguistic modalities and text compared to continuous encoding, discrete modality tokens often struggle to capture the detailed information inherent in raw continuous modalities~\citep{emova,Mini-Omni2}. 
Therefore, some Omni-MLLMs combine both encoding approaches instead of a fully discretized manner, choosing different encoding methods for different modalities. For instance, EMOVA~\citep{emova} uses the discrete S2U tokenizer to encode auditory modalities while employing the continuous encoder InternVit for visual modalities to retain more vision semantic information. Similarly, GroundAction~\cite{groundaction} encodes visual modalities using the CLIP Vit and action modalities with its trained action tokenizer.

\subsection{Multi-modalities Alignment}\label{section:omni_mllm_connector}
Omni-MLLMs align the encoded features of various non-linguistic modalities with the embedding space of LLMs. The multi-modality alignment can be categorized into two approaches: 1) projection alignment and 2) embedding alignment.
\subsubsection{Projection Alignment}
The continuous encoding Omni-MLLMs insert adapters, referred to as \textit{projectors}, between the encoders and the LLMs. These projectors map the continuously encoded modality features $\mathbf{F}_x$ into the text embedding space as $\mathbf{F}_{p}$. As discussed in Section~\ref{section:omni_mllm_encoding_continuous}, $\mathbf{F}_x$ may either reside in distinct feature spaces $\mathbb{R}_{x}$ or share the same feature space $\mathbb{R}_{uni}$. 
For the former, multiple projectors are typically employed to align the $\mathbf{F}_x$ of each modality into $\mathbb{R}_{t}$ as $\mathbf{F}_{p}$ independently, addressing dimensional mismatch and feature misalignment across modalities~\cite{xvila,macawllm,anymal}, formulated as follows: \begin{equation}
\label{eq:continous_projection_multi_branch}
    \mathbf{F}_{p} = \operatorname{SpecificProjector}(
    \mathbf{F}_x
    ),\ 
    \mathbf{F}_{p} \in \mathbb{R}_{t}
\end{equation}where $\operatorname{SpecificProjector}$ refers to the modality-specific projector corresponding to different modalities, called \textit{multi-branch projection}. 

\begin{figure}[t]
    \centering
    \includegraphics[width=1\linewidth]{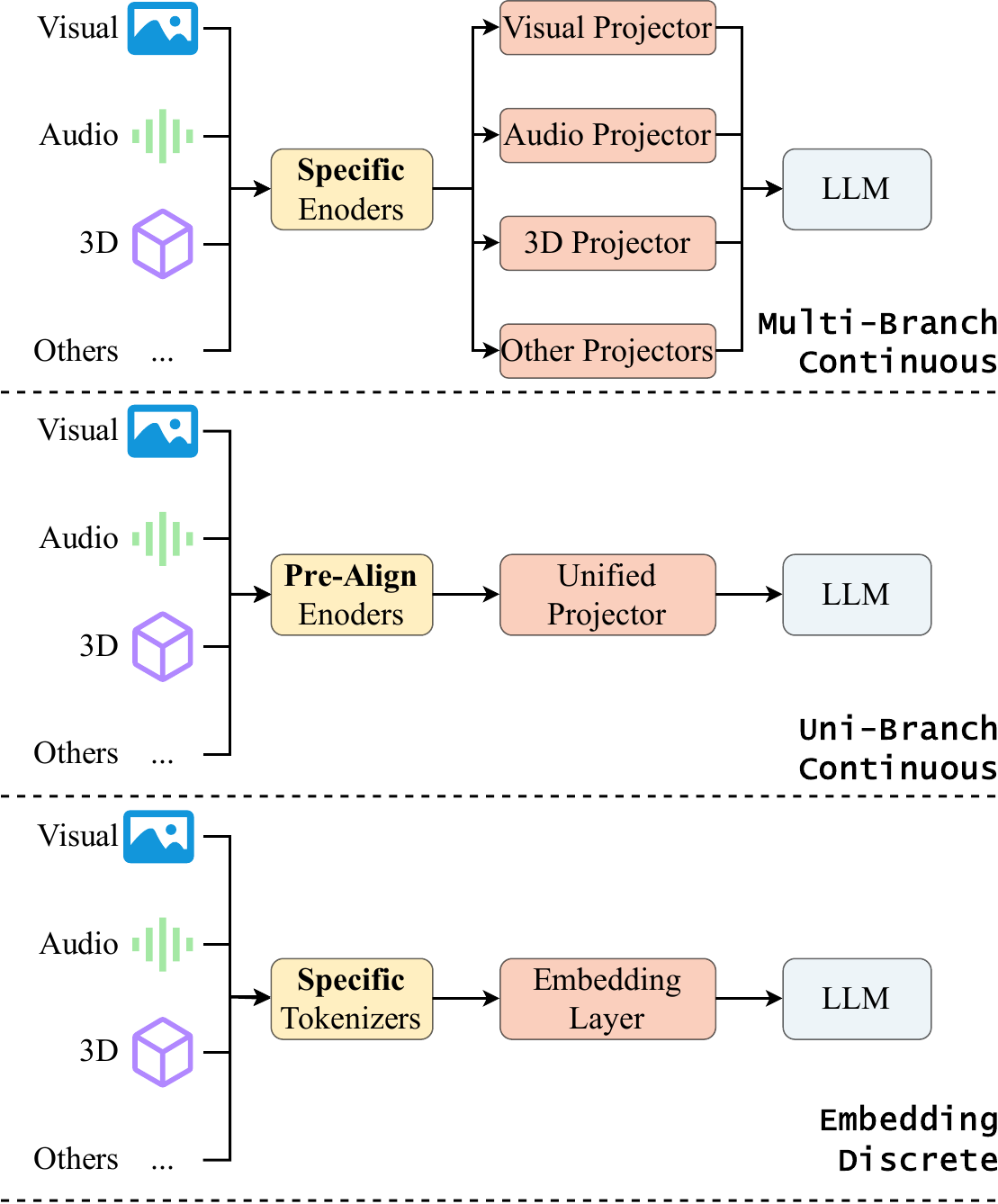}
    \caption{The three combinations of encoding and alignment in Omni-MLLM are based on different encoding spaces and alignment structures.}
    \label{fig:branch}
\end{figure}

For the latter case, besides the multi-branch approach, Omni-MLLMs like PandaGPT~\citep{pandagpt} and WorldGPT~\citep{worldgpt} adopt a shared projector to achieve unified alignment across modalities to reduce the parameters of multiple projectors, as shown in Equation~\ref{eq:continous_projection_uni_branch}.
\begin{equation}
\label{eq:continous_projection_uni_branch}
    \mathbf{F}_{p} = \operatorname{UnifiedProjector}(
    \mathbf{F}_x
    ),\ 
    \mathbf{F}_{p} \in \mathbb{R}_{t}
\end{equation}where $\operatorname{UnifiedProjector}$ refers to the unified projector used to align multiple modalities, a design known as the \textit{uni-branch projection}. A comparison of the two approaches is illustrated in Figure~\ref{fig:branch}.

In terms of the \textbf{\textit{specific implementation}} of the projector, the most straightforward approach is to use a multi-layer perceptron (MLP) or a single linear layer~\citep{nextgpt,videollama2,interomni}. Alternatively, attention mechanisms can be employed to compress the encoded information of non-linguistic modalities. This includes cross-attention-based methods like Q-Former~\citep{xinstruclblip,xllm2023chen} and Perceiver~\citep{chatbridge,qap}, as well as self-attention-based methods such as UPM in OneLLM~\citep{onellm}. Additionally, BaiChuan-Omni~\citep{baichuan-omni} and EMOVA~\citep{emova} incorporate CNNs to compress the projected features, thereby achieving locality preservation~\citep{honeybee}.

It is also worth noting that in multi-branch Omni-MLLMs, different branches may utilize distinct implementations to better accommodate the unique characteristics of each modality~\citep{groundinggpt}. For example, Uni-MoE~\cite{unimoe} uses a linear projection for the visual modality and a Q-Former for the auditory modality. Meanwhile, uni-branch Omni-MLLMs, when using an attention-based projector, typically design multiple modality-specific learnable vectors to extract key information from various non-linguistic modalities~\citep{PathWeave,onellm,CREMA}.

\subsubsection{Embedding Alignment}

As for discrete encoding Omni-MLLMs, the features of non-linguistic modalities are represented as quantized codes, which reside in the same discrete space $\mathbb{V}_{uni}$ as text tokens. Therefore, new modality-specific discrete tokens $\mathbf{T}_x$ are embedded into the continuous feature space $\mathbb{R}_{t}$ by modifying the vocabulary of LLMs and the corresponding embeddings layer, as shown in Equation~\ref{eq:diserect_projection}.\begin{equation}
\label{eq:diserect_projection}
    \mathbf{F}_{p} = \operatorname{Embedding}(
    \mathbf{T}_x
    ),\ 
    \mathbf{F}_{p} \in \mathbb{R}_{t}
\end{equation}where $\operatorname{Embedding}$ refers to the unified embedding layer corresponding to different modalities, which is typically achieved by adding discrete codebooks from various modalities to the vocabulary and expanding the embedding layer of LLMs~\citep{anygpt,teal,occllama}. For instance, AnyGPT extends the vocabulary of the LLaMA-2 by incorporating 17,408 codes across three modalities—image, speech, and music~\citep{anygpt}. Besides, some works like Ground-Action~\citep{groundaction} and LEO~\citep{LEO} overwrite infrequently used tokens in the original vocabulary for alignment, as they extend a smaller set of modality-specific discrete tokens.

Additionally, for hybrid encoding models, alignment is achieved by simultaneously employing both the projection method and the embedding method~\cite{emova,GMA}.


\begin{figure}[t]
    \centering
    \includegraphics[width=1\linewidth]{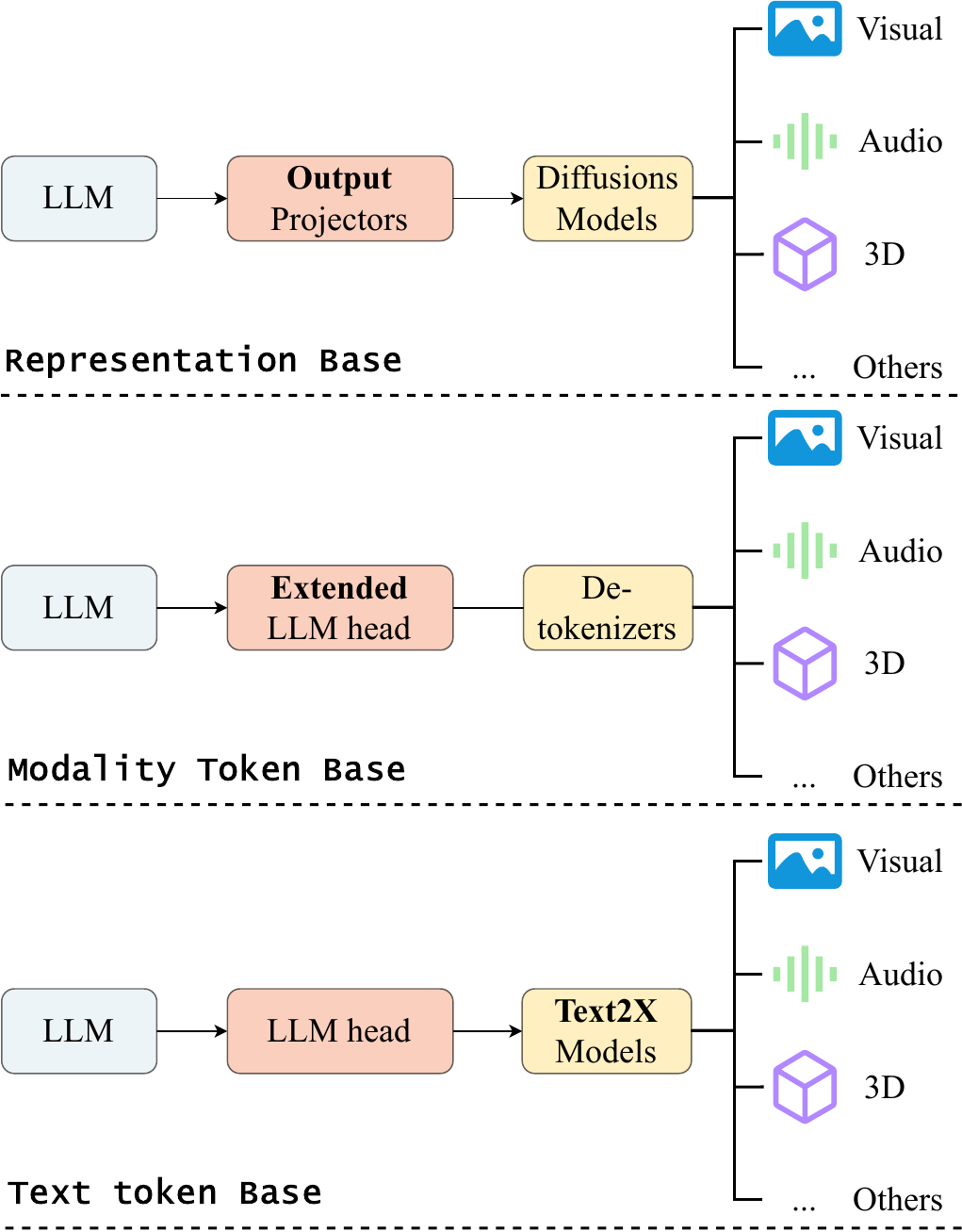}
    \caption{Three generation methods in Omni-MLLMs are implemented based on different output spaces of LLM and the corresponding generative models.}
    \label{fig:generate}
\end{figure}
\subsection{Multi-modalities Interaction}\label{section:omni_mllm_interaction}
Omni-MLLMs utilize transformer-based LLMs to facilitate information interaction between different modalities within a unified feature space $\mathbb{R}_{t}$. Commonly used LLMs include the LLaMA series~\citep{llama}, the Qwen series~\citep{qwen}, and others~\citep{internlm2,chatglm}. 

For interaction, most Omni-MLLMs~\cite{xllm2023chen,xvila,interomni,baichuanomni1.5} concatenate aligned non-linguistic modality features $\mathbf{F}_{p}$ with textual features $\mathbf{F}_{t}$ at the input level, enabling interaction in a progressive and layer-by-layer manner. Meanwhile, some works, such as ImageBind-LLM~\citep{imagebindllm} and TVL-LLaMA~\citep{TVL}, insert $\mathbf{F}_{p}$ into specific layers or all layers of the LLMs to mitigate the loss of modality information~\citep{ePalm}.

In terms of the number of modalities involved in interactions, compared to Specific-MLLMs that are limited to dual-modal interactions between a single non-linguistic modality and text~\citep{llava,pointllm}, Omni-MLLMs not only support multiple dual-modal interactions but also enable omni-multimodal interactions involving more than two non-linguistic modalities~\citep{chatbridge,MIO}. For example, X-InstructBLIP~\citep{xinstruclblip} enables dual-modal interactions such as vision-text, audio-text, and 3D-text, as well as omni-modal interactions like vision-audio-text and 3D-vision-text, showcasing the ability of Omni-MLLMs to handle arbitrary combinations of modalities.
\subsection{Multi-modalities Generation}\label{section:omni_mllm_generation}

Omni-MLLMs can output text while also generating non-linguistic modalities by integrating different generation models. As shown in Figure~\ref{fig:generate}, we categorize multi-modalities generation into three types: text-based generation, representation-based generation, and modality-token-based generation.

\paragraph{Text-based}This approach directly utilizes the discrete text output from the LLM to invoke Text-to-X generation models~\citep{audioldm,VideoFusion,InstructPix2Pix} based on the content of the text. For example, VITA~\citep{vita} employs TTS tools~\citep{gpt-sovits} to convert the output text into corresponding speech, while ModelVerse~\citep{modaverse} and UnifiedMLLM~\citep{unifiedmllm} use the text to specify the generation model and utilize the corresponding descriptions to generate different modalities.

\paragraph{Modality-Token-based}Works like MiniOmni-2~\citep{Mini-Omni2} and AnyGPT~\citep{anygpt} extend the corresponding LLM head with codebooks from different modality tokenizers to generate modality-specific discrete tokens. These tokens are then decoded using the corresponding de-tokenizers~\citep{vqgan,magvit-v2,soundstream,jukebox} to produce various modalities.

\paragraph{Representation-based}To alleviate the potential noise introduced by discrete tokens, works like X-VILA~\citep{xvila} and NextGPT~\citep{nextgpt} incorporate modality-specific signal tokens into the vocabulary. They then use transformers or MLPs to map the signal token representations into the ones that are understandable to the multimodal decoders, typically off-the-shelf latent-conditioned diffusion models~\citep{stablediffusion,CoDi,auffsion,stablevideo}, enabling effective generation capabilities.

\section{Omni-MLLM Training}\label{section:omni_mllm_training}
To achieve alignment across different vector spaces and improve instruction-following ability under arbitrary modality settings, Omni-MLLMs extend the standard two-stage training pipeline of Specific-MLLMs: \textit{multi-modalities alignment pre-training} and \textit{multi-modalities instruction fine-tuning}.

\subsection{Multi-modalities Alignment Pre-training}
Multi-modalities alignment pre-training involves \textit{input alignment} training between the feature spaces of different modalities and the embedding space of LLMs on the encoding side, as well as \textit{output alignment} training between the embedding space and the input spaces of various modality decoders on the decoding side. Input alignment and output alignment can be carried out separately~\citep{nextgpt} or simultaneously~\citep{xvila}.
\subsubsection{Input Alignment}
Input alignment mainly uses X-Text paired datasets of different modalities and minimizes the text generation loss of the corresponding description text to optimize. In this phase, continuous encoding Omni-MLLMs normally update parameters of projectors, while discrete encoding Omni-MLLMs adjust the parameters of the embedding layer.


In terms of \textbf{\textit{training order}} of different modalities alignment, multi-branch Omni-MLLM performs separate alignment training for each modality-specific projector, directly aligning each non-linguistic modality with text and using text as a bridge to align different non-linguistic modalities~\citep{chatbridge,xinstruclblip}. The uni-branch Omni-MLLM, on the other hand, uses the unified projector for different modalities, which may lead to interference in the alignment performance between different modalities. Thus, \citet{onellm} employ a progressive alignment strategy to align multiple modalities in a specific order. In contrast, discrete encoding Omni-MLLMs, like AnyGPT~\citep{anygpt} and M3GPT~\citep{M3GPT}, mix the alignment data from different modalities and perform alignment simultaneously.

Besides, in addition to directly leveraging X-Text paired datasets from different modalities for direct alignment, PandaGPT~\citep{pandagpt}, ImageBind-LLM~\citep{imagebindllm}, and VideoLLaMA~\citep{videollama} utilize the pre-aligned modality feature space $\mathbb{V}_{uni}$ to achieve indirect alignment between other non-linguistic modalities and text by training solely on Image-Text data.

\subsubsection{Output Alignment}\label{section:omni_mllm_training_output_alignment}
The training of output alignment typically utilizes the same X-text paired dataset as input alignment and adheres to the identical training sequence. Meanwhile, the training objectives for output alignment vary depending on the multi-modality generation methods in Section~\ref{section:omni_mllm_generation}. Token-based generative Omni-MLLMs optimize the extended LLM head by minimizing the text generation loss associated with modality-specific discrete tokens~\citep{unifiedio2,occllama}. Representation-based generative Omni-MLLMs generally optimize their output projectors by minimizing the composite loss comprising three components~\citep{Mini-Omni2,teal}: 1) the text generation loss of signal tokens; 2) the L2 distance between the output representation and the condition vector of the corresponding decoder, i.e. MSE loss; and 3) the conditional latent denoising loss~\cite{stablediffusion}. For text-based generative Omni-MLLMs, as there is no additional output structure, the output alignment training is generally not required~\citep{modaverse,unifiedmllm}.\looseness=0

\subsection{Multi-modalities Instruction Fine-tuning}
The instruction fine-tuning phase aims to enhance generalization capability under arbitrary modalities of Omni-MLLMs~\citep{xinstruclblip,nextgpt,xvila}. Instruction fine-tuning primarily utilizes instruction-following datasets and computes the text generation loss for the corresponding responses to optimize. For models with generation capabilities, the loss mentioned in section~\ref{section:omni_mllm_training_output_alignment} may also be incorporated. During this phase, Omni-MLLMs further perform full-scale tuning of the LLM parameters~\citep{onellm,videollama2} or use PEFT techniques~\citep{PEFT_Survey}, such as LoRA~\citep{LORA}, for partial tuning~\citep{internvideo2}.

Compared to Specific-MLLMs, Omni-MLLMs not only leverage multiple uni-modal instruction data of different modalities for training but also use cross-modal instruction data to enhance their cross-modal ability~\citep{xvila,anygpt,baichuan-omni}. In addition to directly mixing different instruction data for training~\citep{xinstruclblip,xvila}, some works like Uni-Moe~\citep{unimoe} and Lyra~\citep{Lyra} adopt a multi-step fine-tuning approach, introducing different uni-modal and cross-modal instruction data in a specific order for training to gradually enhance their uni-modal and cross-modal ability.

\subsection{Other Train Recipes}
In addition to the general training paradigms mentioned in Section~\ref{section:omni_mllm_training}, some other useful training recipes are also used. (1)\textbf{ Prior knowledge from Specific-MLLMs}: Since Specific-MLLMs have already achieved effective alignment in single-modal scenarios, some Omni-MLLMs directly leverage their well-trained projectors to reduce the training overhead during the alignment phase. For example, InstructBLIP~\cite{xinstruclblip} and X-LLM~\cite{xllm2023chen} use the Q-former trained by BLIP2 to align the visual modality, while NaviveMC and DAMC~\cite{ModelComposition} further leverage projectors from multiple models to handle alignment for visual, audio, and 3D modalities separately; (2)\textbf{ Additional human preference training}: \citet{GMA} and \citet{cat} adopt HF training methods like PPO and ADPO to better align with human preferences; (3)\textbf{ Modalities Blending}: During progressive alignment pre-training or multi-step instruction fine-tuning, some works~\cite{onellm,baichuan-omni,emova} mix previously trained modality data with the current new modality data for training to prevent catastrophic forgetting.
\section{Data Construction and Evaluation}\label{section:omni_mllm_resource}
This section summarizes the construction of modality alignment data and instruction data used in the Omni-MLLM training process~(\S\ref{section:omni_mllm_data_construction}), as well as the evaluation across four different capabilities~(\S\ref{section:omni_mllm_benchmark}).
\subsection{Training Data}\label{section:omni_mllm_data_construction}
\paragraph{Alignment Data}
Omni-MLLMs leverage caption datasets from various modalities to construct X-Text paired data for alignment pre-training, such as the WebVid~\citep{webvid} for visual modality and the AudioCaps~\citep{audiocap} for auditory modality. However, for data-scarce modalities like depth maps and thermal maps, large-scale text-paired data is lacking~\citep{languagebind,imagebind}. To address this, synthetic methods that use DPT models~\cite{DPT,ZoeDepth,unimatch} or image translation models~\citep{rgb2ir} to convert image-text pairs into other modality text pairs are widely employed~\cite{onellm,emova,languagebind}. Moreover, interleaved datasets~\citep{Multimodal_c4} are used for alignment pre-training in some works~\citep{codi2} to enhance the contextual understanding capability.  
\paragraph{Instruction Data}

Omni-MLLMs not only leverage uni-modal instruction datasets from Specific-MLLMs, but also construct cross-modal instruction data through diverse methods as follows.

\noindent(1)\textbf{ Template-based Construction}: Most works \cite{favor,chatbridge,REAMO} utilize cross-modal downstream datasets~\citep{how2,vggsound} combined with predefined templates to construct cross-modal instructions; (2)\textbf{ GPT Generation}: Following the paradigm of LLaVA~\citep{llava}, some Omni-MLLMs~\cite{macawllm,chatbridge} leverage the labels from the annotated dataset~\citep{MSCOCO,webvid} or use pre-trained models like SAM~\citep{sam} and GRIT~\citep{GRIT} to extract meta-information (e.g., captions and object categories) of different modalities. Then they employ powerful LLMs~\citep{gpt4,chatgpt} to generate cross-modal instructions based on the obtained meta-information; (3)\textbf{ T2X Generation}: \citet{unimoe} use TTS tools to convert the Image-Text2Text uni-modal instructions from LLaVA-v1.5~\citep{llava1.5} into Image-Speech-Text2Text cross-modal instructions. AnyGPT~\citep{anygpt} and NextGPT~\citep{nextgpt} leverage Text2X models such as DALL-E-3~\citep{dalle3} and MusicGen~\citep{musicgen} to convert the GPT-generated pure text instructions into Xs2Xs cross-modal instructions. Details about training data are shown in Appendix~\ref{section:omnimllm_traing_data_details}

\subsection{Benchmark}\label{section:omni_mllm_benchmark}
We provide a brief overview of the benchmarks used to evaluate Omni-MLLMs. The statistics of the benchmarks are shown in Appendix~\ref{section:omnimllm_benchmark_details}. 
\paragraph{Uni-modal Understanding} 

Uni-modal understanding assesses the ability of Omni-MLLMs to comprehend and reason on different non-linguistic modalities, including downstream X-Text2Text datasets such as X-Caption~\cite{flicker30k,msrvtts}, X-QA~\cite{VQAv2,msvd}, and X-Classification~\cite{objaverse}, as well as comprehensive multi-task benchmarks~\cite{mmb,videomme,MME}.

\paragraph{Uni-modal Generation}
Uni-modal generation aims to evaluate the ability of Omni-MLLMs to generate a single non-linguistic modality, including the Text2X generation task~\citep{audiocap,dreambench} and the Text-X2Text editing task~\cite{VCTK,DAVIS}.

\paragraph{Cross-modal Understanding} 
Cross-modal understanding evaluates the ability of Omni-MLLMs to jointly comprehend and reason across multiple non-linguistic modalities like Image-Speech-Text2Text~\citep{unimoe,interomni}, Video-Audio-Text2Text~\citep{AVQA,musicavqa}, and Image-3D-Text2Text~\citep{xinstruclblip}.

\paragraph{Cross-modal Generation} 

Cross-modal generation further evaluates the ability of Omni-MLLMs to generate non-linguistic modalities in conjunction with other non-linguistic modality inputs. For example, the Xs-Text2X benchmark proposed by X-VILA~\citep{xvila} includes tasks such as Image-Text2Audio and Image-Audio-Text2Video.

\section{Challenges and Future Directions}\label{section:omni_mllm_future}
Despite Omni-MLLMs having showcased remarkable performance on numerous tasks, there are still some challenges that necessitate further research.

\subsection{Expansion of modalities}
Most Omni-MLLMs can only process 2-3 types of non-linguistic modalities, and they still face several challenges when expanding more modalities.
\paragraph{Training efficiency} The common method that introduces new modalities through additional alignment pre-training and instruction fine-tuning can lead to significant training cost. Leveraging prior knowledge from Specific-MLLMs~\citep{xinstruclblip,ModelComposition} or using pre-aligned encoders for indirect alignment~\citep{imagebindllm,pandagpt} can help reduce training overhead but may impact cross-modal performance. 
\paragraph{Catastrophic forgetting} Expanding new modalities may adjust the shared parameters, potentially causing catastrophic forgetting of previously trained modalities knowledge~\citep{PathWeave}. This issue can be partially mitigated by mixing trained modality data~\citep{onellm,unimoe} or fine-tuning only the modality-specific parameters~\cite{PathWeave,CREMA}, but both approaches make the training process more complex. 
\paragraph{Low-resource modalities} Although the data synthesis method in Section~\ref{section:omni_mllm_data_construction} can help alleviate the lack of text-paired data and instruction data for low-resource modalities~\citep{onellm,infraredllava}, the absence of real modality may lead to biases in understanding of that modality.
\subsection{Cross-modal capabilities}

The Omni-MLLMs have achieved promising performance in cross-modal understanding and generation tasks, but there are still some challenges. 
\paragraph{Long Context} When the input contains multiple sequence modalities (video, speech...), the length of the multi-modalities token sequence may exceed the context window of LLMs and lead to memory overflow. While methods such as token compressing~\citep{CREMA,baichuan-omni} or token sampling~\citep{anygpt,Lyra} can reduce the number of input tokens, they also result in a decline in cross-modal performance.
\paragraph{Modality Bias} Due to the imbalance in training data volume and the performance disparity among different modality encoders, Omni-MLLMs may tend to pay attention to the dominant modality while neglecting information from other modalities during cross-modal inference. Balancing the data volume across modalities or enhancing the corresponding modality-specific modules could potentially help mitigate this issue~\citep{curse}.
\paragraph{Temporal Alignment} When dealing with different modalities that have temporal dependencies, retaining their temporal alignment information is crucial for subsequent cross-modal understanding. Some attempts have been made to preserve the temporal alignment information between audio and video, such as interleaved modality-specific tokens of video and audio~\citep{avicuna} and inserting the time-related special tokens into the multi-modalies tokens~\citep{omcat}.
\paragraph{Data and Benchmark}
Although Omni-MLLMs employ various methods in Section~\ref{section:omni_mllm_data_construction} to generate cross-modal instruction data, there is still significant room for improvement and expansion, including enhancing the diversity of instructions, incorporating longer contextual dialogues, and exploring more diverse modality interaction paradigms. Similarly, cross-modal benchmarks such as OmniBench~\citep{li2024omnibench} and OmniR~\citep{omnixr} still fall short in terms of task richness and instruction diversity when compared to uni-modal benchmarks like MMMU~\citep{MMMU} and MME~\citep{MME}. And the variety of modalities they cover is also relatively limited.
\subsection{Application scenarios}

The emergence of Omni-MLLM brings new opportunities and possibilities for various applications. \textbf{(1) Real-time Multi-modalities Interaction}: \citet{vita-1.5} and \citet{Mini-Omni2} achieve robust capabilities in both vision and speech understanding, enabling efficient speech-to-speech interactions with vision in real-time. \textbf{(2) Comprehensive Planning}: \citet{drivemlm} and \citet{groundaction} leverage the complementarity across multiple modalities to achieve better path planning and action planning capabilities than planning with vision information only. 
\textbf{(3) World Simulator}: \citet{worldgpt} not only understands and generates different modalities but also predicts state transitions for any combination of modalities.
\section{Conclusion}\label{section:omni_mllm_conclusion}
In this paper, we provide a comprehensive survey report on Omni-MLLM, offering a comprehensive review of the field. Specifically, we break down Omni-MLLM into four key components and categorize them based on modal encoding and alignment methods. Subsequently, we provide a detailed summary of the training process of Omni-MLLM and the related resources used. We also summarize the current challenges and the future development directions. This paper is the first systematic survey dedicated to Omni-MLLMs. We hope this survey will facilitate further research in this area.

\section*{Limitations}
This study provides the first comprehensive survey of Omni-MLLMs. Related work, architecture statistics, more details of training and evaluation, as well as other training recipes, can be found in Appendix~\ref{section:appendix_related},\ref{section:omnimllm_architures_details},\ref{section:omnimllm_traing_training_evaluation_details}.

We have made our best effort, but there may still be some limitations. On one hand, due to page limitations, we can only provide a concise overview of the core contributions of mainstream Omni-MLLMs, rather than exhaustive technical details. On the other hand, our review primarily covers research from *ACL, NeurIPS, ICLR, ICML, COLING, CVPR, IJCAI, ECCV, and arXiv, and there is a chance that we may have missed some important work published in other venues. We will stay updated with ongoing discussions in the research community and plan to revise our work in the future to include overlooked contributions.



\bibliography{custom}

\begin{thebibliography}{297}
\providecommand{\natexlab}[1]{#1}

\bibitem[{Achlioptas et~al.(2020)Achlioptas, Abdelreheem, Xia, Elhoseiny, and Guibas}]{NR3D}
Panos Achlioptas, Ahmed Abdelreheem, Fei Xia, Mohamed Elhoseiny, and Leonidas~J. Guibas. 2020.
\newblock \href {https://doi.org/10.1007/978-3-030-58452-8\_25} {Referit3d: Neural listeners for fine-grained 3d object identification in real-world scenes}.
\newblock In \emph{Computer Vision - {ECCV} 2020 - 16th European Conference, Glasgow, UK, August 23-28, 2020, Proceedings, Part {I}}, volume 12346 of \emph{Lecture Notes in Computer Science}, pages 422--440. Springer.

\bibitem[{Agostinelli et~al.(2023)Agostinelli, Denk, Borsos, Engel, Verzetti, Caillon, Huang, Jansen, Roberts, Tagliasacchi, Sharifi, Zeghidour, and Frank}]{musiccap}
Andrea Agostinelli, Timo~I. Denk, Zal{\'{a}}n Borsos, Jesse~H. Engel, Mauro Verzetti, Antoine Caillon, Qingqing Huang, Aren Jansen, Adam Roberts, Marco Tagliasacchi, Matthew Sharifi, Neil Zeghidour, and Christian~Havn{\o} Frank. 2023.
\newblock \href {https://doi.org/10.48550/ARXIV.2301.11325} {Musiclm: Generating music from text}.
\newblock \emph{CoRR}, abs/2301.11325.

\bibitem[{Agrawal et~al.(2019)Agrawal, Anderson, Desai, Wang, Chen, Jain, Johnson, Batra, Parikh, and Lee}]{nocaps}
Harsh Agrawal, Peter Anderson, Karan Desai, Yufei Wang, Xinlei Chen, Rishabh Jain, Mark Johnson, Dhruv Batra, Devi Parikh, and Stefan Lee. 2019.
\newblock \href {https://doi.org/10.1109/ICCV.2019.00904} {nocaps: novel object captioning at scale}.
\newblock In \emph{2019 {IEEE/CVF} International Conference on Computer Vision, {ICCV} 2019, Seoul, Korea (South), October 27 - November 2, 2019}, pages 8947--8956. {IEEE}.

\bibitem[{AlAmri et~al.(2018)AlAmri, Cartillier, Lopes, Das, Wang, Essa, Batra, Parikh, Cherian, Marks, and Hori}]{AVSD}
Huda AlAmri, Vincent Cartillier, Raphael~Gontijo Lopes, Abhishek Das, Jue Wang, Irfan Essa, Dhruv Batra, Devi Parikh, Anoop Cherian, Tim~K. Marks, and Chiori Hori. 2018.
\newblock \href {https://arxiv.org/abs/1806.00525} {Audio visual scene-aware dialog {(AVSD)} challenge at {DSTC7}}.
\newblock \emph{CoRR}, abs/1806.00525.

\bibitem[{Allen et~al.(2022)Allen, St-Yves, Wu, Breedlove, Prince, Dowdle, Nau, Caron, Pestilli, Charest et~al.}]{nsd}
Emily~J Allen, Ghislain St-Yves, Yihan Wu, Jesse~L Breedlove, Jacob~S Prince, Logan~T Dowdle, Matthias Nau, Brad Caron, Franco Pestilli, Ian Charest, et~al. 2022.
\newblock A massive 7t fmri dataset to bridge cognitive neuroscience and artificial intelligence.
\newblock \emph{Nature neuroscience}, 25(1):116--126.

\bibitem[{Anonymous(2024)}]{Dolphin}
Anonymous. 2024.
\newblock \href {https://openreview.net/forum?id=1SYUKPeM12} {Aligned better, listen better for audio-visual large language models}.
\newblock In \emph{Submitted to The Thirteenth International Conference on Learning Representations}.
\newblock Under review.

\bibitem[{Ardila et~al.(2020)Ardila, Branson, Davis, Kohler, Meyer, Henretty, Morais, Saunders, Tyers, and Weber}]{commonvoice}
Rosana Ardila, Megan Branson, Kelly Davis, Michael Kohler, Josh Meyer, Michael Henretty, Reuben Morais, Lindsay Saunders, Francis~M. Tyers, and Gregor Weber. 2020.
\newblock \href {https://aclanthology.org/2020.lrec-1.520/} {Common voice: {A} massively-multilingual speech corpus}.
\newblock In \emph{Proceedings of The 12th Language Resources and Evaluation Conference, {LREC} 2020, Marseille, France, May 11-16, 2020}, pages 4218--4222. European Language Resources Association.

\bibitem[{Arnab et~al.(2021)Arnab, Dehghani, Heigold, Sun, Lucic, and Schmid}]{vivit}
Anurag Arnab, Mostafa Dehghani, Georg Heigold, Chen Sun, Mario Lucic, and Cordelia Schmid. 2021.
\newblock \href {https://doi.org/10.1109/ICCV48922.2021.00676} {Vivit: {A} video vision transformer}.
\newblock In \emph{2021 {IEEE/CVF} International Conference on Computer Vision, {ICCV} 2021, Montreal, QC, Canada, October 10-17, 2021}, pages 6816--6826. {IEEE}.

\bibitem[{Azuma et~al.(2022)Azuma, Miyanishi, Kurita, and Kawanabe}]{scanqa}
Daichi Azuma, Taiki Miyanishi, Shuhei Kurita, and Motoaki Kawanabe. 2022.
\newblock \href {https://doi.org/10.1109/CVPR52688.2022.01854} {Scanqa: 3d question answering for spatial scene understanding}.
\newblock In \emph{{IEEE/CVF} Conference on Computer Vision and Pattern Recognition, {CVPR} 2022, New Orleans, LA, USA, June 18-24, 2022}, pages 19107--19117. {IEEE}.

\bibitem[{Bai et~al.(2024{\natexlab{a}})Bai, Du, Huang, Meng, and Zhao}]{M3DCLIP}
Fan Bai, Yuxin Du, Tiejun Huang, Max~Qinghu Meng, and Bo~Zhao. 2024{\natexlab{a}}.
\newblock \href {https://doi.org/10.48550/ARXIV.2404.00578} {{M3D:} advancing 3d medical image analysis with multi-modal large language models}.
\newblock \emph{CoRR}, abs/2404.00578.

\bibitem[{Bai et~al.(2023)Bai, Bai, Chu, Cui, Dang, Deng, Fan, Ge, Han, Huang, Hui, Ji, Li, Lin, Lin, Liu, Liu, Lu, Lu, Ma, Men, Ren, Ren, Tan, Tan, Tu, Wang, Wang, Wang, Wu, Xu, Xu, Yang, Yang, Yang, Yang, Yao, Yu, Yuan, Yuan, Zhang, Zhang, Zhang, Zhang, Zhou, Zhou, Zhou, and Zhu}]{qwen}
Jinze Bai, Shuai Bai, Yunfei Chu, Zeyu Cui, Kai Dang, Xiaodong Deng, Yang Fan, Wenbin Ge, Yu~Han, Fei Huang, Binyuan Hui, Luo Ji, Mei Li, Junyang Lin, Runji Lin, Dayiheng Liu, Gao Liu, Chengqiang Lu, Keming Lu, Jianxin Ma, Rui Men, Xingzhang Ren, Xuancheng Ren, Chuanqi Tan, Sinan Tan, Jianhong Tu, Peng Wang, Shijie Wang, Wei Wang, Shengguang Wu, Benfeng Xu, Jin Xu, An~Yang, Hao Yang, Jian Yang, Shusheng Yang, Yang Yao, Bowen Yu, Hongyi Yuan, Zheng Yuan, Jianwei Zhang, Xingxuan Zhang, Yichang Zhang, Zhenru Zhang, Chang Zhou, Jingren Zhou, Xiaohuan Zhou, and Tianhang Zhu. 2023.
\newblock \href {https://doi.org/10.48550/ARXIV.2309.16609} {Qwen technical report}.
\newblock \emph{CoRR}, abs/2309.16609.

\bibitem[{Bai et~al.(2024{\natexlab{b}})Bai, Wang, Xiao, He, Han, Zhang, and Shou}]{survey_hallucination_1}
Zechen Bai, Pichao Wang, Tianjun Xiao, Tong He, Zongbo Han, Zheng Zhang, and Mike~Zheng Shou. 2024{\natexlab{b}}.
\newblock \href {https://doi.org/10.48550/ARXIV.2404.18930} {Hallucination of multimodal large language models: {A} survey}.
\newblock \emph{CoRR}, abs/2404.18930.

\bibitem[{Bain et~al.(2021)Bain, Nagrani, Varol, and Zisserman}]{webvid}
Max Bain, Arsha Nagrani, G{\"{u}}l Varol, and Andrew Zisserman. 2021.
\newblock \href {https://doi.org/10.1109/ICCV48922.2021.00175} {Frozen in time: {A} joint video and image encoder for end-to-end retrieval}.
\newblock In \emph{2021 {IEEE/CVF} International Conference on Computer Vision, {ICCV} 2021, Montreal, QC, Canada, October 10-17, 2021}, pages 1708--1718. {IEEE}.

\bibitem[{Bertasius et~al.(2021)Bertasius, Wang, and Torresani}]{TimesFormer}
Gedas Bertasius, Heng Wang, and Lorenzo Torresani. 2021.
\newblock \href {http://proceedings.mlr.press/v139/bertasius21a.html} {Is space-time attention all you need for video understanding?}
\newblock In \emph{Proceedings of the 38th International Conference on Machine Learning, {ICML} 2021, 18-24 July 2021, Virtual Event}, volume 139 of \emph{Proceedings of Machine Learning Research}, pages 813--824. {PMLR}.

\bibitem[{Bhat et~al.(2023)Bhat, Birkl, Wofk, Wonka, and M{\"{u}}ller}]{ZoeDepth}
Shariq~Farooq Bhat, Reiner Birkl, Diana Wofk, Peter Wonka, and Matthias M{\"{u}}ller. 2023.
\newblock \href {https://doi.org/10.48550/ARXIV.2302.12288} {Zoedepth: Zero-shot transfer by combining relative and metric depth}.
\newblock \emph{CoRR}, abs/2302.12288.

\bibitem[{Biten et~al.(2019)Biten, Tito, Mafla, G{\'{o}}mez, Rusi{\~{n}}ol, Mathew, Jawahar, Valveny, and Karatzas}]{stvqa}
Ali~Furkan Biten, Rub{\`{e}}n Tito, Andr{\'{e}}s Mafla, Llu{\'{\i}}s G{\'{o}}mez, Mar{\c{c}}al Rusi{\~{n}}ol, Minesh Mathew, C.~V. Jawahar, Ernest Valveny, and Dimosthenis Karatzas. 2019.
\newblock \href {https://doi.org/10.1109/ICDAR.2019.00251} {{ICDAR} 2019 competition on scene text visual question answering}.
\newblock In \emph{2019 International Conference on Document Analysis and Recognition, {ICDAR} 2019, Sydney, Australia, September 20-25, 2019}, pages 1563--1570. {IEEE}.

\bibitem[{Blattmann et~al.(2023)Blattmann, Dockhorn, Kulal, Mendelevitch, Kilian, Lorenz, Levi, English, Voleti, Letts, Jampani, and Rombach}]{stablevideo}
Andreas Blattmann, Tim Dockhorn, Sumith Kulal, Daniel Mendelevitch, Maciej Kilian, Dominik Lorenz, Yam Levi, Zion English, Vikram Voleti, Adam Letts, Varun Jampani, and Robin Rombach. 2023.
\newblock \href {https://doi.org/10.48550/ARXIV.2311.15127} {Stable video diffusion: Scaling latent video diffusion models to large datasets}.
\newblock \emph{CoRR}, abs/2311.15127.

\bibitem[{Brooks et~al.(2023)Brooks, Holynski, and Efros}]{InstructPix2Pix}
Tim Brooks, Aleksander Holynski, and Alexei~A. Efros. 2023.
\newblock \href {https://doi.org/10.1109/CVPR52729.2023.01764} {Instructpix2pix: Learning to follow image editing instructions}.
\newblock In \emph{{IEEE/CVF} Conference on Computer Vision and Pattern Recognition, {CVPR} 2023, Vancouver, BC, Canada, June 17-24, 2023}, pages 18392--18402. {IEEE}.

\bibitem[{Bu et~al.(2017)Bu, Du, Na, Wu, and Zheng}]{AISHELL-1}
Hui Bu, Jiayu Du, Xingyu Na, Bengu Wu, and Hao Zheng. 2017.
\newblock \href {https://doi.org/10.1109/ICSDA.2017.8384449} {{AISHELL-1:} an open-source mandarin speech corpus and a speech recognition baseline}.
\newblock In \emph{20th Conference of the Oriental Chapter of the International Coordinating Committee on Speech Databases and Speech {I/O} Systems and Assessment, {O-COCOSDA} 2017, Seoul, South Korea, November 1-3, 2017}, pages 1--5. {IEEE}.

\bibitem[{Bubeck et~al.(2023)Bubeck, Chandrasekaran, Eldan, Gehrke, Horvitz, Kamar, Lee, Lee, Li, Lundberg, Nori, Palangi, Ribeiro, and Zhang}]{spark_gpt4}
S{\'{e}}bastien Bubeck, Varun Chandrasekaran, Ronen Eldan, Johannes Gehrke, Eric Horvitz, Ece Kamar, Peter Lee, Yin~Tat Lee, Yuanzhi Li, Scott~M. Lundberg, Harsha Nori, Hamid Palangi, Marco~T{\'{u}}lio Ribeiro, and Yi~Zhang. 2023.
\newblock \href {https://doi.org/10.48550/ARXIV.2303.12712} {Sparks of artificial general intelligence: Early experiments with {GPT-4}}.
\newblock \emph{CoRR}, abs/2303.12712.

\bibitem[{Caffagni et~al.(2024)Caffagni, Cocchi, Barsellotti, Moratelli, Sarto, Baraldi, Cornia, and Cucchiara}]{survey_m}
Davide Caffagni, Federico Cocchi, Luca Barsellotti, Nicholas Moratelli, Sara Sarto, Lorenzo Baraldi, Marcella Cornia, and Rita Cucchiara. 2024.
\newblock \href {https://doi.org/10.18653/V1/2024.FINDINGS-ACL.807} {The revolution of multimodal large language models: {A} survey}.
\newblock In \emph{Findings of the Association for Computational Linguistics, {ACL} 2024, Bangkok, Thailand and virtual meeting, August 11-16, 2024}, pages 13590--13618. Association for Computational Linguistics.

\bibitem[{Cai et~al.(2024)Cai, Cao, Chen, Chen, Chen, Chen, Chen, Chen, Chen, Chu, Dong, Duan, Fan, Fei, Gao, Ge, Gu, Gu, Gui, Guo, Guo, He, Hu, Huang, Jiang, Jiao, Jin, Lei, Li, Li, Li, Li, Li, Li, Liu, Liu, Hong, Liu, Liu, Liu, Lv, Lv, Lv, Ma, Ma, Ma, Ning, Ouyang, Qiu, Qu, Shang, Shao, Song, Song, Sui, Sun, Sun, Tang, Wang, Wang, Wang, Wang, Wang, Wang, Wang, Wei, Weng, Wu, Xiong, Zhao, and et~al.}]{internlm2}
Zheng Cai, Maosong Cao, Haojiong Chen, Kai Chen, Keyu Chen, Xin Chen, Xun Chen, Zehui Chen, Zhi Chen, Pei Chu, Xiaoyi Dong, Haodong Duan, Qi~Fan, Zhaoye Fei, Yang Gao, Jiaye Ge, Chenya Gu, Yuzhe Gu, Tao Gui, Aijia Guo, Qipeng Guo, Conghui He, Yingfan Hu, Ting Huang, Tao Jiang, Penglong Jiao, Zhenjiang Jin, Zhikai Lei, Jiaxing Li, Jingwen Li, Linyang Li, Shuaibin Li, Wei Li, Yining Li, Hongwei Liu, Jiangning Liu, Jiawei Hong, Kaiwen Liu, Kuikun Liu, Xiaoran Liu, Chengqi Lv, Haijun Lv, Kai Lv, Li~Ma, Runyuan Ma, Zerun Ma, Wenchang Ning, Linke Ouyang, Jiantao Qiu, Yuan Qu, Fukai Shang, Yunfan Shao, Demin Song, Zifan Song, Zhihao Sui, Peng Sun, Yu~Sun, Huanze Tang, Bin Wang, Guoteng Wang, Jiaqi Wang, Jiayu Wang, Rui Wang, Yudong Wang, Ziyi Wang, Xingjian Wei, Qizhen Weng, Fan Wu, Yingtong Xiong, Xiaomeng Zhao, and et~al. 2024.
\newblock \href {https://doi.org/10.48550/ARXIV.2403.17297} {Internlm2 technical report}.
\newblock \emph{CoRR}, abs/2403.17297.

\bibitem[{Cappellazzo et~al.(2024)Cappellazzo, Kim, Chen, Ma, Petridis, Falavigna, Brutti, and Pantic}]{Llama-AVSR}
Umberto Cappellazzo, Minsu Kim, Honglie Chen, Pingchuan Ma, Stavros Petridis, Daniele Falavigna, Alessio Brutti, and Maja Pantic. 2024.
\newblock \href {https://doi.org/10.48550/ARXIV.2409.12319} {Large language models are strong audio-visual speech recognition learners}.
\newblock \emph{CoRR}, abs/2409.12319.

\bibitem[{Cerspense(2023)}]{zeroscope}
Cerspense. 2023.
\newblock \href {https://huggingface.co/cerspense} {Zeroscope: Diffusion-based text-to-video synthesis}.

\bibitem[{Cha et~al.(2024)Cha, Kang, Mun, and Roh}]{honeybee}
Junbum Cha, Wooyoung Kang, Jonghwan Mun, and Byungseok Roh. 2024.
\newblock \href {https://doi.org/10.1109/CVPR52733.2024.01311} {Honeybee: Locality-enhanced projector for multimodal {LLM}}.
\newblock In \emph{{IEEE/CVF} Conference on Computer Vision and Pattern Recognition, {CVPR} 2024, Seattle, WA, USA, June 16-22, 2024}, pages 13817--13827. {IEEE}.

\bibitem[{Changpinyo et~al.(2021)Changpinyo, Sharma, Ding, and Soricut}]{cc12m}
Soravit Changpinyo, Piyush Sharma, Nan Ding, and Radu Soricut. 2021.
\newblock \href {https://doi.org/10.1109/CVPR46437.2021.00356} {Conceptual 12m: Pushing web-scale image-text pre-training to recognize long-tail visual concepts}.
\newblock In \emph{{IEEE} Conference on Computer Vision and Pattern Recognition, {CVPR} 2021, virtual, June 19-25, 2021}, pages 3558--3568. Computer Vision Foundation / {IEEE}.

\bibitem[{Chen et~al.(2024{\natexlab{a}})Chen, Du, Fang, Wang, Luo, Li, Yan, Zhang, Huang, Sun, and Liu}]{ModelComposition}
Chi Chen, Yiyang Du, Zheng Fang, Ziyue Wang, Fuwen Luo, Peng Li, Ming Yan, Ji~Zhang, Fei Huang, Maosong Sun, and Yang Liu. 2024{\natexlab{a}}.
\newblock \href {https://doi.org/10.18653/V1/2024.ACL-LONG.606} {Model composition for multimodal large language models}.
\newblock In \emph{Proceedings of the 62nd Annual Meeting of the Association for Computational Linguistics (Volume 1: Long Papers), {ACL} 2024, Bangkok, Thailand, August 11-16, 2024}, pages 11246--11262. Association for Computational Linguistics.

\bibitem[{Chen et~al.(2020{\natexlab{a}})Chen, Chang, and Nie{\ss}ner}]{ScanRefer}
Dave~Zhenyu Chen, Angel~X. Chang, and Matthias Nie{\ss}ner. 2020{\natexlab{a}}.
\newblock \href {https://doi.org/10.1007/978-3-030-58565-5\_13} {Scanrefer: 3d object localization in {RGB-D} scans using natural language}.
\newblock In \emph{Computer Vision - {ECCV} 2020 - 16th European Conference, Glasgow, UK, August 23-28, 2020, Proceedings, Part {XX}}, volume 12365 of \emph{Lecture Notes in Computer Science}, pages 202--221. Springer.

\bibitem[{Chen et~al.(2023{\natexlab{a}})Chen, Han, Zhao, Zhang, Shi, Xu, and Xu}]{xllm2023chen}
Feilong Chen, Minglun Han, Haozhi Zhao, Qingyang Zhang, Jing Shi, Shuang Xu, and Bo~Xu. 2023{\natexlab{a}}.
\newblock \href {https://doi.org/10.48550/ARXIV.2305.04160} {{X-LLM:} bootstrapping advanced large language models by treating multi-modalities as foreign languages}.
\newblock \emph{CoRR}, abs/2305.04160.

\bibitem[{Chen et~al.(2021)Chen, Chai, Wang, Du, Zhang, Weng, Su, Povey, Trmal, Zhang, Jin, Khudanpur, Watanabe, Zhao, Zou, Li, Yao, Wang, You, and Yan}]{gigaspeech}
Guoguo Chen, Shuzhou Chai, Guan{-}Bo Wang, Jiayu Du, Wei{-}Qiang Zhang, Chao Weng, Dan Su, Daniel Povey, Jan Trmal, Junbo Zhang, Mingjie Jin, Sanjeev Khudanpur, Shinji Watanabe, Shuaijiang Zhao, Wei Zou, Xiangang Li, Xuchen Yao, Yongqing Wang, Zhao You, and Zhiyong Yan. 2021.
\newblock \href {https://doi.org/10.21437/INTERSPEECH.2021-1965} {Gigaspeech: An evolving, multi-domain {ASR} corpus with 10, 000 hours of transcribed audio}.
\newblock In \emph{22nd Annual Conference of the International Speech Communication Association, Interspeech 2021, Brno, Czechia, August 30 - September 3, 2021}, pages 3670--3674. {ISCA}.

\bibitem[{Chen et~al.(2023{\natexlab{b}})Chen, Xia, He, Zhang, Cun, Yang, Xing, Liu, Chen, Wang, Weng, and Shan}]{VideoCrafter}
Haoxin Chen, Menghan Xia, Yingqing He, Yong Zhang, Xiaodong Cun, Shaoshu Yang, Jinbo Xing, Yaofang Liu, Qifeng Chen, Xintao Wang, Chao Weng, and Ying Shan. 2023{\natexlab{b}}.
\newblock \href {https://doi.org/10.48550/ARXIV.2310.19512} {Videocrafter1: Open diffusion models for high-quality video generation}.
\newblock \emph{CoRR}, abs/2310.19512.

\bibitem[{Chen et~al.(2024{\natexlab{b}})Chen, Wang, Zhou, Huang, Zhang, Feng, Chen, Zhang, Tang, and Zhu}]{survey_h}
Hong Chen, Xin Wang, Yuwei Zhou, Bin Huang, Yipeng Zhang, Wei Feng, Houlun Chen, Zeyang Zhang, Siao Tang, and Wenwu Zhu. 2024{\natexlab{b}}.
\newblock \href {https://doi.org/10.48550/ARXIV.2409.14993} {Multi-modal generative {AI:} multi-modal llm, diffusion and beyond}.
\newblock \emph{CoRR}, abs/2409.14993.

\bibitem[{Chen et~al.(2020{\natexlab{b}})Chen, Xie, Vedaldi, and Zisserman}]{vggsound}
Honglie Chen, Weidi Xie, Andrea Vedaldi, and Andrew Zisserman. 2020{\natexlab{b}}.
\newblock \href {https://doi.org/10.1109/ICASSP40776.2020.9053174} {Vggsound: {A} large-scale audio-visual dataset}.
\newblock In \emph{2020 {IEEE} International Conference on Acoustics, Speech and Signal Processing, {ICASSP} 2020, Barcelona, Spain, May 4-8, 2020}, pages 721--725. {IEEE}.

\bibitem[{Chen et~al.(2023{\natexlab{c}})Chen, Yang, and Zhang}]{sam}
Jiaqi Chen, Zeyu Yang, and Li~Zhang. 2023{\natexlab{c}}.
\newblock Semantic segment anything.
\newblock \url{https://github.com/fudan-zvg/Semantic-Segment-Anything}.

\bibitem[{Chen et~al.(2024{\natexlab{c}})Chen, Gou, Huang, Liu, Tan, Xu, Wang, Zhu, Zeng, Yang, Wang, Xiang, Li, Bai, Han, Li, Jin, Xie, Zhang, Kwok, Zhao, Liang, Yeung, Chen, Li, Zhang, Liu, Yao, Hong, Hou, and Xu}]{emova}
Kai Chen, Yunhao Gou, Runhui Huang, Zhili Liu, Daxin Tan, Jing Xu, Chunwei Wang, Yi~Zhu, Yihan Zeng, Kuo Yang, Dingdong Wang, Kun Xiang, Haoyuan Li, Haoli Bai, Jianhua Han, Xiaohui Li, Weike Jin, Nian Xie, Yu~Zhang, James~T. Kwok, Hengshuang Zhao, Xiaodan Liang, Dit{-}Yan Yeung, Xiao Chen, Zhenguo Li, Wei Zhang, Qun Liu, Jun Yao, Lanqing Hong, Lu~Hou, and Hang Xu. 2024{\natexlab{c}}.
\newblock \href {https://doi.org/10.48550/ARXIV.2409.18042} {{EMOVA:} empowering language models to see, hear and speak with vivid emotions}.
\newblock \emph{CoRR}, abs/2409.18042.

\bibitem[{Chen et~al.(2024{\natexlab{d}})Chen, Hu, Zhang, Chen, Wang, Li, Shyam, Zhou, Huang, Yang, and Gong}]{omnixr}
Lichang Chen, Hexiang Hu, Mingda Zhang, Yiwen Chen, Zifeng Wang, Yandong Li, Pranav Shyam, Tianyi Zhou, Heng Huang, Ming{-}Hsuan Yang, and Boqing Gong. 2024{\natexlab{d}}.
\newblock \href {https://doi.org/10.48550/ARXIV.2410.12219} {Omnixr: Evaluating omni-modality language models on reasoning across modalities}.
\newblock \emph{CoRR}, abs/2410.12219.

\bibitem[{Chen et~al.(2024{\natexlab{e}})Chen, Li, Dong, Zhang, He, Wang, Zhao, and Lin}]{sharegpt4v}
Lin Chen, Jinsong Li, Xiaoyi Dong, Pan Zhang, Conghui He, Jiaqi Wang, Feng Zhao, and Dahua Lin. 2024{\natexlab{e}}.
\newblock \href {https://doi.org/10.1007/978-3-031-72643-9\_22} {Sharegpt4v: Improving large multi-modal models with better captions}.
\newblock In \emph{Computer Vision - {ECCV} 2024 - 18th European Conference, Milan, Italy, September 29-October 4, 2024, Proceedings, Part {XVII}}, volume 15075 of \emph{Lecture Notes in Computer Science}, pages 370--387. Springer.

\bibitem[{Chen et~al.(2023{\natexlab{d}})Chen, Wu, Wang, Liu, Tompkins, Chen, Che, Yu, and Wei}]{Beats}
Sanyuan Chen, Yu~Wu, Chengyi Wang, Shujie Liu, Daniel Tompkins, Zhuo Chen, Wanxiang Che, Xiangzhan Yu, and Furu Wei. 2023{\natexlab{d}}.
\newblock \href {https://proceedings.mlr.press/v202/chen23ag.html} {Beats: Audio pre-training with acoustic tokenizers}.
\newblock In \emph{International Conference on Machine Learning, {ICML} 2023, 23-29 July 2023, Honolulu, Hawaii, {USA}}, volume 202 of \emph{Proceedings of Machine Learning Research}, pages 5178--5193. {PMLR}.

\bibitem[{Chen et~al.(2023{\natexlab{e}})Chen, He, Guo, Zhu, Wang, Tang, and Liu}]{VALOR}
Sihan Chen, Xingjian He, Longteng Guo, Xinxin Zhu, Weining Wang, Jinhui Tang, and Jing Liu. 2023{\natexlab{e}}.
\newblock \href {https://doi.org/10.48550/ARXIV.2304.08345} {{VALOR:} vision-audio-language omni-perception pretraining model and dataset}.
\newblock \emph{CoRR}, abs/2304.08345.

\bibitem[{Chen et~al.(2023{\natexlab{f}})Chen, Li, Wang, Zhao, Sun, Zhu, and Liu}]{vast}
Sihan Chen, Handong Li, Qunbo Wang, Zijia Zhao, Mingzhen Sun, Xinxin Zhu, and Jing Liu. 2023{\natexlab{f}}.
\newblock \href {http://papers.nips.cc/paper\_files/paper/2023/hash/e6b2b48b5ed90d07c305932729927781-Abstract-Conference.html} {{VAST:} {A} vision-audio-subtitle-text omni-modality foundation model and dataset}.
\newblock In \emph{Advances in Neural Information Processing Systems 36: Annual Conference on Neural Information Processing Systems 2023, NeurIPS 2023, New Orleans, LA, USA, December 10 - 16, 2023}.

\bibitem[{Chen et~al.(2024{\natexlab{f}})Chen, Siarohin, Menapace, Deyneka, Chao, Jeon, Fang, Lee, Ren, Yang, and Tulyakov}]{panda-70m}
Tsai{-}Shien Chen, Aliaksandr Siarohin, Willi Menapace, Ekaterina Deyneka, Hsiang{-}wei Chao, Byung~Eun Jeon, Yuwei Fang, Hsin{-}Ying Lee, Jian Ren, Ming{-}Hsuan Yang, and Sergey Tulyakov. 2024{\natexlab{f}}.
\newblock \href {https://doi.org/10.1109/CVPR52733.2024.01265} {Panda-70m: Captioning 70m videos with multiple cross-modality teachers}.
\newblock In \emph{{IEEE/CVF} Conference on Computer Vision and Pattern Recognition, {CVPR} 2024, Seattle, WA, USA, June 16-22, 2024}, pages 13320--13331. {IEEE}.

\bibitem[{Chen et~al.(2024{\natexlab{g}})Chen, Wang, Wang, Liu, Yin, Liu, Sheng, Ouyang, and Shao}]{Octavius}
Zeren Chen, Ziqin Wang, Zhen Wang, Huayang Liu, Zhenfei Yin, Si~Liu, Lu~Sheng, Wanli Ouyang, and Jing Shao. 2024{\natexlab{g}}.
\newblock \href {https://openreview.net/forum?id=rTDyN8yajn} {Octavius: Mitigating task interference in mllms via lora-moe}.
\newblock In \emph{The Twelfth International Conference on Learning Representations, {ICLR} 2024, Vienna, Austria, May 7-11, 2024}. OpenReview.net.

\bibitem[{Chen et~al.(2023{\natexlab{g}})Chen, Wu, Wang, Su, Chen, Xing, Zhong, Zhang, Zhu, Lu, Li, Luo, Lu, Qiao, and Dai}]{internvl}
Zhe Chen, Jiannan Wu, Wenhai Wang, Weijie Su, Guo Chen, Sen Xing, Muyan Zhong, Qinglong Zhang, Xizhou Zhu, Lewei Lu, Bin Li, Ping Luo, Tong Lu, Yu~Qiao, and Jifeng Dai. 2023{\natexlab{g}}.
\newblock \href {https://doi.org/10.48550/ARXIV.2312.14238} {Internvl: Scaling up vision foundation models and aligning for generic visual-linguistic tasks}.
\newblock \emph{CoRR}, abs/2312.14238.

\bibitem[{Cheng et~al.(2024{\natexlab{a}})Cheng, Cheng, He, Sun, Wang, Lin, Lian, Peng, and Hauptmann}]{emotionllama}
Zebang Cheng, Zhi{-}Qi Cheng, Jun{-}Yan He, Jingdong Sun, Kai Wang, Yuxiang Lin, Zheng Lian, Xiaojiang Peng, and Alexander Hauptmann. 2024{\natexlab{a}}.
\newblock \href {https://doi.org/10.48550/ARXIV.2406.11161} {Emotion-llama: Multimodal emotion recognition and reasoning with instruction tuning}.
\newblock \emph{CoRR}, abs/2406.11161.

\bibitem[{Cheng et~al.(2024{\natexlab{b}})Cheng, Leng, Zhang, Xin, Li, Chen, Zhu, Zhang, Luo, Zhao, and Bing}]{videollama2}
Zesen Cheng, Sicong Leng, Hang Zhang, Yifei Xin, Xin Li, Guanzheng Chen, Yongxin Zhu, Wenqi Zhang, Ziyang Luo, Deli Zhao, and Lidong Bing. 2024{\natexlab{b}}.
\newblock \href {https://doi.org/10.48550/ARXIV.2406.07476} {Videollama 2: Advancing spatial-temporal modeling and audio understanding in video-llms}.
\newblock \emph{CoRR}, abs/2406.07476.

\bibitem[{Chng and Chan(2017)}]{TotalText}
Chee~Kheng Chng and Chee~Seng Chan. 2017.
\newblock \href {https://doi.org/10.1109/ICDAR.2017.157} {Total-text: {A} comprehensive dataset for scene text detection and recognition}.
\newblock In \emph{14th {IAPR} International Conference on Document Analysis and Recognition, {ICDAR} 2017, Kyoto, Japan, November 9-15, 2017}, pages 935--942. {IEEE}.

\bibitem[{Chowdhury et~al.(2024)Chowdhury, Nag, Dasgupta, Chen, Elhoseiny, Gao, and Manocha}]{meerkat}
Sanjoy Chowdhury, Sayan Nag, Subhrajyoti Dasgupta, Jun Chen, Mohamed Elhoseiny, Ruohan Gao, and Dinesh Manocha. 2024.
\newblock \href {https://doi.org/10.1007/978-3-031-73039-9\_4} {{MEERKAT:} audio-visual large language model for grounding in space and time}.
\newblock In \emph{Computer Vision - {ECCV} 2024 - 18th European Conference, Milan, Italy, September 29-October 4, 2024, Proceedings, Part {LXIV}}, volume 15122 of \emph{Lecture Notes in Computer Science}, pages 52--70. Springer.

\bibitem[{Chu et~al.(2023)Chu, Xu, Zhou, Yang, Zhang, Yan, Zhou, and Zhou}]{qwenaudio}
Yunfei Chu, Jin Xu, Xiaohuan Zhou, Qian Yang, Shiliang Zhang, Zhijie Yan, Chang Zhou, and Jingren Zhou. 2023.
\newblock \href {https://doi.org/10.48550/ARXIV.2311.07919} {Qwen-audio: Advancing universal audio understanding via unified large-scale audio-language models}.
\newblock \emph{CoRR}, abs/2311.07919.

\bibitem[{{\c{C}}oban et~al.(2024){\c{C}}oban, Mandel, and Devaney}]{survey_audio}
Enis~Berk {\c{C}}oban, Michael~I. Mandel, and Johanna Devaney. 2024.
\newblock \href {https://doi.org/10.48550/ARXIV.2406.04615} {What do mllms hear? examining reasoning with text and sound components in multimodal large language models}.
\newblock \emph{CoRR}, abs/2406.04615.

\bibitem[{Copet et~al.(2023)Copet, Kreuk, Gat, Remez, Kant, Synnaeve, Adi, and D{\'{e}}fossez}]{musicgen}
Jade Copet, Felix Kreuk, Itai Gat, Tal Remez, David Kant, Gabriel Synnaeve, Yossi Adi, and Alexandre D{\'{e}}fossez. 2023.
\newblock \href {http://papers.nips.cc/paper\_files/paper/2023/hash/94b472a1842cd7c56dcb125fb2765fbd-Abstract-Conference.html} {Simple and controllable music generation}.
\newblock In \emph{Advances in Neural Information Processing Systems 36: Annual Conference on Neural Information Processing Systems 2023, NeurIPS 2023, New Orleans, LA, USA, December 10 - 16, 2023}.

\bibitem[{Cui et~al.(2024)Cui, Ma, Cao, Ye, Zhou, Liang, Chen, Lu, Yang, Liao, Gao, Li, Tang, Cao, Zhou, Liu, Yan, Mei, Cao, Wang, and Zheng}]{survey_driving_1}
Can Cui, Yunsheng Ma, Xu~Cao, Wenqian Ye, Yang Zhou, Kaizhao Liang, Jintai Chen, Juanwu Lu, Zichong Yang, Kuei{-}Da Liao, Tianren Gao, Erlong Li, Kun Tang, Zhipeng Cao, Tong Zhou, Ao~Liu, Xinrui Yan, Shuqi Mei, Jianguo Cao, Ziran Wang, and Chao Zheng. 2024.
\newblock \href {https://doi.org/10.1109/WACVW60836.2024.00106} {A survey on multimodal large language models for autonomous driving}.
\newblock In \emph{{IEEE/CVF} Winter Conference on Applications of Computer Vision Workshops, {WACVW} 2024 - Workshops, Waikoloa, HI, USA, January 1-6, 2024}, pages 958--979. {IEEE}.

\bibitem[{Das et~al.(2017)Das, Kottur, Gupta, Singh, Yadav, Moura, Parikh, and Batra}]{VisualDialog}
Abhishek Das, Satwik Kottur, Khushi Gupta, Avi Singh, Deshraj Yadav, Jos{\'{e}} M.~F. Moura, Devi Parikh, and Dhruv Batra. 2017.
\newblock \href {https://doi.org/10.1109/CVPR.2017.121} {Visual dialog}.
\newblock In \emph{2017 {IEEE} Conference on Computer Vision and Pattern Recognition, {CVPR} 2017, Honolulu, HI, USA, July 21-26, 2017}, pages 1080--1089. {IEEE} Computer Society.

\bibitem[{D{\'{e}}fossez et~al.(2023)D{\'{e}}fossez, Copet, Synnaeve, and Adi}]{encodec}
Alexandre D{\'{e}}fossez, Jade Copet, Gabriel Synnaeve, and Yossi Adi. 2023.
\newblock \href {https://openreview.net/forum?id=ivCd8z8zR2} {High fidelity neural audio compression}.
\newblock \emph{Trans. Mach. Learn. Res.}, 2023.

\bibitem[{Deitke et~al.(2023)Deitke, Schwenk, Salvador, Weihs, Michel, VanderBilt, Schmidt, Ehsani, Kembhavi, and Farhadi}]{objaverse}
Matt Deitke, Dustin Schwenk, Jordi Salvador, Luca Weihs, Oscar Michel, Eli VanderBilt, Ludwig Schmidt, Kiana Ehsani, Aniruddha Kembhavi, and Ali Farhadi. 2023.
\newblock \href {https://doi.org/10.1109/CVPR52729.2023.01263} {Objaverse: {A} universe of annotated 3d objects}.
\newblock In \emph{{IEEE/CVF} Conference on Computer Vision and Pattern Recognition, {CVPR} 2023, Vancouver, BC, Canada, June 17-24, 2023}, pages 13142--13153. {IEEE}.

\bibitem[{Desai et~al.(2021)Desai, Kaul, Aysola, and Johnson}]{RedCaps}
Karan Desai, Gaurav Kaul, Zubin Aysola, and Justin Johnson. 2021.
\newblock \href {https://datasets-benchmarks-proceedings.neurips.cc/paper/2021/hash/e00da03b685a0dd18fb6a08af0923de0-Abstract-round1.html} {Redcaps: Web-curated image-text data created by the people, for the people}.
\newblock In \emph{Proceedings of the Neural Information Processing Systems Track on Datasets and Benchmarks 1, NeurIPS Datasets and Benchmarks 2021, December 2021, virtual}.

\bibitem[{Dhariwal et~al.(2020)Dhariwal, Jun, Payne, Kim, Radford, and Sutskever}]{jukebox}
Prafulla Dhariwal, Heewoo Jun, Christine Payne, Jong~Wook Kim, Alec Radford, and Ilya Sutskever. 2020.
\newblock \href {https://arxiv.org/abs/2005.00341} {Jukebox: {A} generative model for music}.
\newblock \emph{CoRR}, abs/2005.00341.

\bibitem[{Dosovitskiy et~al.(2021)Dosovitskiy, Beyer, Kolesnikov, Weissenborn, Zhai, Unterthiner, Dehghani, Minderer, Heigold, Gelly, Uszkoreit, and Houlsby}]{vit}
Alexey Dosovitskiy, Lucas Beyer, Alexander Kolesnikov, Dirk Weissenborn, Xiaohua Zhai, Thomas Unterthiner, Mostafa Dehghani, Matthias Minderer, Georg Heigold, Sylvain Gelly, Jakob Uszkoreit, and Neil Houlsby. 2021.
\newblock \href {https://openreview.net/forum?id=YicbFdNTTy} {An image is worth 16x16 words: Transformers for image recognition at scale}.
\newblock In \emph{9th International Conference on Learning Representations, {ICLR} 2021, Virtual Event, Austria, May 3-7, 2021}. OpenReview.net.

\bibitem[{Drossos et~al.(2020)Drossos, Lipping, and Virtanen}]{clotho}
Konstantinos Drossos, Samuel Lipping, and Tuomas Virtanen. 2020.
\newblock \href {https://doi.org/10.1109/ICASSP40776.2020.9052990} {Clotho: an audio captioning dataset}.
\newblock In \emph{2020 {IEEE} International Conference on Acoustics, Speech and Signal Processing, {ICASSP} 2020, Barcelona, Spain, May 4-8, 2020}, pages 736--740. {IEEE}.

\bibitem[{Du et~al.(2018)Du, Na, Liu, and Bu}]{aishell-2}
Jiayu Du, Xingyu Na, Xuechen Liu, and Hui Bu. 2018.
\newblock \href {https://arxiv.org/abs/1808.10583} {{AISHELL-2:} transforming mandarin {ASR} research into industrial scale}.
\newblock \emph{CoRR}, abs/1808.10583.

\bibitem[{Elizalde et~al.(2023)Elizalde, Deshmukh, Ismail, and Wang}]{CLAP}
Benjamin Elizalde, Soham Deshmukh, Mahmoud~Al Ismail, and Huaming Wang. 2023.
\newblock \href {https://doi.org/10.1109/ICASSP49357.2023.10095889} {{CLAP} learning audio concepts from natural language supervision}.
\newblock In \emph{{IEEE} International Conference on Acoustics, Speech and Signal Processing {ICASSP} 2023, Rhodes Island, Greece, June 4-10, 2023}, pages 1--5. {IEEE}.

\bibitem[{Esser et~al.(2021)Esser, Rombach, and Ommer}]{vqgan}
Patrick Esser, Robin Rombach, and Bj{\"{o}}rn Ommer. 2021.
\newblock \href {https://doi.org/10.1109/CVPR46437.2021.01268} {Taming transformers for high-resolution image synthesis}.
\newblock In \emph{{IEEE} Conference on Computer Vision and Pattern Recognition, {CVPR} 2021, virtual, June 19-25, 2021}, pages 12873--12883. Computer Vision Foundation / {IEEE}.

\bibitem[{Fan et~al.(2024)Fan, Cao, Zhao, Liu, and Li}]{survey_safety_1}
Yihe Fan, Yuxin Cao, Ziyu Zhao, Ziyao Liu, and Shaofeng Li. 2024.
\newblock \href {https://doi.org/10.48550/ARXIV.2404.05264} {Unbridled icarus: {A} survey of the potential perils of image inputs in multimodal large language model security}.
\newblock \emph{CoRR}, abs/2404.05264.

\bibitem[{Fang et~al.(2024{\natexlab{a}})Fang, Jose, Jain, Schmidt, Toshev, and Shankar}]{dfnclip}
Alex Fang, Albin~Madappally Jose, Amit Jain, Ludwig Schmidt, Alexander~T. Toshev, and Vaishaal Shankar. 2024{\natexlab{a}}.
\newblock \href {https://openreview.net/forum?id=KAk6ngZ09F} {Data filtering networks}.
\newblock In \emph{The Twelfth International Conference on Learning Representations, {ICLR} 2024, Vienna, Austria, May 7-11, 2024}. OpenReview.net.

\bibitem[{Fang et~al.(2024{\natexlab{b}})Fang, Guo, Zhou, Ma, Zhang, and Feng}]{llama-omni}
Qingkai Fang, Shoutao Guo, Yan Zhou, Zhengrui Ma, Shaolei Zhang, and Yang Feng. 2024{\natexlab{b}}.
\newblock \href {https://doi.org/10.48550/ARXIV.2409.06666} {Llama-omni: Seamless speech interaction with large language models}.
\newblock \emph{CoRR}, abs/2409.06666.

\bibitem[{Fei et~al.(2024)Fei, Zhang, Wang, Liao, Liu, and Cambria}]{EmpathyEar}
Hao Fei, Han Zhang, Bin Wang, Lizi Liao, Qian Liu, and Erik Cambria. 2024.
\newblock \href {https://doi.org/10.48550/ARXIV.2406.15177} {Empathyear: An open-source avatar multimodal empathetic chatbot}.
\newblock \emph{CoRR}, abs/2406.15177.

\bibitem[{Fu et~al.(2023)Fu, Chen, Shen, Qin, Zhang, Lin, Qiu, Lin, Yang, Zheng, Li, Sun, and Ji}]{MME}
Chaoyou Fu, Peixian Chen, Yunhang Shen, Yulei Qin, Mengdan Zhang, Xu~Lin, Zhenyu Qiu, Wei Lin, Jinrui Yang, Xiawu Zheng, Ke~Li, Xing Sun, and Rongrong Ji. 2023.
\newblock \href {https://doi.org/10.48550/ARXIV.2306.13394} {{MME:} {A} comprehensive evaluation benchmark for multimodal large language models}.
\newblock \emph{CoRR}, abs/2306.13394.

\bibitem[{Fu et~al.(2024{\natexlab{a}})Fu, Dai, Luo, Li, Ren, Zhang, Wang, Zhou, Shen, Zhang, Chen, Li, Lin, Zhao, Li, Xu, Zheng, Chen, Ji, and Sun}]{videomme}
Chaoyou Fu, Yuhan Dai, Yondong Luo, Lei Li, Shuhuai Ren, Renrui Zhang, Zihan Wang, Chenyu Zhou, Yunhang Shen, Mengdan Zhang, Peixian Chen, Yanwei Li, Shaohui Lin, Sirui Zhao, Ke~Li, Tong Xu, Xiawu Zheng, Enhong Chen, Rongrong Ji, and Xing Sun. 2024{\natexlab{a}}.
\newblock \href {https://doi.org/10.48550/ARXIV.2405.21075} {Video-mme: The first-ever comprehensive evaluation benchmark of multi-modal llms in video analysis}.
\newblock \emph{CoRR}, abs/2405.21075.

\bibitem[{Fu et~al.(2024{\natexlab{b}})Fu, Lin, Long, Shen, Zhao, Zhang, Wang, Yin, Ma, Zheng, He, Ji, Wu, Shan, and Sun}]{vita}
Chaoyou Fu, Haojia Lin, Zuwei Long, Yunhang Shen, Meng Zhao, Yifan Zhang, Xiong Wang, Di~Yin, Long Ma, Xiawu Zheng, Ran He, Rongrong Ji, Yunsheng Wu, Caifeng Shan, and Xing Sun. 2024{\natexlab{b}}.
\newblock \href {https://doi.org/10.48550/ARXIV.2408.05211} {{VITA:} towards open-source interactive omni multimodal {LLM}}.
\newblock \emph{CoRR}, abs/2408.05211.

\bibitem[{Fu et~al.(2025)Fu, Lin, Wang, Zhang, Shen, Liu, Li, Long, Gao, Li, Ma, Zheng, Ji, Sun, Shan, and He}]{vita-1.5}
Chaoyou Fu, Haojia Lin, Xiong Wang, Yi-Fan Zhang, Yunhang Shen, Xiaoyu Liu, Yangze Li, Zuwei Long, Heting Gao, Ke~Li, Long Ma, Xiawu Zheng, Rongrong Ji, Xing Sun, Caifeng Shan, and Ran He. 2025.
\newblock \href {https://arxiv.org/abs/2501.01957} {Vita-1.5: Towards gpt-4o level real-time vision and speech interaction}.
\newblock \emph{Preprint}, arXiv:2501.01957.

\bibitem[{Fu et~al.(2024{\natexlab{c}})Fu, Datta, Huang, Panitch, Drake, Ortiz, Mukadam, Lambeta, Calandra, and Goldberg}]{TVL}
Letian Fu, Gaurav Datta, Huang Huang, William~Chung{-}Ho Panitch, Jaimyn Drake, Joseph Ortiz, Mustafa Mukadam, Mike Lambeta, Roberto Calandra, and Ken Goldberg. 2024{\natexlab{c}}.
\newblock \href {https://openreview.net/forum?id=tFEOOH9eH0} {A touch, vision, and language dataset for multimodal alignment}.
\newblock In \emph{Forty-first International Conference on Machine Learning, {ICML} 2024, Vienna, Austria, July 21-27, 2024}. OpenReview.net.

\bibitem[{Ge et~al.(2023)Ge, Ge, Zeng, Wang, and Shan}]{SEED}
Yuying Ge, Yixiao Ge, Ziyun Zeng, Xintao Wang, and Ying Shan. 2023.
\newblock \href {https://doi.org/10.48550/ARXIV.2307.08041} {Planting a {SEED} of vision in large language model}.
\newblock \emph{CoRR}, abs/2307.08041.

\bibitem[{Ge et~al.(2024{\natexlab{a}})Ge, Zhao, Zeng, Ge, Li, Wang, and Shan}]{seed_tokenizer}
Yuying Ge, Sijie Zhao, Ziyun Zeng, Yixiao Ge, Chen Li, Xintao Wang, and Ying Shan. 2024{\natexlab{a}}.
\newblock \href {https://openreview.net/forum?id=0Nui91LBQS} {Making llama {SEE} and draw with {SEED} tokenizer}.
\newblock In \emph{The Twelfth International Conference on Learning Representations, {ICLR} 2024, Vienna, Austria, May 7-11, 2024}. OpenReview.net.

\bibitem[{Ge et~al.(2024{\natexlab{b}})Ge, Huang, Zhou, Li, Wang, Tang, and Zhuang}]{worldgpt}
Zhiqi Ge, Hongzhe Huang, Mingze Zhou, Juncheng Li, Guoming Wang, Siliang Tang, and Yueting Zhuang. 2024{\natexlab{b}}.
\newblock \href {https://doi.org/10.1145/3664647.3681488} {Worldgpt: Empowering {LLM} as multimodal world model}.
\newblock In \emph{Proceedings of the 32nd {ACM} International Conference on Multimedia, {MM} 2024, Melbourne, VIC, Australia, 28 October 2024 - 1 November 2024}, pages 7346--7355. {ACM}.

\bibitem[{Gemmeke et~al.(2017)Gemmeke, Ellis, Freedman, Jansen, Lawrence, Moore, Plakal, and Ritter}]{audioset}
Jort~F. Gemmeke, Daniel P.~W. Ellis, Dylan Freedman, Aren Jansen, Wade Lawrence, R.~Channing Moore, Manoj Plakal, and Marvin Ritter. 2017.
\newblock \href {https://doi.org/10.1109/ICASSP.2017.7952261} {Audio set: An ontology and human-labeled dataset for audio events}.
\newblock In \emph{2017 {IEEE} International Conference on Acoustics, Speech and Signal Processing, {ICASSP} 2017, New Orleans, LA, USA, March 5-9, 2017}, pages 776--780. {IEEE}.

\bibitem[{Geng et~al.(2023)Geng, Wang, Duan, Cong, and Zheng}]{unav-100}
Tiantian Geng, Teng Wang, Jinming Duan, Runmin Cong, and Feng Zheng. 2023.
\newblock \href {https://doi.org/10.1109/CVPR52729.2023.02197} {Dense-localizing audio-visual events in untrimmed videos: {A} large-scale benchmark and baseline}.
\newblock In \emph{{IEEE/CVF} Conference on Computer Vision and Pattern Recognition, {CVPR} 2023, Vancouver, BC, Canada, June 17-24, 2023}, pages 22942--22951. {IEEE}.

\bibitem[{Geng et~al.(2024)Geng, Zhang, Wang, Wang, Duan, and Zheng}]{LongVALE-LLM}
Tiantian Geng, Jinrui Zhang, Qingni Wang, Teng Wang, Jinming Duan, and Feng Zheng. 2024.
\newblock \href {https://doi.org/10.48550/ARXIV.2411.19772} {Longvale: Vision-audio-language-event benchmark towards time-aware omni-modal perception of long videos}.
\newblock \emph{CoRR}, abs/2411.19772.

\bibitem[{Girdhar et~al.(2023)Girdhar, El{-}Nouby, Liu, Singh, Alwala, Joulin, and Misra}]{imagebind}
Rohit Girdhar, Alaaeldin El{-}Nouby, Zhuang Liu, Mannat Singh, Kalyan~Vasudev Alwala, Armand Joulin, and Ishan Misra. 2023.
\newblock \href {https://doi.org/10.1109/CVPR52729.2023.01457} {Imagebind one embedding space to bind them all}.
\newblock In \emph{{IEEE/CVF} Conference on Computer Vision and Pattern Recognition, {CVPR} 2023, Vancouver, BC, Canada, June 17-24, 2023}, pages 15180--15190. {IEEE}.

\bibitem[{Goel et~al.(2024)Goel, Sapra, Le, Valle, Tao, and Catanzaro}]{omcat}
Arushi Goel, Karan Sapra, Matthieu Le, Rafael Valle, Andrew Tao, and Bryan Catanzaro. 2024.
\newblock \href {https://doi.org/10.48550/ARXIV.2410.12109} {{OMCAT:} omni context aware transformer}.
\newblock \emph{CoRR}, abs/2410.12109.

\bibitem[{Gong et~al.(2021)Gong, Chung, and Glass}]{AST}
Yuan Gong, Yu{-}An Chung, and James~R. Glass. 2021.
\newblock \href {https://doi.org/10.21437/INTERSPEECH.2021-698} {{AST:} audio spectrogram transformer}.
\newblock In \emph{22nd Annual Conference of the International Speech Communication Association, Interspeech 2021, Brno, Czechia, August 30 - September 3, 2021}, pages 571--575. {ISCA}.

\bibitem[{Gong et~al.(2022)Gong, Yu, and Glass}]{vocalsound}
Yuan Gong, Jin Yu, and James~R. Glass. 2022.
\newblock \href {https://doi.org/10.1109/ICASSP43922.2022.9746828} {Vocalsound: {A} dataset for improving human vocal sounds recognition}.
\newblock In \emph{{IEEE} International Conference on Acoustics, Speech and Signal Processing, {ICASSP} 2022, Virtual and Singapore, 23-27 May 2022}, pages 151--155. {IEEE}.

\bibitem[{Goyal et~al.(2017)Goyal, Khot, Summers{-}Stay, Batra, and Parikh}]{VQAv2}
Yash Goyal, Tejas Khot, Douglas Summers{-}Stay, Dhruv Batra, and Devi Parikh. 2017.
\newblock \href {https://doi.org/10.1109/CVPR.2017.670} {Making the {V} in {VQA} matter: Elevating the role of image understanding in visual question answering}.
\newblock In \emph{2017 {IEEE} Conference on Computer Vision and Pattern Recognition, {CVPR} 2017, Honolulu, HI, USA, July 21-26, 2017}, pages 6325--6334. {IEEE} Computer Society.

\bibitem[{Grauman et~al.(2022)Grauman, Westbury, Byrne, Chavis, Furnari, Girdhar, Hamburger, Jiang, Liu, Liu, Martin, Nagarajan, Radosavovic, Ramakrishnan, Ryan, Sharma, Wray, Xu, Xu, Zhao, Bansal, Batra, Cartillier, Crane, Do, Doulaty, Erapalli, Feichtenhofer, Fragomeni, Fu, Gebreselasie, Gonz{\'{a}}lez, Hillis, Huang, Huang, Jia, Khoo, Kol{\'{a}}r, Kottur, Kumar, Landini, Li, Li, Li, Mangalam, Modhugu, Munro, Murrell, Nishiyasu, Price, Puentes, Ramazanova, Sari, Somasundaram, Southerland, Sugano, Tao, Vo, Wang, Wu, Yagi, Zhao, Zhu, Arbel{\'{a}}ez, Crandall, Damen, Farinella, Fuegen, Ghanem, Ithapu, Jawahar, Joo, Kitani, Li, Newcombe, Oliva, Park, Rehg, Sato, Shi, Shou, Torralba, Torresani, Yan, and Malik}]{ego4d}
Kristen Grauman, Andrew Westbury, Eugene Byrne, Zachary Chavis, Antonino Furnari, Rohit Girdhar, Jackson Hamburger, Hao Jiang, Miao Liu, Xingyu Liu, Miguel Martin, Tushar Nagarajan, Ilija Radosavovic, Santhosh~Kumar Ramakrishnan, Fiona Ryan, Jayant Sharma, Michael Wray, Mengmeng Xu, Eric~Zhongcong Xu, Chen Zhao, Siddhant Bansal, Dhruv Batra, Vincent Cartillier, Sean Crane, Tien Do, Morrie Doulaty, Akshay Erapalli, Christoph Feichtenhofer, Adriano Fragomeni, Qichen Fu, Abrham Gebreselasie, Cristina Gonz{\'{a}}lez, James Hillis, Xuhua Huang, Yifei Huang, Wenqi Jia, Weslie Khoo, J{\'{a}}chym Kol{\'{a}}r, Satwik Kottur, Anurag Kumar, Federico Landini, Chao Li, Yanghao Li, Zhenqiang Li, Karttikeya Mangalam, Raghava Modhugu, Jonathan Munro, Tullie Murrell, Takumi Nishiyasu, Will Price, Paola~Ruiz Puentes, Merey Ramazanova, Leda Sari, Kiran Somasundaram, Audrey Southerland, Yusuke Sugano, Ruijie Tao, Minh Vo, Yuchen Wang, Xindi Wu, Takuma Yagi, Ziwei Zhao, Yunyi Zhu, Pablo Arbel{\'{a}}ez, David Crandall, Dima Damen,
  Giovanni~Maria Farinella, Christian Fuegen, Bernard Ghanem, Vamsi~Krishna Ithapu, C.~V. Jawahar, Hanbyul Joo, Kris Kitani, Haizhou Li, Richard~A. Newcombe, Aude Oliva, Hyun~Soo Park, James~M. Rehg, Yoichi Sato, Jianbo Shi, Mike~Zheng Shou, Antonio Torralba, Lorenzo Torresani, Mingfei Yan, and Jitendra Malik. 2022.
\newblock \href {https://doi.org/10.1109/CVPR52688.2022.01842} {Ego4d: Around the world in 3, 000 hours of egocentric video}.
\newblock In \emph{{IEEE/CVF} Conference on Computer Vision and Pattern Recognition, {CVPR} 2022, New Orleans, LA, USA, June 18-24, 2022}, pages 18973--18990. {IEEE}.

\bibitem[{Gu et~al.(2022)Gu, Meng, Lu, Hou, Minzhe, Liang, Yao, Huang, Zhang, Jiang, Xu, and Xu}]{wukong}
Jiaxi Gu, Xiaojun Meng, Guansong Lu, Lu~Hou, Niu Minzhe, Xiaodan Liang, Lewei Yao, Runhui Huang, Wei Zhang, Xin Jiang, Chunjing Xu, and Hang Xu. 2022.
\newblock \href {http://papers.nips.cc/paper\_files/paper/2022/hash/a90b9a09a6ee43d6631cf42e225d73b4-Abstract-Datasets\_and\_Benchmarks.html} {Wukong: {A} 100 million large-scale chinese cross-modal pre-training benchmark}.
\newblock In \emph{Advances in Neural Information Processing Systems 35: Annual Conference on Neural Information Processing Systems 2022, NeurIPS 2022, New Orleans, LA, USA, November 28 - December 9, 2022}.

\bibitem[{Gulati et~al.(2020)Gulati, Qin, Chiu, Parmar, Zhang, Yu, Han, Wang, Zhang, Wu, and Pang}]{Conformer}
Anmol Gulati, James Qin, Chung{-}Cheng Chiu, Niki Parmar, Yu~Zhang, Jiahui Yu, Wei Han, Shibo Wang, Zhengdong Zhang, Yonghui Wu, and Ruoming Pang. 2020.
\newblock \href {https://doi.org/10.21437/INTERSPEECH.2020-3015} {Conformer: Convolution-augmented transformer for speech recognition}.
\newblock In \emph{21st Annual Conference of the International Speech Communication Association, Interspeech 2020, Virtual Event, Shanghai, China, October 25-29, 2020}, pages 5036--5040. {ISCA}.

\bibitem[{Gunasekar et~al.(2023)Gunasekar, Zhang, Aneja, Mendes, Giorno, Gopi, Javaheripi, Kauffmann, de~Rosa, Saarikivi, Salim, Shah, Behl, Wang, Bubeck, Eldan, Kalai, Lee, and Li}]{phi}
Suriya Gunasekar, Yi~Zhang, Jyoti Aneja, Caio C{\'{e}}sar~Teodoro Mendes, Allie~Del Giorno, Sivakanth Gopi, Mojan Javaheripi, Piero Kauffmann, Gustavo de~Rosa, Olli Saarikivi, Adil Salim, Shital Shah, Harkirat~Singh Behl, Xin Wang, S{\'{e}}bastien Bubeck, Ronen Eldan, Adam~Tauman Kalai, Yin~Tat Lee, and Yuanzhi Li. 2023.
\newblock \href {https://doi.org/10.48550/ARXIV.2306.11644} {Textbooks are all you need}.
\newblock \emph{CoRR}, abs/2306.11644.

\bibitem[{Gurari et~al.(2018)Gurari, Li, Stangl, Guo, Lin, Grauman, Luo, and Bigham}]{vizwiz}
Danna Gurari, Qing Li, Abigale~J. Stangl, Anhong Guo, Chi Lin, Kristen Grauman, Jiebo Luo, and Jeffrey~P. Bigham. 2018.
\newblock \href {https://doi.org/10.1109/CVPR.2018.00380} {Vizwiz grand challenge: Answering visual questions from blind people}.
\newblock In \emph{2018 {IEEE} Conference on Computer Vision and Pattern Recognition, {CVPR} 2018, Salt Lake City, UT, USA, June 18-22, 2018}, pages 3608--3617. Computer Vision Foundation / {IEEE} Computer Society.

\bibitem[{Han et~al.(2024{\natexlab{a}})Han, Gong, Zhang, Wang, Zhang, Lin, Qiao, Gao, and Yue}]{onellm}
Jiaming Han, Kaixiong Gong, Yiyuan Zhang, Jiaqi Wang, Kaipeng Zhang, Dahua Lin, Yu~Qiao, Peng Gao, and Xiangyu Yue. 2024{\natexlab{a}}.
\newblock \href {https://doi.org/10.1109/CVPR52733.2024.02510} {Onellm: One framework to align all modalities with language}.
\newblock In \emph{{IEEE/CVF} Conference on Computer Vision and Pattern Recognition, {CVPR} 2024, Seattle, WA, USA, June 16-22, 2024}, pages 26574--26585. {IEEE}.

\bibitem[{Han et~al.(2023)Han, Zhang, Shao, Gao, Xu, Xiao, Zhang, Liu, Wen, Guo, Lu, Ren, Wen, Chen, Yue, Li, and Qiao}]{imagebindllm}
Jiaming Han, Renrui Zhang, Wenqi Shao, Peng Gao, Peng Xu, Han Xiao, Kaipeng Zhang, Chris Liu, Song Wen, Ziyu Guo, Xudong Lu, Shuai Ren, Yafei Wen, Xiaoxin Chen, Xiangyu Yue, Hongsheng Li, and Yu~Qiao. 2023.
\newblock \href {https://doi.org/10.48550/ARXIV.2309.03905} {Imagebind-llm: Multi-modality instruction tuning}.
\newblock \emph{CoRR}, abs/2309.03905.

\bibitem[{Han et~al.(2024{\natexlab{b}})Han, Gao, Liu, Zhang, and Zhang}]{PEFT_Survey}
Zeyu Han, Chao Gao, Jinyang Liu, Jeff Zhang, and Sai~Qian Zhang. 2024{\natexlab{b}}.
\newblock \href {https://doi.org/10.48550/ARXIV.2403.14608} {Parameter-efficient fine-tuning for large models: {A} comprehensive survey}.
\newblock \emph{CoRR}, abs/2403.14608.

\bibitem[{He et~al.(2024)He, Liu, Chen, Tian, Liu, Chi, Liu, Yuan, Xing, Wang, Dai, Zhang, Xue, Liu, Guo, and Chen}]{survey_g}
Yingqing He, Zhaoyang Liu, Jingye Chen, Zeyue Tian, Hongyu Liu, Xiaowei Chi, Runtao Liu, Ruibin Yuan, Yazhou Xing, Wenhai Wang, Jifeng Dai, Yong Zhang, Wei Xue, Qifeng Liu, Yike Guo, and Qifeng Chen. 2024.
\newblock \href {https://doi.org/10.48550/ARXIV.2405.19334} {Llms meet multimodal generation and editing: {A} survey}.
\newblock \emph{CoRR}, abs/2405.19334.

\bibitem[{Hong et~al.(2024)Hong, Zheng, Chen, Wang, Li, and Gan}]{MultiPLY}
Yining Hong, Zishuo Zheng, Peihao Chen, Yian Wang, Junyan Li, and Chuang Gan. 2024.
\newblock \href {https://doi.org/10.1109/CVPR52733.2024.02494} {Multiply: {A} multisensory object-centric embodied large language model in 3d world}.
\newblock In \emph{{IEEE/CVF} Conference on Computer Vision and Pattern Recognition, {CVPR} 2024, Seattle, WA, USA, June 16-22, 2024}, pages 26396--26406. {IEEE}.

\bibitem[{Hsu et~al.(2021)Hsu, Bolte, Tsai, Lakhotia, Salakhutdinov, and Mohamed}]{hubert}
Wei{-}Ning Hsu, Benjamin Bolte, Yao{-}Hung~Hubert Tsai, Kushal Lakhotia, Ruslan Salakhutdinov, and Abdelrahman Mohamed. 2021.
\newblock \href {https://doi.org/10.1109/TASLP.2021.3122291} {Hubert: Self-supervised speech representation learning by masked prediction of hidden units}.
\newblock \emph{{IEEE} {ACM} Trans. Audio Speech Lang. Process.}, 29:3451--3460.

\bibitem[{Hu et~al.(2022)Hu, Shen, Wallis, Allen{-}Zhu, Li, Wang, Wang, and Chen}]{LORA}
Edward~J. Hu, Yelong Shen, Phillip Wallis, Zeyuan Allen{-}Zhu, Yuanzhi Li, Shean Wang, Lu~Wang, and Weizhu Chen. 2022.
\newblock \href {https://openreview.net/forum?id=nZeVKeeFYf9} {Lora: Low-rank adaptation of large language models}.
\newblock In \emph{The Tenth International Conference on Learning Representations, {ICLR} 2022, Virtual Event, April 25-29, 2022}. OpenReview.net.

\bibitem[{Huang et~al.(2024)Huang, Yong, Ma, Linghu, Li, Wang, Li, Zhu, Jia, and Huang}]{LEO}
Jiangyong Huang, Silong Yong, Xiaojian Ma, Xiongkun Linghu, Puhao Li, Yan Wang, Qing Li, Song{-}Chun Zhu, Baoxiong Jia, and Siyuan Huang. 2024.
\newblock \href {https://openreview.net/forum?id=V4qV08Vk6S} {An embodied generalist agent in 3d world}.
\newblock In \emph{Forty-first International Conference on Machine Learning, {ICML} 2024, Vienna, Austria, July 21-27, 2024}. OpenReview.net.

\bibitem[{Huang and Zhang(2024)}]{survey_evaluation_2}
Jiaxing Huang and Jingyi Zhang. 2024.
\newblock \href {https://doi.org/10.48550/ARXIV.2408.15769} {A survey on evaluation of multimodal large language models}.
\newblock \emph{CoRR}, abs/2408.15769.

\bibitem[{Huang et~al.(2022)Huang, Li, Qu, He, Zuo, and Ouyang}]{frozenclip}
Xiaoshui Huang, Sheng Li, Wentao Qu, Tong He, Yifan Zuo, and Wanli Ouyang. 2022.
\newblock \href {https://doi.org/10.48550/ARXIV.2212.04098} {Frozen {CLIP} model is an efficient point cloud backbone}.
\newblock \emph{CoRR}, abs/2212.04098.

\bibitem[{Hudson and Manning(2019)}]{GQA}
Drew~A. Hudson and Christopher~D. Manning. 2019.
\newblock \href {https://doi.org/10.1109/CVPR.2019.00686} {{GQA:} {A} new dataset for real-world visual reasoning and compositional question answering}.
\newblock In \emph{{IEEE} Conference on Computer Vision and Pattern Recognition, {CVPR} 2019, Long Beach, CA, USA, June 16-20, 2019}, pages 6700--6709. Computer Vision Foundation / {IEEE}.

\bibitem[{Jiang et~al.(2023)Jiang, Sablayrolles, Mensch, Bamford, Chaplot, de~Las~Casas, Bressand, Lengyel, Lample, Saulnier, Lavaud, Lachaux, Stock, Scao, Lavril, Wang, Lacroix, and Sayed}]{Mistral}
Albert~Q. Jiang, Alexandre Sablayrolles, Arthur Mensch, Chris Bamford, Devendra~Singh Chaplot, Diego de~Las~Casas, Florian Bressand, Gianna Lengyel, Guillaume Lample, Lucile Saulnier, L{\'{e}}lio~Renard Lavaud, Marie{-}Anne Lachaux, Pierre Stock, Teven~Le Scao, Thibaut Lavril, Thomas Wang, Timoth{\'{e}}e Lacroix, and William~El Sayed. 2023.
\newblock \href {https://doi.org/10.48550/ARXIV.2310.06825} {Mistral 7b}.
\newblock \emph{CoRR}, abs/2310.06825.

\bibitem[{Jiang et~al.(2024{\natexlab{a}})Jiang, Sablayrolles, Roux, Mensch, Savary, Bamford, Chaplot, de~Las~Casas, Hanna, Bressand, Lengyel, Bour, Lample, Lavaud, Saulnier, Lachaux, Stock, Subramanian, Yang, Antoniak, Scao, Gervet, Lavril, Wang, Lacroix, and Sayed}]{mixtral}
Albert~Q. Jiang, Alexandre Sablayrolles, Antoine Roux, Arthur Mensch, Blanche Savary, Chris Bamford, Devendra~Singh Chaplot, Diego de~Las~Casas, Emma~Bou Hanna, Florian Bressand, Gianna Lengyel, Guillaume Bour, Guillaume Lample, L{\'{e}}lio~Renard Lavaud, Lucile Saulnier, Marie{-}Anne Lachaux, Pierre Stock, Sandeep Subramanian, Sophia Yang, Szymon Antoniak, Teven~Le Scao, Th{\'{e}}ophile Gervet, Thibaut Lavril, Thomas Wang, Timoth{\'{e}}e Lacroix, and William~El Sayed. 2024{\natexlab{a}}.
\newblock \href {https://doi.org/10.48550/ARXIV.2401.04088} {Mixtral of experts}.
\newblock \emph{CoRR}, abs/2401.04088.

\bibitem[{Jiang et~al.(2024{\natexlab{b}})Jiang, Xiao, Lin, Zhang, Ren, Gao, Lin, Cai, Yang, and Liu}]{solami}
Jianping Jiang, Weiye Xiao, Zhengyu Lin, Huaizhong Zhang, Tianxiang Ren, Yang Gao, Zhiqian Lin, Zhongang Cai, Lei Yang, and Ziwei Liu. 2024{\natexlab{b}}.
\newblock \href {https://doi.org/10.48550/ARXIV.2412.00174} {{SOLAMI:} social vision-language-action modeling for immersive interaction with 3d autonomous characters}.
\newblock \emph{CoRR}, abs/2412.00174.

\bibitem[{Jin et~al.(2024)Jin, Sun, Xu, Xu, Chen, Jiang, Huang, Song, Liu, Zhang, Song, Gai, and Mu}]{Video-lavit}
Yang Jin, Zhicheng Sun, Kun Xu, Kun Xu, Liwei Chen, Hao Jiang, Quzhe Huang, Chengru Song, Yuliang Liu, Di~Zhang, Yang Song, Kun Gai, and Yadong Mu. 2024.
\newblock \href {https://openreview.net/forum?id=S9lk6dk4LL} {Video-lavit: Unified video-language pre-training with decoupled visual-motional tokenization}.
\newblock In \emph{Forty-first International Conference on Machine Learning, {ICML} 2024, Vienna, Austria, July 21-27, 2024}. OpenReview.net.

\bibitem[{Karatzas et~al.(2015)Karatzas, Gomez{-}Bigorda, Nicolaou, Ghosh, Bagdanov, Iwamura, Matas, Neumann, Chandrasekhar, Lu, Shafait, Uchida, and Valveny}]{IC15}
Dimosthenis Karatzas, Lluis Gomez{-}Bigorda, Anguelos Nicolaou, Suman~K. Ghosh, Andrew~D. Bagdanov, Masakazu Iwamura, Jiri Matas, Luk{\'{a}}s Neumann, Vijay~Ramaseshan Chandrasekhar, Shijian Lu, Faisal Shafait, Seiichi Uchida, and Ernest Valveny. 2015.
\newblock \href {https://doi.org/10.1109/ICDAR.2015.7333942} {{ICDAR} 2015 competition on robust reading}.
\newblock In \emph{13th International Conference on Document Analysis and Recognition, {ICDAR} 2015, Nancy, France, August 23-26, 2015}, pages 1156--1160. {IEEE} Computer Society.

\bibitem[{Karatzas et~al.(2013)Karatzas, Shafait, Uchida, Iwamura, i~Bigorda, Mestre, Mas, Mota, Almaz{\'{a}}n, and de~las Heras}]{IC13}
Dimosthenis Karatzas, Faisal Shafait, Seiichi Uchida, Masakazu Iwamura, Lluis~Gomez i~Bigorda, Sergi~Robles Mestre, Joan Mas, David~Fern{\'{a}}ndez Mota, Jon Almaz{\'{a}}n, and Llu{\'{\i}}s{-}Pere de~las Heras. 2013.
\newblock \href {https://doi.org/10.1109/ICDAR.2013.221} {{ICDAR} 2013 robust reading competition}.
\newblock In \emph{12th International Conference on Document Analysis and Recognition, {ICDAR} 2013, Washington, DC, USA, August 25-28, 2013}, pages 1484--1493. {IEEE} Computer Society.

\bibitem[{Kerr et~al.(2023)Kerr, Huang, Wilcox, Hoque, Ichnowski, Calandra, and Goldberg}]{SSVTP}
Justin Kerr, Huang Huang, Albert Wilcox, Ryan Hoque, Jeffrey Ichnowski, Roberto Calandra, and Ken Goldberg. 2023.
\newblock \href {https://doi.org/10.15607/RSS.2023.XIX.018} {Self-supervised visuo-tactile pretraining to locate and follow garment features}.
\newblock In \emph{Robotics: Science and Systems XIX, Daegu, Republic of Korea, July 10-14, 2023}.

\bibitem[{Kiela et~al.(2020)Kiela, Firooz, Mohan, Goswami, Singh, Ringshia, and Testuggine}]{HatefulMeme}
Douwe Kiela, Hamed Firooz, Aravind Mohan, Vedanuj Goswami, Amanpreet Singh, Pratik Ringshia, and Davide Testuggine. 2020.
\newblock \href {https://proceedings.neurips.cc/paper/2020/hash/1b84c4cee2b8b3d823b30e2d604b1878-Abstract.html} {The hateful memes challenge: Detecting hate speech in multimodal memes}.
\newblock In \emph{Advances in Neural Information Processing Systems 33: Annual Conference on Neural Information Processing Systems 2020, NeurIPS 2020, December 6-12, 2020, virtual}.

\bibitem[{Kim et~al.(2019)Kim, Kim, Lee, and Kim}]{audiocap}
Chris~Dongjoo Kim, Byeongchang Kim, Hyunmin Lee, and Gunhee Kim. 2019.
\newblock \href {https://doi.org/10.18653/V1/N19-1011} {Audiocaps: Generating captions for audios in the wild}.
\newblock In \emph{Proceedings of the 2019 Conference of the North American Chapter of the Association for Computational Linguistics: Human Language Technologies, {NAACL-HLT} 2019, Minneapolis, MN, USA, June 2-7, 2019, Volume 1 (Long and Short Papers)}, pages 119--132. Association for Computational Linguistics.

\bibitem[{Kim et~al.(2024{\natexlab{a}})Kim, BAEK, and Oh}]{egmi}
Taemin Kim, WOOYEOL BAEK, and Heeseok Oh. 2024{\natexlab{a}}.
\newblock \href {https://openreview.net/forum?id=5fWY2ZlsKj} {Efficient generative multimodal integration ({EGMI}): Enabling audio generation from text-image pairs through alignment with large language models}.
\newblock In \emph{Audio Imagination: NeurIPS 2024 Workshop AI-Driven Speech, Music, and Sound Generation}.

\bibitem[{Kim et~al.(2024{\natexlab{b}})Kim, Park, and Ro}]{EmpatheticLLM}
Yeonju Kim, Se~Jin Park, and Yong~Man Ro. 2024{\natexlab{b}}.
\newblock \href {https://doi.org/10.48550/ARXIV.2412.17572} {Empathetic response in audio-visual conversations using emotion preference optimization and mambacompressor}.
\newblock \emph{CoRR}, abs/2412.17572.

\bibitem[{Kong et~al.(2020)Kong, Cao, Iqbal, Wang, Wang, and Plumbley}]{pann}
Qiuqiang Kong, Yin Cao, Turab Iqbal, Yuxuan Wang, Wenwu Wang, and Mark~D. Plumbley. 2020.
\newblock \href {https://doi.org/10.1109/TASLP.2020.3030497} {Panns: Large-scale pretrained audio neural networks for audio pattern recognition}.
\newblock \emph{{IEEE} {ACM} Trans. Audio Speech Lang. Process.}, 28:2880--2894.

\bibitem[{Koska and Horv{\'{a}}th(2024)}]{eagel}
Ben Koska and Mojm{\'{\i}}r Horv{\'{a}}th. 2024.
\newblock \href {https://doi.org/10.48550/ARXIV.2411.05903} {Towards multi-modal mastery: {A} 4.5b parameter truly multi-modal small language model}.
\newblock \emph{CoRR}, abs/2411.05903.

\bibitem[{Krishna et~al.(2017{\natexlab{a}})Krishna, Hata, Ren, Fei{-}Fei, and Niebles}]{activitynet-caption}
Ranjay Krishna, Kenji Hata, Frederic Ren, Li~Fei{-}Fei, and Juan~Carlos Niebles. 2017{\natexlab{a}}.
\newblock \href {https://doi.org/10.1109/ICCV.2017.83} {Dense-captioning events in videos}.
\newblock In \emph{{IEEE} International Conference on Computer Vision, {ICCV} 2017, Venice, Italy, October 22-29, 2017}, pages 706--715. {IEEE} Computer Society.

\bibitem[{Krishna et~al.(2017{\natexlab{b}})Krishna, Zhu, Groth, Johnson, Hata, Kravitz, Chen, Kalantidis, Li, Shamma, Bernstein, and Fei{-}Fei}]{visualgenome}
Ranjay Krishna, Yuke Zhu, Oliver Groth, Justin Johnson, Kenji Hata, Joshua Kravitz, Stephanie Chen, Yannis Kalantidis, Li{-}Jia Li, David~A. Shamma, Michael~S. Bernstein, and Li~Fei{-}Fei. 2017{\natexlab{b}}.
\newblock \href {https://doi.org/10.1007/S11263-016-0981-7} {Visual genome: Connecting language and vision using crowdsourced dense image annotations}.
\newblock \emph{Int. J. Comput. Vis.}, 123(1):32--73.

\bibitem[{Lai et~al.(2024)Lai, Zhang, Liu, Li, Lu, and Guo}]{Spider}
Jinxiang Lai, Jie Zhang, Jun Liu, Jian Li, Xiaocheng Lu, and Song Guo. 2024.
\newblock \href {https://doi.org/10.48550/ARXIV.2411.09439} {Spider: Any-to-many multimodal {LLM}}.
\newblock \emph{CoRR}, abs/2411.09439.

\bibitem[{Lauren{\c{c}}on et~al.(2023)Lauren{\c{c}}on, Saulnier, Tronchon, Bekman, Singh, Lozhkov, Wang, Karamcheti, Rush, Kiela, Cord, and Sanh}]{obelics}
Hugo Lauren{\c{c}}on, Lucile Saulnier, L{\'{e}}o Tronchon, Stas Bekman, Amanpreet Singh, Anton Lozhkov, Thomas Wang, Siddharth Karamcheti, Alexander~M. Rush, Douwe Kiela, Matthieu Cord, and Victor Sanh. 2023.
\newblock \href {http://papers.nips.cc/paper\_files/paper/2023/hash/e2cfb719f58585f779d0a4f9f07bd618-Abstract-Datasets\_and\_Benchmarks.html} {{OBELICS:} an open web-scale filtered dataset of interleaved image-text documents}.
\newblock In \emph{Advances in Neural Information Processing Systems 36: Annual Conference on Neural Information Processing Systems 2023, NeurIPS 2023, New Orleans, LA, USA, December 10 - 16, 2023}.

\bibitem[{Lee et~al.(2023)Lee, Jeon, Cho, and Kim}]{rgb2ir}
Dong{-}Guw Lee, Myung{-}Hwan Jeon, Younggun Cho, and Ayoung Kim. 2023.
\newblock \href {https://doi.org/10.1109/ICRA48891.2023.10161210} {Edge-guided multi-domain rgb-to-tir image translation for training vision tasks with challenging labels}.
\newblock In \emph{{IEEE} International Conference on Robotics and Automation, {ICRA} 2023, London, UK, May 29 - June 2, 2023}, pages 8291--8298. {IEEE}.

\bibitem[{Leng et~al.(2024)Leng, Xing, Cheng, Zhou, Zhang, Li, Zhao, Lu, Miao, and Bing}]{curse}
Sicong Leng, Yun Xing, Zesen Cheng, Yang Zhou, Hang Zhang, Xin Li, Deli Zhao, Shijian Lu, Chunyan Miao, and Lidong Bing. 2024.
\newblock \href {https://doi.org/10.48550/ARXIV.2410.12787} {The curse of multi-modalities: Evaluating hallucinations of large multimodal models across language, visual, and audio}.
\newblock \emph{CoRR}, abs/2410.12787.

\bibitem[{Li et~al.(2022{\natexlab{a}})Li, Wei, Tian, Xu, Wen, and Hu}]{AVQA}
Guangyao Li, Yake Wei, Yapeng Tian, Chenliang Xu, Ji{-}Rong Wen, and Di~Hu. 2022{\natexlab{a}}.
\newblock \href {https://doi.org/10.1109/CVPR52688.2022.01852} {Learning to answer questions in dynamic audio-visual scenarios}.
\newblock In \emph{{IEEE/CVF} Conference on Computer Vision and Pattern Recognition, {CVPR} 2022, New Orleans, LA, USA, June 18-24, 2022}, pages 19086--19096. {IEEE}.

\bibitem[{Li et~al.(2022{\natexlab{b}})Li, Wei, Tian, Xu, Wen, and Hu}]{musicavqa}
Guangyao Li, Yake Wei, Yapeng Tian, Chenliang Xu, Ji{-}Rong Wen, and Di~Hu. 2022{\natexlab{b}}.
\newblock \href {https://doi.org/10.1109/CVPR52688.2022.01852} {Learning to answer questions in dynamic audio-visual scenarios}.
\newblock In \emph{{IEEE/CVF} Conference on Computer Vision and Pattern Recognition, {CVPR} 2022, New Orleans, LA, USA, June 18-24, 2022}, pages 19086--19096. {IEEE}.

\bibitem[{Li and Lu(2024)}]{survey_evaluation_1}
Jian Li and Weiheng Lu. 2024.
\newblock \href {https://doi.org/10.48550/ARXIV.2408.08632} {A survey on benchmarks of multimodal large language models}.
\newblock \emph{CoRR}, abs/2408.08632.

\bibitem[{Li et~al.(2022{\natexlab{c}})Li, Li, Xiong, and Hoi}]{BLIP}
Junnan Li, Dongxu Li, Caiming Xiong, and Steven C.~H. Hoi. 2022{\natexlab{c}}.
\newblock \href {https://proceedings.mlr.press/v162/li22n.html} {{BLIP:} bootstrapping language-image pre-training for unified vision-language understanding and generation}.
\newblock In \emph{International Conference on Machine Learning, {ICML} 2022, 17-23 July 2022, Baltimore, Maryland, {USA}}, volume 162 of \emph{Proceedings of Machine Learning Research}, pages 12888--12900. {PMLR}.

\bibitem[{Li et~al.(2024{\natexlab{a}})Li, Wang, He, Li, Wang, Liu, Wang, Xu, Chen, Lou, Wang, and Qiao}]{mvbench}
Kunchang Li, Yali Wang, Yinan He, Yizhuo Li, Yi~Wang, Yi~Liu, Zun Wang, Jilan Xu, Guo Chen, Ping Lou, Limin Wang, and Yu~Qiao. 2024{\natexlab{a}}.
\newblock \href {https://doi.org/10.1109/CVPR52733.2024.02095} {Mvbench: {A} comprehensive multi-modal video understanding benchmark}.
\newblock In \emph{{IEEE/CVF} Conference on Computer Vision and Pattern Recognition, {CVPR} 2024, Seattle, WA, USA, June 16-22, 2024}, pages 22195--22206. {IEEE}.

\bibitem[{Li et~al.(2020)Li, Chen, Cheng, Gan, Yu, and Liu}]{how2qa}
Linjie Li, Yen{-}Chun Chen, Yu~Cheng, Zhe Gan, Licheng Yu, and Jingjing Liu. 2020.
\newblock \href {https://doi.org/10.18653/V1/2020.EMNLP-MAIN.161} {{HERO:} hierarchical encoder for video+language omni-representation pre-training}.
\newblock In \emph{Proceedings of the 2020 Conference on Empirical Methods in Natural Language Processing, {EMNLP} 2020, Online, November 16-20, 2020}, pages 2046--2065. Association for Computational Linguistics.

\bibitem[{Li et~al.(2024{\natexlab{b}})Li, Chen, Zhang, Chen, Zhu, Yin, Li, Yu, and Chen}]{Omni-3D}
Mingsheng Li, Xin Chen, Chi Zhang, Sijin Chen, Hongyuan Zhu, Fukun Yin, Zhuoyuan Li, Gang Yu, and Tao Chen. 2024{\natexlab{b}}.
\newblock \href {https://doi.org/10.1007/978-3-031-73636-0\_3} {M3dbench: Towards omni 3d assistant with interleaved multi-modal instructions}.
\newblock In \emph{Computer Vision - {ECCV} 2024 - 18th European Conference, Milan, Italy, September 29-October 4, 2024, Proceedings, Part {LVIII}}, volume 15116 of \emph{Lecture Notes in Computer Science}, pages 41--59. Springer.

\bibitem[{Li et~al.(2025)Li, Liu, Zhang, Zhang, Chen, Li, Li, Liu, Ming, Dong, Pan, Li, Fang, Kuang, Wang, Zhu, Zhang, Guo, Zhang, Wang, Ding, Song, Li, Huo, Liang, Zhang, Wu, Zhao, Xiong, Wu, Ye, Lu, Li, Zhang, Zhou, Chen, Su, Zhang, Chen, Dong, Nie, Wu, Xiao, Li, Dang, Zhang, Sun, Wu, Yang, Lin, Ma, Wu, li, Yang, Liu, Zhang, Chen, Ai, Zhang, Chen, Huang, Li, Luo, Duan, Zhu, Xiao, Su, Pu, Wang, Jia, Zhang, Ai, Wang, Qiao, Zhang, Shen, Yang, Zhen, Zhou, Chen, Li, Zhu, Lu, Zhao, Liang, Li, Qin, Sun, Xu, Sun, Lin, Zhou, and Chen}]{baichuanomni1.5}
Yadong Li, Jun Liu, Tao Zhang, Tao Zhang, Song Chen, Tianpeng Li, Zehuan Li, Lijun Liu, Lingfeng Ming, Guosheng Dong, Da~Pan, Chong Li, Yuanbo Fang, Dongdong Kuang, Mingrui Wang, Chenglin Zhu, Youwei Zhang, Hongyu Guo, Fengyu Zhang, Yuran Wang, Bowen Ding, Wei Song, Xu~Li, Yuqi Huo, Zheng Liang, Shusen Zhang, Xin Wu, Shuai Zhao, Linchu Xiong, Yozhen Wu, Jiahui Ye, Wenhao Lu, Bowen Li, Yan Zhang, Yaqi Zhou, Xin Chen, Lei Su, Hongda Zhang, Fuzhong Chen, Xuezhen Dong, Na~Nie, Zhiying Wu, Bin Xiao, Ting Li, Shunya Dang, Ping Zhang, Yijia Sun, Jincheng Wu, Jinjie Yang, Xionghai Lin, Zhi Ma, Kegeng Wu, Jia li, Aiyuan Yang, Hui Liu, Jianqiang Zhang, Xiaoxi Chen, Guangwei Ai, Wentao Zhang, Yicong Chen, Xiaoqin Huang, Kun Li, Wenjing Luo, Yifei Duan, Lingling Zhu, Ran Xiao, Zhe Su, Jiani Pu, Dian Wang, Xu~Jia, Tianyu Zhang, Mengyu Ai, Mang Wang, Yujing Qiao, Lei Zhang, Yanjun Shen, Fan Yang, Miao Zhen, Yijie Zhou, Mingyang Chen, Fei Li, Chenzheng Zhu, Keer Lu, Yaqi Zhao, Hao Liang, Youquan Li, Yanzhao Qin, Linzhuang
  Sun, Jianhua Xu, Haoze Sun, Mingan Lin, Zenan Zhou, and Weipeng Chen. 2025.
\newblock \href {https://arxiv.org/abs/2501.15368} {Baichuan-omni-1.5 technical report}.
\newblock \emph{Preprint}, arXiv:2501.15368.

\bibitem[{Li et~al.(2024{\natexlab{c}})Li, Sun, Lin, Li, Dong, Zhang, Ding, Song, Cheng, Huo, Chen, Li, Pan, Zhang, Wu, Liang, Liu, Zhang, Lu, Zhao, Shen, Yang, Yu, Lin, Xu, Zhou, and Chen}]{baichuan-omni}
Yadong Li, Haoze Sun, Mingan Lin, Tianpeng Li, Guosheng Dong, Tao Zhang, Bowen Ding, Wei Song, Zhenglin Cheng, Yuqi Huo, Song Chen, Xu~Li, Da~Pan, Shusen Zhang, Xin Wu, Zheng Liang, Jun Liu, Tao Zhang, Keer Lu, Yaqi Zhao, Yanjun Shen, Fan Yang, Kaicheng Yu, Tao Lin, Jianhua Xu, Zenan Zhou, and Weipeng Chen. 2024{\natexlab{c}}.
\newblock \href {https://doi.org/10.48550/ARXIV.2410.08565} {Baichuan-omni technical report}.
\newblock \emph{CoRR}, abs/2410.08565.

\bibitem[{Li et~al.(2023)Li, Du, Zhou, Wang, Zhao, and Wen}]{POPE}
Yifan Li, Yifan Du, Kun Zhou, Jinpeng Wang, Wayne~Xin Zhao, and Ji{-}Rong Wen. 2023.
\newblock \href {https://doi.org/10.18653/V1/2023.EMNLP-MAIN.20} {Evaluating object hallucination in large vision-language models}.
\newblock In \emph{Proceedings of the 2023 Conference on Empirical Methods in Natural Language Processing, {EMNLP} 2023, Singapore, December 6-10, 2023}, pages 292--305. Association for Computational Linguistics.

\bibitem[{Li et~al.(2022{\natexlab{d}})Li, Han, and Mesgarani}]{styletts}
Yinghao~Aaron Li, Cong Han, and Nima Mesgarani. 2022{\natexlab{d}}.
\newblock \href {https://doi.org/10.48550/ARXIV.2205.15439} {Styletts: {A} style-based generative model for natural and diverse text-to-speech synthesis}.
\newblock \emph{CoRR}, abs/2205.15439.

\bibitem[{Li et~al.(2024{\natexlab{d}})Li, Yuan, Zhang, Ma, Chen, Yin, Xiao, Lin, Ragni, Benetos, Gyenge, Dannenberg, Liu, Chen, Xia, Shi, Huang, Wang, Guo, and Fu}]{MERT}
Yizhi Li, Ruibin Yuan, Ge~Zhang, Yinghao Ma, Xingran Chen, Hanzhi Yin, Chenghao Xiao, Chenghua Lin, Anton Ragni, Emmanouil Benetos, Norbert Gyenge, Roger~B. Dannenberg, Ruibo Liu, Wenhu Chen, Gus Xia, Yemin Shi, Wenhao Huang, Zili Wang, Yike Guo, and Jie Fu. 2024{\natexlab{d}}.
\newblock \href {https://openreview.net/forum?id=w3YZ9MSlBu} {{MERT:} acoustic music understanding model with large-scale self-supervised training}.
\newblock In \emph{The Twelfth International Conference on Learning Representations, {ICLR} 2024, Vienna, Austria, May 7-11, 2024}. OpenReview.net.

\bibitem[{Li et~al.(2024{\natexlab{e}})Li, Zhang, Ma, Yuan, Zhu, Guo, Liang, Liu, Yang, Wu, Qu, Shi, Zhang, Yang, Wang, Zhang, Liu, Benetos, Huang, and Lin}]{li2024omnibench}
Yizhi Li, Ge~Zhang, Yinghao Ma, Ruibin Yuan, Kang Zhu, Hangyu Guo, Yiming Liang, Jiaheng Liu, Jian Yang, Siwei Wu, Xingwei Qu, Jinjie Shi, Xinyue Zhang, Zhenzhu Yang, Xiangzhou Wang, Zhaoxiang Zhang, Zachary Liu, Emmanouil Benetos, Wenhao Huang, and Chenghua Lin. 2024{\natexlab{e}}.
\newblock \href {https://doi.org/10.48550/ARXIV.2409.15272} {Omnibench: Towards the future of universal omni-language models}.
\newblock \emph{CoRR}, abs/2409.15272.

\bibitem[{Li et~al.(2024{\natexlab{f}})Li, Jiang, Hu, Wang, Zhong, Luo, Ma, and Zhang}]{unimoe}
Yunxin Li, Shenyuan Jiang, Baotian Hu, Longyue Wang, Wanqi Zhong, Wenhan Luo, Lin Ma, and Min Zhang. 2024{\natexlab{f}}.
\newblock \href {https://doi.org/10.48550/ARXIV.2405.11273} {Uni-moe: Scaling unified multimodal llms with mixture of experts}.
\newblock \emph{CoRR}, abs/2405.11273.

\bibitem[{Li et~al.(2024{\natexlab{g}})Li, Wang, Cai, Xu, Wang, Zhang, Song, Jiang, Huang, and Wang}]{unifiedmllm}
Zhaowei Li, Wei Wang, Yiqing Cai, Qi~Xu, Pengyu Wang, Dong Zhang, Hang Song, Botian Jiang, Zhida Huang, and Tao Wang. 2024{\natexlab{g}}.
\newblock \href {https://doi.org/10.48550/ARXIV.2408.02503} {Unifiedmllm: Enabling unified representation for multi-modal multi-tasks with large language model}.
\newblock \emph{CoRR}, abs/2408.02503.

\bibitem[{Li et~al.(2024{\natexlab{h}})Li, Xu, Zhang, Song, Cai, Qi, Zhou, Pan, Li, Tu, Huang, and Wang}]{groundinggpt}
Zhaowei Li, Qi~Xu, Dong Zhang, Hang Song, Yiqing Cai, Qi~Qi, Ran Zhou, Junting Pan, Zefeng Li, Vu~Tu, Zhida Huang, and Tao Wang. 2024{\natexlab{h}}.
\newblock \href {https://doi.org/10.18653/V1/2024.ACL-LONG.360} {Groundinggpt: Language enhanced multi-modal grounding model}.
\newblock In \emph{Proceedings of the 62nd Annual Meeting of the Association for Computational Linguistics (Volume 1: Long Papers), {ACL} 2024, Bangkok, Thailand, August 11-16, 2024}, pages 6657--6678. Association for Computational Linguistics.

\bibitem[{Liang et~al.(2024)Liang, Huang, Kong, Chen, and Zhu}]{qap}
Tian Liang, Jing Huang, Ming Kong, Luyuan Chen, and Qiang Zhu. 2024.
\newblock \href {https://doi.org/10.1109/CVPR52733.2024.02536} {Querying as prompt: Parameter-efficient learning for multimodal language model}.
\newblock In \emph{{IEEE/CVF} Conference on Computer Vision and Pattern Recognition, {CVPR} 2024, Seattle, WA, USA, June 16-22, 2024}, pages 26845--26855. {IEEE}.

\bibitem[{Lin et~al.(2024)Lin, Yin, Ping, Molchanov, Shoeybi, and Han}]{vila}
Ji~Lin, Hongxu Yin, Wei Ping, Pavlo Molchanov, Mohammad Shoeybi, and Song Han. 2024.
\newblock \href {https://doi.org/10.1109/CVPR52733.2024.02520} {{VILA:} on pre-training for visual language models}.
\newblock In \emph{{IEEE/CVF} Conference on Computer Vision and Pattern Recognition, {CVPR} 2024, Seattle, WA, USA, June 16-22, 2024}, pages 26679--26689. {IEEE}.

\bibitem[{Lin et~al.(2014)Lin, Maire, Belongie, Hays, Perona, Ramanan, Doll{\'{a}}r, and Zitnick}]{MSCOCO}
Tsung{-}Yi Lin, Michael Maire, Serge~J. Belongie, James Hays, Pietro Perona, Deva Ramanan, Piotr Doll{\'{a}}r, and C.~Lawrence Zitnick. 2014.
\newblock \href {https://doi.org/10.1007/978-3-319-10602-1\_48} {Microsoft {COCO:} common objects in context}.
\newblock In \emph{Computer Vision - {ECCV} 2014 - 13th European Conference, Zurich, Switzerland, September 6-12, 2014, Proceedings, Part {V}}, volume 8693 of \emph{Lecture Notes in Computer Science}, pages 740--755. Springer.

\bibitem[{Lipping et~al.(2022)Lipping, Sudarsanam, Drossos, and Virtanen}]{clothoaqa}
Samuel Lipping, Parthasaarathy Sudarsanam, Konstantinos Drossos, and Tuomas Virtanen. 2022.
\newblock \href {https://ieeexplore.ieee.org/document/9909680} {Clotho-aqa: {A} crowdsourced dataset for audio question answering}.
\newblock In \emph{30th European Signal Processing Conference, {EUSIPCO} 2022, Belgrade, Serbia, August 29 - Sept. 2, 2022}, pages 1140--1144. {IEEE}.

\bibitem[{Liu et~al.(2023{\natexlab{a}})Liu, Emerson, and Collier}]{VSR}
Fangyu Liu, Guy Emerson, and Nigel Collier. 2023{\natexlab{a}}.
\newblock \href {https://doi.org/10.1162/TACL\_A\_00566} {Visual spatial reasoning}.
\newblock \emph{Trans. Assoc. Comput. Linguistics}, 11:635--651.

\bibitem[{Liu et~al.(2023{\natexlab{b}})Liu, Chen, Yuan, Mei, Liu, Mandic, Wang, and Plumbley}]{audioldm}
Haohe Liu, Zehua Chen, Yi~Yuan, Xinhao Mei, Xubo Liu, Danilo~P. Mandic, Wenwu Wang, and Mark~D. Plumbley. 2023{\natexlab{b}}.
\newblock \href {https://proceedings.mlr.press/v202/liu23f.html} {Audioldm: Text-to-audio generation with latent diffusion models}.
\newblock In \emph{International Conference on Machine Learning, {ICML} 2023, 23-29 July 2023, Honolulu, Hawaii, {USA}}, volume 202 of \emph{Proceedings of Machine Learning Research}, pages 21450--21474. {PMLR}.

\bibitem[{Liu et~al.(2024{\natexlab{a}})Liu, Li, Li, and Lee}]{llava1.5}
Haotian Liu, Chunyuan Li, Yuheng Li, and Yong~Jae Lee. 2024{\natexlab{a}}.
\newblock \href {https://doi.org/10.1109/CVPR52733.2024.02484} {Improved baselines with visual instruction tuning}.
\newblock In \emph{{IEEE/CVF} Conference on Computer Vision and Pattern Recognition, {CVPR} 2024, Seattle, WA, USA, June 16-22, 2024}, pages 26286--26296. {IEEE}.

\bibitem[{Liu et~al.(2023{\natexlab{c}})Liu, Li, Wu, and Lee}]{llava}
Haotian Liu, Chunyuan Li, Qingyang Wu, and Yong~Jae Lee. 2023{\natexlab{c}}.
\newblock \href {http://papers.nips.cc/paper\_files/paper/2023/hash/6dcf277ea32ce3288914faf369fe6de0-Abstract-Conference.html} {Visual instruction tuning}.
\newblock In \emph{Advances in Neural Information Processing Systems 36: Annual Conference on Neural Information Processing Systems 2023, NeurIPS 2023, New Orleans, LA, USA, December 10 - 16, 2023}.

\bibitem[{Liu et~al.(2024{\natexlab{b}})Liu, Hussain, Wu, Sun, and Shan}]{mumullama}
Shansong Liu, Atin~Sakkeer Hussain, Qilong Wu, Chenshuo Sun, and Ying Shan. 2024{\natexlab{b}}.
\newblock \href {https://doi.org/10.48550/ARXIV.2412.06660} {Mumu-llama: Multi-modal music understanding and generation via large language models}.
\newblock \emph{CoRR}, abs/2412.06660.

\bibitem[{Liu et~al.(2024{\natexlab{c}})Liu, Cun, Liu, Wang, Zhang, Chen, Liu, Zeng, Chan, and Shan}]{evalcrafter}
Yaofang Liu, Xiaodong Cun, Xuebo Liu, Xintao Wang, Yong Zhang, Haoxin Chen, Yang Liu, Tieyong Zeng, Raymond Chan, and Ying Shan. 2024{\natexlab{c}}.
\newblock \href {https://doi.org/10.1109/CVPR52733.2024.02090} {Evalcrafter: Benchmarking and evaluating large video generation models}.
\newblock In \emph{{IEEE/CVF} Conference on Computer Vision and Pattern Recognition, {CVPR} 2024, Seattle, WA, USA, June 16-22, 2024}, pages 22139--22149. {IEEE}.

\bibitem[{Liu et~al.(2024{\natexlab{d}})Liu, Duan, Zhang, Li, Zhang, Zhao, Yuan, Wang, He, Liu, Chen, and Lin}]{mmb}
Yuan Liu, Haodong Duan, Yuanhan Zhang, Bo~Li, Songyang Zhang, Wangbo Zhao, Yike Yuan, Jiaqi Wang, Conghui He, Ziwei Liu, Kai Chen, and Dahua Lin. 2024{\natexlab{d}}.
\newblock \href {https://doi.org/10.1007/978-3-031-72658-3\_13} {Mmbench: Is your multi-modal model an all-around player?}
\newblock In \emph{Computer Vision - {ECCV} 2024 - 18th European Conference, Milan, Italy, September 29-October 4, 2024, Proceedings, Part {VI}}, volume 15064 of \emph{Lecture Notes in Computer Science}, pages 216--233. Springer.

\bibitem[{Liu et~al.(2022{\natexlab{a}})Liu, Ning, Cao, Wei, Zhang, Lin, and Hu}]{videoswin}
Ze~Liu, Jia Ning, Yue Cao, Yixuan Wei, Zheng Zhang, Stephen Lin, and Han Hu. 2022{\natexlab{a}}.
\newblock \href {https://doi.org/10.1109/CVPR52688.2022.00320} {Video swin transformer}.
\newblock In \emph{{IEEE/CVF} Conference on Computer Vision and Pattern Recognition, {CVPR} 2022, New Orleans, LA, USA, June 18-24, 2022}, pages 3192--3201. {IEEE}.

\bibitem[{Liu et~al.(2022{\natexlab{b}})Liu, Mao, Wu, Feichtenhofer, Darrell, and Xie}]{convnext}
Zhuang Liu, Hanzi Mao, Chao{-}Yuan Wu, Christoph Feichtenhofer, Trevor Darrell, and Saining Xie. 2022{\natexlab{b}}.
\newblock \href {https://doi.org/10.1109/CVPR52688.2022.01167} {A convnet for the 2020s}.
\newblock In \emph{{IEEE/CVF} Conference on Computer Vision and Pattern Recognition, {CVPR} 2022, New Orleans, LA, USA, June 18-24, 2022}, pages 11966--11976. {IEEE}.

\bibitem[{Lu et~al.(2024{\natexlab{a}})Lu, Clark, Lee, Zhang, Khosla, Marten, Hoiem, and Kembhavi}]{unifiedio2}
Jiasen Lu, Christopher Clark, Sangho Lee, Zichen Zhang, Savya Khosla, Ryan Marten, Derek Hoiem, and Aniruddha Kembhavi. 2024{\natexlab{a}}.
\newblock \href {https://doi.org/10.1109/CVPR52733.2024.02497} {Unified-io 2: Scaling autoregressive multimodal models with vision, language, audio, and action}.
\newblock In \emph{{IEEE/CVF} Conference on Computer Vision and Pattern Recognition, {CVPR} 2024, Seattle, WA, USA, June 16-22, 2024}, pages 26429--26445. {IEEE}.

\bibitem[{Lu et~al.(2022)Lu, Mishra, Xia, Qiu, Chang, Zhu, Tafjord, Clark, and Kalyan}]{ScienceQA}
Pan Lu, Swaroop Mishra, Tanglin Xia, Liang Qiu, Kai{-}Wei Chang, Song{-}Chun Zhu, Oyvind Tafjord, Peter Clark, and Ashwin Kalyan. 2022.
\newblock \href {http://papers.nips.cc/paper\_files/paper/2022/hash/11332b6b6cf4485b84afadb1352d3a9a-Abstract-Conference.html} {Learn to explain: Multimodal reasoning via thought chains for science question answering}.
\newblock In \emph{Advances in Neural Information Processing Systems 35: Annual Conference on Neural Information Processing Systems 2022, NeurIPS 2022, New Orleans, LA, USA, November 28 - December 9, 2022}.

\bibitem[{Lu et~al.(2021)Lu, Qiu, Chen, Xia, Zhao, Zhang, Yu, Liang, and Zhu}]{IconQA}
Pan Lu, Liang Qiu, Jiaqi Chen, Tanglin Xia, Yizhou Zhao, Wei Zhang, Zhou Yu, Xiaodan Liang, and Song{-}Chun Zhu. 2021.
\newblock \href {https://datasets-benchmarks-proceedings.neurips.cc/paper/2021/hash/d3d9446802a44259755d38e6d163e820-Abstract-round2.html} {Iconqa: {A} new benchmark for abstract diagram understanding and visual language reasoning}.
\newblock In \emph{Proceedings of the Neural Information Processing Systems Track on Datasets and Benchmarks 1, NeurIPS Datasets and Benchmarks 2021, December 2021, virtual}.

\bibitem[{Lu et~al.(2024{\natexlab{b}})Lu, Song, Chang, Bian, Maiti, and Watanabe}]{SynesLM}
Yichen Lu, Jiaqi Song, Xuankai Chang, Hengwei Bian, Soumi Maiti, and Shinji Watanabe. 2024{\natexlab{b}}.
\newblock \href {https://doi.org/10.48550/ARXIV.2408.00624} {Syneslm: {A} unified approach for audio-visual speech recognition and translation via language model and synthetic data}.
\newblock \emph{CoRR}, abs/2408.00624.

\bibitem[{Luo et~al.(2024)Luo, Hou, Chang, Liu, Wang, and Shan}]{M3GPT}
Mingshuang Luo, Ruibing Hou, Hong Chang, Zimo Liu, Yaowei Wang, and Shiguang Shan. 2024.
\newblock \href {https://doi.org/10.48550/ARXIV.2405.16273} {M\({}^{\mbox{3}}\)gpt: An advanced multimodal, multitask framework for motion comprehension and generation}.
\newblock \emph{CoRR}, abs/2405.16273.

\bibitem[{Luo et~al.(2023{\natexlab{a}})Luo, Zhao, Yang, Dong, Qiu, Lu, Wang, and Wei}]{valley}
Ruipu Luo, Ziwang Zhao, Min Yang, Junwei Dong, Minghui Qiu, Pengcheng Lu, Tao Wang, and Zhongyu Wei. 2023{\natexlab{a}}.
\newblock \href {https://doi.org/10.48550/ARXIV.2306.07207} {Valley: Video assistant with large language model enhanced ability}.
\newblock \emph{CoRR}, abs/2306.07207.

\bibitem[{Luo et~al.(2025)Luo, Lin, Zhang, Wu, Liu, Yang, Li, Chen, Li, Zhang, Chen, Alinejad-Rokny, and Huang}]{openomni}
Run Luo, Ting-En Lin, Haonan Zhang, Yuchuan Wu, Xiong Liu, Min Yang, Yongbin Li, Longze Chen, Jiaming Li, Lei Zhang, Yangyi Chen, Hamid Alinejad-Rokny, and Fei Huang. 2025.
\newblock \href {https://arxiv.org/abs/2501.04561} {Openomni: Large language models pivot zero-shot omnimodal alignment across language with real-time self-aware emotional speech synthesis}.
\newblock \emph{Preprint}, arXiv:2501.04561.

\bibitem[{Luo et~al.(2023{\natexlab{b}})Luo, Rockwell, Lee, and Johnson}]{cap3d}
Tiange Luo, Chris Rockwell, Honglak Lee, and Justin Johnson. 2023{\natexlab{b}}.
\newblock \href {http://papers.nips.cc/paper\_files/paper/2023/hash/ee4814f9bce0cae7991d3341bb081b55-Abstract-Datasets\_and\_Benchmarks.html} {Scalable 3d captioning with pretrained models}.
\newblock In \emph{Advances in Neural Information Processing Systems 36: Annual Conference on Neural Information Processing Systems 2023, NeurIPS 2023, New Orleans, LA, USA, December 10 - 16, 2023}.

\bibitem[{Luo et~al.(2023{\natexlab{c}})Luo, Chen, Zhang, Huang, Wang, Shen, Zhao, Zhou, and Tan}]{VideoFusion}
Zhengxiong Luo, Dayou Chen, Yingya Zhang, Yan Huang, Liang Wang, Yujun Shen, Deli Zhao, Jingren Zhou, and Tieniu Tan. 2023{\natexlab{c}}.
\newblock \href {https://doi.org/10.48550/ARXIV.2303.08320} {Videofusion: Decomposed diffusion models for high-quality video generation}.
\newblock \emph{CoRR}, abs/2303.08320.

\bibitem[{Lyu et~al.(2023)Lyu, Wu, Wang, Huang, Liu, Du, Shi, and Tu}]{macawllm}
Chenyang Lyu, Minghao Wu, Longyue Wang, Xinting Huang, Bingshuai Liu, Zefeng Du, Shuming Shi, and Zhaopeng Tu. 2023.
\newblock \href {https://doi.org/10.48550/ARXIV.2306.09093} {Macaw-llm: Multi-modal language modeling with image, audio, video, and text integration}.
\newblock \emph{CoRR}, abs/2306.09093.

\bibitem[{Ma et~al.(2024)Ma, Bhalgat, Smart, Chen, Li, Ding, Gu, Chen, Peng, Bian, Torr, Pollefeys, Nie{\ss}ner, Reid, Chang, Laina, and Prisacariu}]{3d_survey}
Xianzheng Ma, Yash Bhalgat, Brandon Smart, Shuai Chen, Xinghui Li, Jian Ding, Jindong Gu, Dave~Zhenyu Chen, Songyou Peng, Jia{-}Wang Bian, Philip H.~S. Torr, Marc Pollefeys, Matthias Nie{\ss}ner, Ian~D. Reid, Angel~X. Chang, Iro Laina, and Victor~Adrian Prisacariu. 2024.
\newblock \href {https://doi.org/10.48550/ARXIV.2405.10255} {When llms step into the 3d world: {A} survey and meta-analysis of 3d tasks via multi-modal large language models}.
\newblock \emph{CoRR}, abs/2405.10255.

\bibitem[{Ma et~al.(2023)Ma, Yong, Zheng, Li, Liang, Zhu, and Huang}]{SQA3D}
Xiaojian Ma, Silong Yong, Zilong Zheng, Qing Li, Yitao Liang, Song{-}Chun Zhu, and Siyuan Huang. 2023.
\newblock \href {https://openreview.net/forum?id=IDJx97BC38} {{SQA3D:} situated question answering in 3d scenes}.
\newblock In \emph{The Eleventh International Conference on Learning Representations, {ICLR} 2023, Kigali, Rwanda, May 1-5, 2023}. OpenReview.net.

\bibitem[{Maaz et~al.(2024)Maaz, Rasheed, Khan, and Khan}]{videochatgpt}
Muhammad Maaz, Hanoona~Abdul Rasheed, Salman Khan, and Fahad Khan. 2024.
\newblock \href {https://doi.org/10.18653/V1/2024.ACL-LONG.679} {Video-chatgpt: Towards detailed video understanding via large vision and language models}.
\newblock In \emph{Proceedings of the 62nd Annual Meeting of the Association for Computational Linguistics (Volume 1: Long Papers), {ACL} 2024, Bangkok, Thailand, August 11-16, 2024}, pages 12585--12602. Association for Computational Linguistics.

\bibitem[{Mangalam et~al.(2023)Mangalam, Akshulakov, and Malik}]{egoschema}
Karttikeya Mangalam, Raiymbek Akshulakov, and Jitendra Malik. 2023.
\newblock \href {http://papers.nips.cc/paper\_files/paper/2023/hash/90ce332aff156b910b002ce4e6880dec-Abstract-Datasets\_and\_Benchmarks.html} {Egoschema: {A} diagnostic benchmark for very long-form video language understanding}.
\newblock In \emph{Advances in Neural Information Processing Systems 36: Annual Conference on Neural Information Processing Systems 2023, NeurIPS 2023, New Orleans, LA, USA, December 10 - 16, 2023}.

\bibitem[{Mao et~al.(2016)Mao, Huang, Toshev, Camburu, Yuille, and Murphy}]{refcocog}
Junhua Mao, Jonathan Huang, Alexander Toshev, Oana Camburu, Alan~L. Yuille, and Kevin Murphy. 2016.
\newblock \href {https://doi.org/10.1109/CVPR.2016.9} {Generation and comprehension of unambiguous object descriptions}.
\newblock In \emph{2016 {IEEE} Conference on Computer Vision and Pattern Recognition, {CVPR} 2016, Las Vegas, NV, USA, June 27-30, 2016}, pages 11--20. {IEEE} Computer Society.

\bibitem[{Marino et~al.(2019)Marino, Rastegari, Farhadi, and Mottaghi}]{OKVQA}
Kenneth Marino, Mohammad Rastegari, Ali Farhadi, and Roozbeh Mottaghi. 2019.
\newblock \href {https://doi.org/10.1109/CVPR.2019.00331} {{OK-VQA:} {A} visual question answering benchmark requiring external knowledge}.
\newblock In \emph{{IEEE} Conference on Computer Vision and Pattern Recognition, {CVPR} 2019, Long Beach, CA, USA, June 16-20, 2019}, pages 3195--3204. Computer Vision Foundation / {IEEE}.

\bibitem[{Mathew et~al.(2021)Mathew, Karatzas, and Jawahar}]{DocVQA}
Minesh Mathew, Dimosthenis Karatzas, and C.~V. Jawahar. 2021.
\newblock \href {https://doi.org/10.1109/WACV48630.2021.00225} {Docvqa: {A} dataset for {VQA} on document images}.
\newblock In \emph{{IEEE} Winter Conference on Applications of Computer Vision, {WACV} 2021, Waikoloa, HI, USA, January 3-8, 2021}, pages 2199--2208. {IEEE}.

\bibitem[{Mei et~al.(2024)Mei, Meng, Liu, Kong, Ko, Zhao, Plumbley, Zou, and Wang}]{wavcaps}
Xinhao Mei, Chutong Meng, Haohe Liu, Qiuqiang Kong, Tom Ko, Chengqi Zhao, Mark~D. Plumbley, Yuexian Zou, and Wenwu Wang. 2024.
\newblock \href {https://doi.org/10.1109/TASLP.2024.3419446} {Wavcaps: {A} chatgpt-assisted weakly-labelled audio captioning dataset for audio-language multimodal research}.
\newblock \emph{{IEEE} {ACM} Trans. Audio Speech Lang. Process.}, 32:3339--3354.

\bibitem[{Mesaros et~al.(2016)Mesaros, Heittola, and Virtanen}]{tut2017}
Annamaria Mesaros, Toni Heittola, and Tuomas Virtanen. 2016.
\newblock \href {https://doi.org/10.1109/EUSIPCO.2016.7760424} {{TUT} database for acoustic scene classification and sound event detection}.
\newblock In \emph{24th European Signal Processing Conference, {EUSIPCO} 2016, Budapest, Hungary, August 29 - September 2, 2016}, pages 1128--1132. {IEEE}.

\bibitem[{MiniCPM-o Team(2025)}]{minicpm-o-2.6}
OpenBMB MiniCPM-o Team. 2025.
\newblock Minicpm-o 2.6: A gpt-4o level mllm for vision, speech, and multimodal live streaming on your phone.
\newblock \url{https://github.com/OpenBMB/MiniCPM-o}.
\newblock Accessed: 2025-02-10.

\bibitem[{Mishra et~al.(2012)Mishra, Alahari, and Jawahar}]{IIIT5K}
Anand Mishra, Karteek Alahari, and C.~V. Jawahar. 2012.
\newblock \href {https://doi.org/10.5244/C.26.127} {Scene text recognition using higher order language priors}.
\newblock In \emph{British Machine Vision Conference, {BMVC} 2012, Surrey, UK, September 3-7, 2012}, pages 1--11. {BMVA} Press.

\bibitem[{Mishra et~al.(2019)Mishra, Shekhar, Singh, and Chakraborty}]{OCR-VQA}
Anand Mishra, Shashank Shekhar, Ajeet~Kumar Singh, and Anirban Chakraborty. 2019.
\newblock \href {https://doi.org/10.1109/ICDAR.2019.00156} {{OCR-VQA:} visual question answering by reading text in images}.
\newblock In \emph{2019 International Conference on Document Analysis and Recognition, {ICDAR} 2019, Sydney, Australia, September 20-25, 2019}, pages 947--952. {IEEE}.

\bibitem[{Moon et~al.(2022)Moon, Madotto, Lin, Dirafzoon, Saraf, Bearman, and Damavandi}]{IMU2CLIP}
Seungwhan Moon, Andrea Madotto, Zhaojiang Lin, Alireza Dirafzoon, Aparajita Saraf, Amy Bearman, and Babak Damavandi. 2022.
\newblock \href {https://doi.org/10.48550/ARXIV.2210.14395} {{IMU2CLIP:} multimodal contrastive learning for {IMU} motion sensors from egocentric videos and text}.
\newblock \emph{CoRR}, abs/2210.14395.

\bibitem[{Moon et~al.(2024)Moon, Madotto, Lin, Nagarajan, Smith, Jain, Yeh, Murugesan, Heidari, Liu, Srinet, Damavandi, and Kumar}]{anymal}
Seungwhan Moon, Andrea Madotto, Zhaojiang Lin, Tushar Nagarajan, Matt Smith, Shashank Jain, Chun{-}Fu Yeh, Prakash Murugesan, Peyman Heidari, Yue Liu, Kavya Srinet, Babak Damavandi, and Anuj Kumar. 2024.
\newblock \href {https://aclanthology.org/2024.emnlp-industry.98} {Anymal: An efficient and scalable any-modality augmented language model}.
\newblock In \emph{Proceedings of the 2024 Conference on Empirical Methods in Natural Language Processing: {EMNLP} 2024 - Industry Track, Miami, Florida, USA, November 12-16, 2024}, pages 1314--1332. Association for Computational Linguistics.

\bibitem[{OpenAI()}]{gpt4o}
OpenAI.
\newblock Hello gpt-4o.
\newblock \url{https://openai.com/index/hello-gpt-4o/}.
\newblock Accessed: January 3, 2025.

\bibitem[{OpenAI(2023{\natexlab{a}})}]{chatgpt}
OpenAI. 2023{\natexlab{a}}.
\newblock Chatgpt.
\newblock Technical report, OpenAI.

\bibitem[{OpenAI(2023{\natexlab{b}})}]{gpt4}
OpenAI. 2023{\natexlab{b}}.
\newblock \href {https://doi.org/10.48550/ARXIV.2303.08774} {{GPT-4} technical report}.
\newblock \emph{CoRR}, abs/2303.08774.

\bibitem[{OpenGV(2024)}]{interomni}
OpenGV. 2024.
\newblock \href {https://internvl.github.io/blog/2024-07-27-InternOmni/} {Interomni}.
\newblock Accessed: 2024-07-27.

\bibitem[{Oquab et~al.(2024)Oquab, Darcet, Moutakanni, Vo, Szafraniec, Khalidov, Fernandez, Haziza, Massa, El{-}Nouby, Assran, Ballas, Galuba, Howes, Huang, Li, Misra, Rabbat, Sharma, Synnaeve, Xu, J{\'{e}}gou, Mairal, Labatut, Joulin, and Bojanowski}]{dinov2}
Maxime Oquab, Timoth{\'{e}}e Darcet, Th{\'{e}}o Moutakanni, Huy~V. Vo, Marc Szafraniec, Vasil Khalidov, Pierre Fernandez, Daniel Haziza, Francisco Massa, Alaaeldin El{-}Nouby, Mido Assran, Nicolas Ballas, Wojciech Galuba, Russell Howes, Po{-}Yao Huang, Shang{-}Wen Li, Ishan Misra, Michael Rabbat, Vasu Sharma, Gabriel Synnaeve, Hu~Xu, Herv{\'{e}} J{\'{e}}gou, Julien Mairal, Patrick Labatut, Armand Joulin, and Piotr Bojanowski. 2024.
\newblock \href {https://openreview.net/forum?id=a68SUt6zFt} {Dinov2: Learning robust visual features without supervision}.
\newblock \emph{Trans. Mach. Learn. Res.}, 2024.

\bibitem[{Ordonez et~al.(2011)Ordonez, Kulkarni, and Berg}]{SBU}
Vicente Ordonez, Girish Kulkarni, and Tamara~L. Berg. 2011.
\newblock \href {https://proceedings.neurips.cc/paper/2011/hash/5dd9db5e033da9c6fb5ba83c7a7ebea9-Abstract.html} {Im2text: Describing images using 1 million captioned photographs}.
\newblock In \emph{Advances in Neural Information Processing Systems 24: 25th Annual Conference on Neural Information Processing Systems 2011. Proceedings of a meeting held 12-14 December 2011, Granada, Spain}, pages 1143--1151.

\bibitem[{Ormazabal et~al.(2024)Ormazabal, Zheng, de~Masson~d'Autume, Yogatama, Fu, Ong, Chen, Lamprecht, Pham, Ong, Aleksiev, Li, Henderson, Bain, Artetxe, Relan, Padlewski, Liu, Chen, Phua, Yang, Tay, Wang, Zhu, and Xie}]{reka}
Aitor Ormazabal, Che Zheng, Cyprien de~Masson~d'Autume, Dani Yogatama, Deyu Fu, Donovan Ong, Eric Chen, Eugenie Lamprecht, Hai Pham, Isaac Ong, Kaloyan Aleksiev, Lei Li, Matthew Henderson, Max Bain, Mikel Artetxe, Nishant Relan, Piotr Padlewski, Qi~Liu, Ren Chen, Samuel Phua, Yazheng Yang, Yi~Tay, Yuqi Wang, Zhongkai Zhu, and Zhihui Xie. 2024.
\newblock \href {https://doi.org/10.48550/ARXIV.2404.12387} {Reka core, flash, and edge: {A} series of powerful multimodal language models}.
\newblock \emph{CoRR}, abs/2404.12387.

\bibitem[{Panagopoulou et~al.(2024)Panagopoulou, Xue, Yu, Li, Li, Joty, Xu, Savarese, Xiong, and Niebles}]{xinstruclblip}
Artemis Panagopoulou, Le~Xue, Ning Yu, Junnan Li, Dongxu Li, Shafiq Joty, Ran Xu, Silvio Savarese, Caiming Xiong, and Juan~Carlos Niebles. 2024.
\newblock \href {https://doi.org/10.1007/978-3-031-72995-9\_11} {X-instructblip: {A} framework for aligning image, 3d, audio, video to llms and its emergent cross-modal reasoning}.
\newblock In \emph{Computer Vision - {ECCV} 2024 - 18th European Conference, Milan, Italy, September 29-October 4, 2024, Proceedings, Part {XLV}}, volume 15103 of \emph{Lecture Notes in Computer Science}, pages 177--197. Springer.

\bibitem[{Panayotov et~al.(2015)Panayotov, Chen, Povey, and Khudanpur}]{librispeech}
Vassil Panayotov, Guoguo Chen, Daniel Povey, and Sanjeev Khudanpur. 2015.
\newblock \href {https://doi.org/10.1109/ICASSP.2015.7178964} {Librispeech: An {ASR} corpus based on public domain audio books}.
\newblock In \emph{2015 {IEEE} International Conference on Acoustics, Speech and Signal Processing, {ICASSP} 2015, South Brisbane, Queensland, Australia, April 19-24, 2015}, pages 5206--5210. {IEEE}.

\bibitem[{Pang et~al.(2024)Pang, Ding, He, Tao, Zhang, and Gan}]{Gesticulator}
Haozhou Pang, Tianwei Ding, Lanshan He, Ming Tao, Lu~Zhang, and Qi~Gan. 2024.
\newblock \href {https://doi.org/10.48550/ARXIV.2410.10851} {{LLM} gesticulator: Leveraging large language models for scalable and controllable co-speech gesture synthesis}.
\newblock \emph{CoRR}, abs/2410.10851.

\bibitem[{Patraucean et~al.(2023)Patraucean, Smaira, Gupta, Recasens, Markeeva, Banarse, Koppula, Heyward, Malinowski, Yang, Doersch, Matejovicova, Sulsky, Miech, Fr{\'{e}}chette, Klimczak, Koster, Zhang, Winkler, Aytar, Osindero, Damen, Zisserman, and Carreira}]{preception-test}
Viorica Patraucean, Lucas Smaira, Ankush Gupta, Adri{\`{a}} Recasens, Larisa Markeeva, Dylan Banarse, Skanda Koppula, Joseph Heyward, Mateusz Malinowski, Yi~Yang, Carl Doersch, Tatiana Matejovicova, Yury Sulsky, Antoine Miech, Alexandre Fr{\'{e}}chette, Hanna Klimczak, Raphael Koster, Junlin Zhang, Stephanie Winkler, Yusuf Aytar, Simon Osindero, Dima Damen, Andrew Zisserman, and Jo{\~{a}}o Carreira. 2023.
\newblock \href {http://papers.nips.cc/paper\_files/paper/2023/hash/8540fba4abdc7f9f7a7b1cc6cd60e409-Abstract-Datasets\_and\_Benchmarks.html} {Perception test: {A} diagnostic benchmark for multimodal video models}.
\newblock In \emph{Advances in Neural Information Processing Systems 36: Annual Conference on Neural Information Processing Systems 2023, NeurIPS 2023, New Orleans, LA, USA, December 10 - 16, 2023}.

\bibitem[{Peng et~al.(2022)Peng, Dong, Bao, Ye, and Wei}]{Beitv2}
Zhiliang Peng, Li~Dong, Hangbo Bao, Qixiang Ye, and Furu Wei. 2022.
\newblock \href {https://doi.org/10.48550/ARXIV.2208.06366} {Beit v2: Masked image modeling with vector-quantized visual tokenizers}.
\newblock \emph{CoRR}, abs/2208.06366.

\bibitem[{Perazzi et~al.(2016)Perazzi, Pont{-}Tuset, McWilliams, Gool, Gross, and Sorkine{-}Hornung}]{DAVIS}
Federico Perazzi, Jordi Pont{-}Tuset, Brian McWilliams, Luc~Van Gool, Markus~H. Gross, and Alexander Sorkine{-}Hornung. 2016.
\newblock \href {https://doi.org/10.1109/CVPR.2016.85} {A benchmark dataset and evaluation methodology for video object segmentation}.
\newblock In \emph{2016 {IEEE} Conference on Computer Vision and Pattern Recognition, {CVPR} 2016, Las Vegas, NV, USA, June 27-30, 2016}, pages 724--732. {IEEE} Computer Society.

\bibitem[{Phan et~al.(2013)Phan, Shivakumara, Tian, and Tan}]{SVTP}
Trung~Quy Phan, Palaiahnakote Shivakumara, Shangxuan Tian, and Chew~Lim Tan. 2013.
\newblock \href {https://doi.org/10.1109/ICCV.2013.76} {Recognizing text with perspective distortion in natural scenes}.
\newblock In \emph{{IEEE} International Conference on Computer Vision, {ICCV} 2013, Sydney, Australia, December 1-8, 2013}, pages 569--576. {IEEE} Computer Society.

\bibitem[{Piczak(2015)}]{ecs50}
Karol~J. Piczak. 2015.
\newblock \href {https://doi.org/10.1145/2733373.2806390} {{ESC:} dataset for environmental sound classification}.
\newblock In \emph{Proceedings of the 23rd Annual {ACM} Conference on Multimedia Conference, {MM} '15, Brisbane, Australia, October 26 - 30, 2015}, pages 1015--1018. {ACM}.

\bibitem[{Plummer et~al.(2015)Plummer, Wang, Cervantes, Caicedo, Hockenmaier, and Lazebnik}]{flicker30k}
Bryan~A. Plummer, Liwei Wang, Chris~M. Cervantes, Juan~C. Caicedo, Julia Hockenmaier, and Svetlana Lazebnik. 2015.
\newblock \href {https://doi.org/10.1109/ICCV.2015.303} {Flickr30k entities: Collecting region-to-phrase correspondences for richer image-to-sentence models}.
\newblock In \emph{2015 {IEEE} International Conference on Computer Vision, {ICCV} 2015, Santiago, Chile, December 7-13, 2015}, pages 2641--2649. {IEEE} Computer Society.

\bibitem[{Pratap et~al.(2020)Pratap, Xu, Sriram, Synnaeve, and Collobert}]{MLS}
Vineel Pratap, Qiantong Xu, Anuroop Sriram, Gabriel Synnaeve, and Ronan Collobert. 2020.
\newblock \href {https://doi.org/10.21437/INTERSPEECH.2020-2826} {{MLS:} {A} large-scale multilingual dataset for speech research}.
\newblock In \emph{21st Annual Conference of the International Speech Communication Association, Interspeech 2020, Virtual Event, Shanghai, China, October 25-29, 2020}, pages 2757--2761. {ISCA}.

\bibitem[{Radenovic et~al.(2023)Radenovic, Dubey, Kadian, Mihaylov, Vandenhende, Patel, Wen, Ramanathan, and Mahajan}]{LAION-Cat}
Filip Radenovic, Abhimanyu Dubey, Abhishek Kadian, Todor Mihaylov, Simon Vandenhende, Yash Patel, Yi~Wen, Vignesh Ramanathan, and Dhruv Mahajan. 2023.
\newblock \href {https://doi.org/10.1109/CVPR52729.2023.00673} {Filtering, distillation, and hard negatives for vision-language pre-training}.
\newblock In \emph{{IEEE/CVF} Conference on Computer Vision and Pattern Recognition, {CVPR} 2023, Vancouver, BC, Canada, June 17-24, 2023}, pages 6967--6977. {IEEE}.

\bibitem[{Radford et~al.(2021)Radford, Kim, Hallacy, Ramesh, Goh, Agarwal, Sastry, Askell, Mishkin, Clark, Krueger, and Sutskever}]{clip}
Alec Radford, Jong~Wook Kim, Chris Hallacy, Aditya Ramesh, Gabriel Goh, Sandhini Agarwal, Girish Sastry, Amanda Askell, Pamela Mishkin, Jack Clark, Gretchen Krueger, and Ilya Sutskever. 2021.
\newblock \href {http://proceedings.mlr.press/v139/radford21a.html} {Learning transferable visual models from natural language supervision}.
\newblock In \emph{Proceedings of the 38th International Conference on Machine Learning, {ICML} 2021, 18-24 July 2021, Virtual Event}, volume 139 of \emph{Proceedings of Machine Learning Research}, pages 8748--8763. {PMLR}.

\bibitem[{Radford et~al.(2023)Radford, Kim, Xu, Brockman, McLeavey, and Sutskever}]{Whisper}
Alec Radford, Jong~Wook Kim, Tao Xu, Greg Brockman, Christine McLeavey, and Ilya Sutskever. 2023.
\newblock \href {https://proceedings.mlr.press/v202/radford23a.html} {Robust speech recognition via large-scale weak supervision}.
\newblock In \emph{International Conference on Machine Learning, {ICML} 2023, 23-29 July 2023, Honolulu, Hawaii, {USA}}, volume 202 of \emph{Proceedings of Machine Learning Research}, pages 28492--28518. {PMLR}.

\bibitem[{Raffel et~al.(2020)Raffel, Shazeer, Roberts, Lee, Narang, Matena, Zhou, Li, and Liu}]{T5}
Colin Raffel, Noam Shazeer, Adam Roberts, Katherine Lee, Sharan Narang, Michael Matena, Yanqi Zhou, Wei Li, and Peter~J. Liu. 2020.
\newblock \href {https://jmlr.org/papers/v21/20-074.html} {Exploring the limits of transfer learning with a unified text-to-text transformer}.
\newblock \emph{J. Mach. Learn. Res.}, 21:140:1--140:67.

\bibitem[{Ramesh et~al.(2021)Ramesh, Pavlov, Goh, Gray, Voss, Radford, Chen, and Sutskever}]{Dalle1}
Aditya Ramesh, Mikhail Pavlov, Gabriel Goh, Scott Gray, Chelsea Voss, Alec Radford, Mark Chen, and Ilya Sutskever. 2021.
\newblock \href {https://arxiv.org/abs/2102.12092} {Zero-shot text-to-image generation}.
\newblock \emph{CoRR}, abs/2102.12092.

\bibitem[{Ranftl et~al.(2021)Ranftl, Bochkovskiy, and Koltun}]{DPT}
Ren{\'{e}} Ranftl, Alexey Bochkovskiy, and Vladlen Koltun. 2021.
\newblock \href {https://doi.org/10.1109/ICCV48922.2021.01196} {Vision transformers for dense prediction}.
\newblock In \emph{2021 {IEEE/CVF} International Conference on Computer Vision, {ICCV} 2021, Montreal, QC, Canada, October 10-17, 2021}, pages 12159--12168. {IEEE}.

\bibitem[{Reid et~al.(2024)Reid, Savinov, Teplyashin, Lepikhin, Lillicrap, Alayrac, Soricut, Lazaridou, Firat, Schrittwieser, Antonoglou, Anil, Borgeaud, Dai, Millican, Dyer, Glaese, Sottiaux, Lee, Viola, Reynolds, Xu, Molloy, Chen, Isard, Barham, Hennigan, McIlroy, Johnson, Schalkwyk, Collins, Rutherford, Moreira, Ayoub, Goel, Meyer, Thornton, Yang, Michalewski, Abbas, Schucher, Anand, Ives, Keeling, Lenc, Haykal, Shakeri, Shyam, Chowdhery, Ring, Spencer, Sezener, and et~al.}]{Gemini1.5}
Machel Reid, Nikolay Savinov, Denis Teplyashin, Dmitry Lepikhin, Timothy~P. Lillicrap, Jean{-}Baptiste Alayrac, Radu Soricut, Angeliki Lazaridou, Orhan Firat, Julian Schrittwieser, Ioannis Antonoglou, Rohan Anil, Sebastian Borgeaud, Andrew~M. Dai, Katie Millican, Ethan Dyer, Mia Glaese, Thibault Sottiaux, Benjamin Lee, Fabio Viola, Malcolm Reynolds, Yuanzhong Xu, James Molloy, Jilin Chen, Michael Isard, Paul Barham, Tom Hennigan, Ross McIlroy, Melvin Johnson, Johan Schalkwyk, Eli Collins, Eliza Rutherford, Erica Moreira, Kareem Ayoub, Megha Goel, Clemens Meyer, Gregory Thornton, Zhen Yang, Henryk Michalewski, Zaheer Abbas, Nathan Schucher, Ankesh Anand, Richard Ives, James Keeling, Karel Lenc, Salem Haykal, Siamak Shakeri, Pranav Shyam, Aakanksha Chowdhery, Roman Ring, Stephen Spencer, Eren Sezener, and et~al. 2024.
\newblock \href {https://doi.org/10.48550/ARXIV.2403.05530} {Gemini 1.5: Unlocking multimodal understanding across millions of tokens of context}.
\newblock \emph{CoRR}, abs/2403.05530.

\bibitem[{Ren et~al.(2024)Ren, Wang, Yi, Xu, Tao, Zhang, and Zhou}]{ticodec}
Yong Ren, Tao Wang, Jiangyan Yi, Le~Xu, Jianhua Tao, Chu~Yuan Zhang, and Junzuo Zhou. 2024.
\newblock \href {https://doi.org/10.1109/ICASSP48485.2024.10448454} {Fewer-token neural speech codec with time-invariant codes}.
\newblock In \emph{{IEEE} International Conference on Acoustics, Speech and Signal Processing, {ICASSP} 2024, Seoul, Republic of Korea, April 14-19, 2024}, pages 12737--12741. {IEEE}.

\bibitem[{Risnumawan et~al.(2014)Risnumawan, Shivakumara, Chan, and Tan}]{CUTE80}
Anhar Risnumawan, Palaiahnakote Shivakumara, Chee~Seng Chan, and Chew~Lim Tan. 2014.
\newblock \href {https://doi.org/10.1016/J.ESWA.2014.07.008} {A robust arbitrary text detection system for natural scene images}.
\newblock \emph{Expert Syst. Appl.}, 41(18):8027--8048.

\bibitem[{Rombach et~al.(2022)Rombach, Blattmann, Lorenz, Esser, and Ommer}]{stablediffusion}
Robin Rombach, Andreas Blattmann, Dominik Lorenz, Patrick Esser, and Bj{\"{o}}rn Ommer. 2022.
\newblock \href {https://doi.org/10.1109/CVPR52688.2022.01042} {High-resolution image synthesis with latent diffusion models}.
\newblock In \emph{{IEEE/CVF} Conference on Computer Vision and Pattern Recognition, {CVPR} 2022, New Orleans, LA, USA, June 18-24, 2022}, pages 10674--10685. {IEEE}.

\bibitem[{Ruiz et~al.(2023)Ruiz, Li, Jampani, Pritch, Rubinstein, and Aberman}]{dreambench}
Nataniel Ruiz, Yuanzhen Li, Varun Jampani, Yael Pritch, Michael Rubinstein, and Kfir Aberman. 2023.
\newblock \href {https://doi.org/10.1109/CVPR52729.2023.02155} {Dreambooth: Fine tuning text-to-image diffusion models for subject-driven generation}.
\newblock In \emph{{IEEE/CVF} Conference on Computer Vision and Pattern Recognition, {CVPR} 2023, Vancouver, BC, Canada, June 17-24, 2023}, pages 22500--22510. {IEEE}.

\bibitem[{RVC-Boss()}]{gpt-sovits}
RVC-Boss.
\newblock Gpt-sovits.
\newblock \url{https://github.com/RVC-Boss/GPT-SoVITS}.
\newblock Accessed: January 3, 2025.

\bibitem[{Sanabria et~al.(2018)Sanabria, Caglayan, Palaskar, Elliott, Barrault, Specia, and Metze}]{how2}
Ramon Sanabria, Ozan Caglayan, Shruti Palaskar, Desmond Elliott, Lo{\"{\i}}c Barrault, Lucia Specia, and Florian Metze. 2018.
\newblock \href {https://arxiv.org/abs/1811.00347} {How2: {A} large-scale dataset for multimodal language understanding}.
\newblock \emph{CoRR}, abs/1811.00347.

\bibitem[{Schuhmann et~al.(2022)Schuhmann, Beaumont, Vencu, Gordon, Wightman, Cherti, Coombes, Katta, Mullis, Wortsman, Schramowski, Kundurthy, Crowson, Schmidt, Kaczmarczyk, and Jitsev}]{LAION-5B}
Christoph Schuhmann, Romain Beaumont, Richard Vencu, Cade Gordon, Ross Wightman, Mehdi Cherti, Theo Coombes, Aarush Katta, Clayton Mullis, Mitchell Wortsman, Patrick Schramowski, Srivatsa Kundurthy, Katherine Crowson, Ludwig Schmidt, Robert Kaczmarczyk, and Jenia Jitsev. 2022.
\newblock \href {http://papers.nips.cc/paper\_files/paper/2022/hash/a1859debfb3b59d094f3504d5ebb6c25-Abstract-Datasets\_and\_Benchmarks.html} {{LAION-5B:} an open large-scale dataset for training next generation image-text models}.
\newblock In \emph{Advances in Neural Information Processing Systems 35: Annual Conference on Neural Information Processing Systems 2022, NeurIPS 2022, New Orleans, LA, USA, November 28 - December 9, 2022}.

\bibitem[{Schuhmann et~al.(2022b{(\natexlab{a})})Schuhmann, Köpf, Vencu, Coombes, and Beaumont}]{laion-aesthetics}
Christoph Schuhmann, Andreas Köpf, Richard Vencu, Theo Coombes, and Romain Beaumont. 2022b{(\natexlab{a})}.
\newblock \href {https://laion.ai/blog/laion-aesthetics/} {Laion-aesthetics}.

\bibitem[{Schuhmann et~al.(2022b{(\natexlab{b})})Schuhmann, Köpf, Vencu, Coombes, and Beaumont}]{laioncoco}
Christoph Schuhmann, Andreas Köpf, Richard Vencu, Theo Coombes, and Romain Beaumont. 2022b{(\natexlab{b})}.
\newblock \href {https://laion.ai/blog/laion-coco/} {Laion coco: 600m synthetic captions from laion2b-en}.

\bibitem[{Schuhmann et~al.(2021)Schuhmann, Vencu, Beaumont, Kaczmarczyk, Mullis, Katta, Coombes, Jitsev, and Komatsuzaki}]{laion-400m}
Christoph Schuhmann, Richard Vencu, Romain Beaumont, Robert Kaczmarczyk, Clayton Mullis, Aarush Katta, Theo Coombes, Jenia Jitsev, and Aran Komatsuzaki. 2021.
\newblock \href {https://arxiv.org/abs/2111.02114} {{LAION-400M:} open dataset of clip-filtered 400 million image-text pairs}.
\newblock \emph{CoRR}, abs/2111.02114.

\bibitem[{Schwenk et~al.(2022)Schwenk, Khandelwal, Clark, Marino, and Mottaghi}]{A-OKVQA}
Dustin Schwenk, Apoorv Khandelwal, Christopher Clark, Kenneth Marino, and Roozbeh Mottaghi. 2022.
\newblock \href {https://doi.org/10.1007/978-3-031-20074-8\_9} {{A-OKVQA:} {A} benchmark for visual question answering using world knowledge}.
\newblock In \emph{Computer Vision - {ECCV} 2022 - 17th European Conference, Tel Aviv, Israel, October 23-27, 2022, Proceedings, Part {VIII}}, volume 13668 of \emph{Lecture Notes in Computer Science}, pages 146--162. Springer.

\bibitem[{Share(2024)}]{sharegemini}
Share. 2024.
\newblock \href {https://github.com/Share14/ShareGemini} {Sharegemini: Scaling up video caption data for multimodal large language models}.

\bibitem[{Sharma et~al.(2018)Sharma, Ding, Goodman, and Soricut}]{cc3m}
Piyush Sharma, Nan Ding, Sebastian Goodman, and Radu Soricut. 2018.
\newblock \href {https://doi.org/10.18653/V1/P18-1238} {Conceptual captions: {A} cleaned, hypernymed, image alt-text dataset for automatic image captioning}.
\newblock In \emph{Proceedings of the 56th Annual Meeting of the Association for Computational Linguistics, {ACL} 2018, Melbourne, Australia, July 15-20, 2018, Volume 1: Long Papers}, pages 2556--2565. Association for Computational Linguistics.

\bibitem[{Shi et~al.(2024)Shi, Zhu, Hu, Guo, Li, and Wu}]{Med-2E3}
Yiming Shi, Xun Zhu, Ying Hu, Chenyi Guo, Miao Li, and Ji~Wu. 2024.
\newblock \href {https://doi.org/10.48550/ARXIV.2411.12783} {Med-2e3: {A} 2d-enhanced 3d medical multimodal large language model}.
\newblock \emph{CoRR}, abs/2411.12783.

\bibitem[{Shi et~al.(2020)Shi, Zhou, Qiu, and Zhu}]{dalle3}
Zhan Shi, Xu~Zhou, Xipeng Qiu, and Xiaodan Zhu. 2020.
\newblock \href {https://arxiv.org/abs/2006.11807} {Improving image captioning with better use of captions}.
\newblock \emph{CoRR}, abs/2006.11807.

\bibitem[{Shu et~al.(2023)Shu, Zhang, Jiang, and Xie}]{avllm}
Fangxun Shu, Lei Zhang, Hao Jiang, and Cihang Xie. 2023.
\newblock \href {https://doi.org/10.48550/ARXIV.2312.06720} {Audio-visual {LLM} for video understanding}.
\newblock \emph{CoRR}, abs/2312.06720.

\bibitem[{Shukor et~al.(2023)Shukor, Dancette, and Cord}]{ePalm}
Mustafa Shukor, Corentin Dancette, and Matthieu Cord. 2023.
\newblock \href {https://doi.org/10.1109/ICCV51070.2023.02016} {ep-alm: Efficient perceptual augmentation of language models}.
\newblock In \emph{{IEEE/CVF} International Conference on Computer Vision, {ICCV} 2023, Paris, France, October 1-6, 2023}, pages 21999--22012. {IEEE}.

\bibitem[{Sigurdsson et~al.(2016)Sigurdsson, Varol, Wang, Farhadi, Laptev, and Gupta}]{charades-sta}
Gunnar~A. Sigurdsson, G{\"{u}}l Varol, Xiaolong Wang, Ali Farhadi, Ivan Laptev, and Abhinav Gupta. 2016.
\newblock \href {https://doi.org/10.1007/978-3-319-46448-0\_31} {Hollywood in homes: Crowdsourcing data collection for activity understanding}.
\newblock In \emph{Computer Vision - {ECCV} 2016 - 14th European Conference, Amsterdam, The Netherlands, October 11-14, 2016, Proceedings, Part {I}}, volume 9905 of \emph{Lecture Notes in Computer Science}, pages 510--526. Springer.

\bibitem[{Silberman et~al.(2012)Silberman, Hoiem, Kohli, and Fergus}]{nyuv2}
Nathan Silberman, Derek Hoiem, Pushmeet Kohli, and Rob Fergus. 2012.
\newblock \href {https://doi.org/10.1007/978-3-642-33715-4\_54} {Indoor segmentation and support inference from {RGBD} images}.
\newblock In \emph{Computer Vision - {ECCV} 2012 - 12th European Conference on Computer Vision, Florence, Italy, October 7-13, 2012, Proceedings, Part {V}}, volume 7576 of \emph{Lecture Notes in Computer Science}, pages 746--760. Springer.

\bibitem[{Singh et~al.(2019)Singh, Natarajan, Shah, Jiang, Chen, Batra, Parikh, and Rohrbach}]{TextVQA}
Amanpreet Singh, Vivek Natarajan, Meet Shah, Yu~Jiang, Xinlei Chen, Dhruv Batra, Devi Parikh, and Marcus Rohrbach. 2019.
\newblock \href {https://doi.org/10.1109/CVPR.2019.00851} {Towards {VQA} models that can read}.
\newblock In \emph{{IEEE} Conference on Computer Vision and Pattern Recognition, {CVPR} 2019, Long Beach, CA, USA, June 16-20, 2019}, pages 8317--8326. Computer Vision Foundation / {IEEE}.

\bibitem[{Siuzdak et~al.(2024)Siuzdak, Gr{\"{o}}tschla, and Lanzend{\"{o}}rfer}]{SNAC}
Hubert Siuzdak, Florian Gr{\"{o}}tschla, and Luca~A. Lanzend{\"{o}}rfer. 2024.
\newblock \href {https://doi.org/10.48550/ARXIV.2410.14411} {{SNAC:} multi-scale neural audio codec}.
\newblock \emph{CoRR}, abs/2410.14411.

\bibitem[{Song et~al.(2015)Song, Lichtenberg, and Xiao}]{sun-rgb-d}
Shuran Song, Samuel~P. Lichtenberg, and Jianxiong Xiao. 2015.
\newblock \href {https://doi.org/10.1109/CVPR.2015.7298655} {{SUN} {RGB-D:} {A} {RGB-D} scene understanding benchmark suite}.
\newblock In \emph{{IEEE} Conference on Computer Vision and Pattern Recognition, {CVPR} 2015, Boston, MA, USA, June 7-12, 2015}, pages 567--576. {IEEE} Computer Society.

\bibitem[{Soomro et~al.(2012)Soomro, Zamir, and Shah}]{ucf-101}
Khurram Soomro, Amir~Roshan Zamir, and Mubarak Shah. 2012.
\newblock \href {https://arxiv.org/abs/1212.0402} {{UCF101:} {A} dataset of 101 human actions classes from videos in the wild}.
\newblock \emph{CoRR}, abs/1212.0402.

\bibitem[{Su et~al.(2023)Su, Lan, Li, Xu, Wang, and Cai}]{pandagpt}
Yixuan Su, Tian Lan, Huayang Li, Jialu Xu, Yan Wang, and Deng Cai. 2023.
\newblock \href {https://doi.org/10.48550/ARXIV.2305.16355} {Pandagpt: One model to instruction-follow them all}.
\newblock \emph{CoRR}, abs/2305.16355.

\bibitem[{Sun et~al.(2024{\natexlab{a}})Sun, Yu, Tang, Chen, Tan, Li, Lu, Ma, Wang, and Zhang}]{video-SALMONN}
Guangzhi Sun, Wenyi Yu, Changli Tang, Xianzhao Chen, Tian Tan, Wei Li, Lu~Lu, Zejun Ma, Yuxuan Wang, and Chao Zhang. 2024{\natexlab{a}}.
\newblock \href {https://openreview.net/forum?id=nYsh5GFIqX} {video-salmonn: Speech-enhanced audio-visual large language models}.
\newblock In \emph{Forty-first International Conference on Machine Learning, {ICML} 2024, Vienna, Austria, July 21-27, 2024}. OpenReview.net.

\bibitem[{Sun et~al.(2023{\natexlab{a}})Sun, Yu, Tang, Chen, Tan, Li, Lu, Ma, and Zhang}]{favor}
Guangzhi Sun, Wenyi Yu, Changli Tang, Xianzhao Chen, Tian Tan, Wei Li, Lu~Lu, Zejun Ma, and Chao Zhang. 2023{\natexlab{a}}.
\newblock \href {https://doi.org/10.48550/ARXIV.2310.05863} {Fine-grained audio-visual joint representations for multimodal large language models}.
\newblock \emph{CoRR}, abs/2310.05863.

\bibitem[{Sun et~al.(2023{\natexlab{b}})Sun, Pan, Ge, Li, Duan, Wu, Zhang, Zhou, Qin, Wang, Dai, Qiao, Wang, and Li}]{Journeydb}
Keqiang Sun, Junting Pan, Yuying Ge, Hao Li, Haodong Duan, Xiaoshi Wu, Renrui Zhang, Aojun Zhou, Zipeng Qin, Yi~Wang, Jifeng Dai, Yu~Qiao, Limin Wang, and Hongsheng Li. 2023{\natexlab{b}}.
\newblock \href {http://papers.nips.cc/paper\_files/paper/2023/hash/9bc59aff4685e39e1a8175d5303248a1-Abstract-Datasets\_and\_Benchmarks.html} {Journeydb: {A} benchmark for generative image understanding}.
\newblock In \emph{Advances in Neural Information Processing Systems 36: Annual Conference on Neural Information Processing Systems 2023, NeurIPS 2023, New Orleans, LA, USA, December 10 - 16, 2023}.

\bibitem[{Sun et~al.(2023{\natexlab{c}})Sun, Lian, Liu, and Tao}]{mae-dfer}
Licai Sun, Zheng Lian, Bin Liu, and Jianhua Tao. 2023{\natexlab{c}}.
\newblock \href {https://doi.org/10.1145/3581783.3612365} {{MAE-DFER:} efficient masked autoencoder for self-supervised dynamic facial expression recognition}.
\newblock In \emph{Proceedings of the 31st {ACM} International Conference on Multimedia, {MM} 2023, Ottawa, ON, Canada, 29 October 2023- 3 November 2023}, pages 6110--6121. {ACM}.

\bibitem[{Sun et~al.(2023{\natexlab{d}})Sun, Lian, Liu, and Tao}]{omnivl}
Licai Sun, Zheng Lian, Bin Liu, and Jianhua Tao. 2023{\natexlab{d}}.
\newblock \href {https://doi.org/10.1145/3581783.3612365} {{MAE-DFER:} efficient masked autoencoder for self-supervised dynamic facial expression recognition}.
\newblock In \emph{Proceedings of the 31st {ACM} International Conference on Multimedia, {MM} 2023, Ottawa, ON, Canada, 29 October 2023- 3 November 2023}, pages 6110--6121. {ACM}.

\bibitem[{Sun et~al.(2024{\natexlab{b}})Sun, Xu, Wu, and Xie}]{autocad}
Luoyi Sun, Xuenan Xu, Mengyue Wu, and Weidi Xie. 2024{\natexlab{b}}.
\newblock \href {https://doi.org/10.1145/3664647.3681472} {Auto-acd: {A} large-scale dataset for audio-language representation learning}.
\newblock In \emph{Proceedings of the 32nd {ACM} International Conference on Multimedia, {MM} 2024, Melbourne, VIC, Australia, 28 October 2024 - 1 November 2024}, pages 5025--5034. {ACM}.

\bibitem[{Sun et~al.(2023{\natexlab{e}})Sun, Fang, Wu, Wang, and Cao}]{evaclip}
Quan Sun, Yuxin Fang, Ledell Wu, Xinlong Wang, and Yue Cao. 2023{\natexlab{e}}.
\newblock \href {https://doi.org/10.48550/ARXIV.2303.15389} {{EVA-CLIP:} improved training techniques for {CLIP} at scale}.
\newblock \emph{CoRR}, abs/2303.15389.

\bibitem[{Sun et~al.(2024{\natexlab{c}})Sun, Yu, Cui, Zhang, Zhang, Wang, Gao, Liu, Huang, and Wang}]{emu}
Quan Sun, Qiying Yu, Yufeng Cui, Fan Zhang, Xiaosong Zhang, Yueze Wang, Hongcheng Gao, Jingjing Liu, Tiejun Huang, and Xinlong Wang. 2024{\natexlab{c}}.
\newblock \href {https://openreview.net/forum?id=mL8Q9OOamV} {Emu: Generative pretraining in multimodality}.
\newblock In \emph{The Twelfth International Conference on Learning Representations, {ICLR} 2024, Vienna, Austria, May 7-11, 2024}. OpenReview.net.

\bibitem[{Szot et~al.(2024{\natexlab{a}})Szot, Mazoure, Agrawal, Hjelm, Kira, and Toshev}]{groundaction}
Andrew Szot, Bogdan Mazoure, Harsh Agrawal, R.~Devon Hjelm, Zsolt Kira, and Alexander Toshev. 2024{\natexlab{a}}.
\newblock \href {https://doi.org/10.48550/ARXIV.2406.07904} {Grounding multimodal large language models in actions}.
\newblock \emph{CoRR}, abs/2406.07904.

\bibitem[{Szot et~al.(2024{\natexlab{b}})Szot, Mazoure, Attia, Timofeev, Agrawal, Hjelm, Gan, Kira, and Toshev}]{GMA}
Andrew Szot, Bogdan Mazoure, Omar Attia, Aleksei Timofeev, Harsh Agrawal, Devon Hjelm, Zhe Gan, Zsolt Kira, and Alexander Toshev. 2024{\natexlab{b}}.
\newblock \href {https://arxiv.org/abs/2412.08442} {From multimodal llms to generalist embodied agents: Methods and lessons}.
\newblock \emph{Preprint}, arXiv:2412.08442.

\bibitem[{Tang et~al.(2024{\natexlab{a}})Tang, Shimada, Bi, and Xu}]{avicuna}
Yunlong Tang, Daiki Shimada, Jing Bi, and Chenliang Xu. 2024{\natexlab{a}}.
\newblock \href {https://doi.org/10.48550/ARXIV.2403.16276} {Avicuna: Audio-visual {LLM} with interleaver and context-boundary alignment for temporal referential dialogue}.
\newblock \emph{CoRR}, abs/2403.16276.

\bibitem[{Tang et~al.(2024{\natexlab{b}})Tang, Yang, Khademi, Liu, Zhu, and Bansal}]{codi2}
Zineng Tang, Ziyi Yang, Mahmoud Khademi, Yang Liu, Chenguang Zhu, and Mohit Bansal. 2024{\natexlab{b}}.
\newblock \href {https://doi.org/10.1109/CVPR52733.2024.02589} {Codi-2: In-context, interleaved, and interactive any-to-any generation}.
\newblock In \emph{{IEEE/CVF} Conference on Computer Vision and Pattern Recognition, {CVPR} 2024, Seattle, WA, USA, June 16-22, 2024}, pages 27415--27424. {IEEE}.

\bibitem[{Tang et~al.(2023)Tang, Yang, Zhu, Zeng, and Bansal}]{CoDi}
Zineng Tang, Ziyi Yang, Chenguang Zhu, Michael Zeng, and Mohit Bansal. 2023.
\newblock \href {http://papers.nips.cc/paper\_files/paper/2023/hash/33edf072fe44f19079d66713a1831550-Abstract-Conference.html} {Any-to-any generation via composable diffusion}.
\newblock In \emph{Advances in Neural Information Processing Systems 36: Annual Conference on Neural Information Processing Systems 2023, NeurIPS 2023, New Orleans, LA, USA, December 10 - 16, 2023}.

\bibitem[{Tian et~al.(2020)Tian, Li, and Xu}]{LLP}
Yapeng Tian, Dingzeyu Li, and Chenliang Xu. 2020.
\newblock \href {https://doi.org/10.1007/978-3-030-58580-8\_26} {Unified multisensory perception: Weakly-supervised audio-visual video parsing}.
\newblock In \emph{Computer Vision - {ECCV} 2020 - 16th European Conference, Glasgow, UK, August 23-28, 2020, Proceedings, Part {III}}, volume 12348 of \emph{Lecture Notes in Computer Science}, pages 436--454. Springer.

\bibitem[{Tong et~al.(2022)Tong, Song, Wang, and Wang}]{videomae}
Zhan Tong, Yibing Song, Jue Wang, and Limin Wang. 2022.
\newblock \href {http://papers.nips.cc/paper\_files/paper/2022/hash/416f9cb3276121c42eebb86352a4354a-Abstract-Conference.html} {Videomae: Masked autoencoders are data-efficient learners for self-supervised video pre-training}.
\newblock In \emph{Advances in Neural Information Processing Systems 35: Annual Conference on Neural Information Processing Systems 2022, NeurIPS 2022, New Orleans, LA, USA, November 28 - December 9, 2022}.

\bibitem[{Touvron et~al.(2023)Touvron, Lavril, Izacard, Martinet, Lachaux, Lacroix, Rozi{\`{e}}re, Goyal, Hambro, Azhar, Rodriguez, Joulin, Grave, and Lample}]{llama}
Hugo Touvron, Thibaut Lavril, Gautier Izacard, Xavier Martinet, Marie{-}Anne Lachaux, Timoth{\'{e}}e Lacroix, Baptiste Rozi{\`{e}}re, Naman Goyal, Eric Hambro, Faisal Azhar, Aur{\'{e}}lien Rodriguez, Armand Joulin, Edouard Grave, and Guillaume Lample. 2023.
\newblock \href {https://doi.org/10.48550/ARXIV.2302.13971} {Llama: Open and efficient foundation language models}.
\newblock \emph{CoRR}, abs/2302.13971.

\bibitem[{Urbanek et~al.(2024)Urbanek, Bordes, Astolfi, Williamson, Sharma, and Romero{-}Soriano}]{DCI}
Jack Urbanek, Florian Bordes, Pietro Astolfi, Mary Williamson, Vasu Sharma, and Adriana Romero{-}Soriano. 2024.
\newblock \href {https://doi.org/10.1109/CVPR52733.2024.02521} {A picture is worth more than 77 text tokens: Evaluating clip-style models on dense captions}.
\newblock In \emph{{IEEE/CVF} Conference on Computer Vision and Pattern Recognition, {CVPR} 2024, Seattle, WA, USA, June 16-22, 2024}, pages 26690--26699. {IEEE}.

\bibitem[{Veaux et~al.(2017)Veaux, Yamagishi, MacDonald et~al.}]{VCTK}
Christophe Veaux, Junichi Yamagishi, Kirsten MacDonald, et~al. 2017.
\newblock Cstr vctk corpus: English multi-speaker corpus for cstr voice cloning toolkit.
\newblock \emph{CSTR}, 6:15.

\bibitem[{Veit et~al.(2016)Veit, Matera, Neumann, Matas, and Belongie}]{COCO-Text}
Andreas Veit, Tomas Matera, Luk{\'{a}}s Neumann, Jiri Matas, and Serge~J. Belongie. 2016.
\newblock \href {https://arxiv.org/abs/1601.07140} {Coco-text: Dataset and benchmark for text detection and recognition in natural images}.
\newblock \emph{CoRR}, abs/1601.07140.

\bibitem[{Wang et~al.(2024{\natexlab{a}})Wang, Jiang, Liu, Ma, Zhang, Pan, Liu, Gu, Xia, Li, Zhang, Wu, Liu, Zhong, Ge, Zhang, Qiang, Hu, Jiang, Zhang, Zhang, Shen, Liu, and Zhang}]{survey_j}
Jiaqi Wang, Hanqi Jiang, Yiheng Liu, Chong Ma, Xu~Zhang, Yi~Pan, Mengyuan Liu, Peiran Gu, Sichen Xia, Wenjun Li, Yutong Zhang, Zihao Wu, Zhengliang Liu, Tianyang Zhong, Bao Ge, Tuo Zhang, Ning Qiang, Xintao Hu, Xi~Jiang, Xin Zhang, Wei Zhang, Dinggang Shen, Tianming Liu, and Shu Zhang. 2024{\natexlab{a}}.
\newblock \href {https://doi.org/10.48550/ARXIV.2408.01319} {A comprehensive review of multimodal large language models: Performance and challenges across different tasks}.
\newblock \emph{CoRR}, abs/2408.01319.

\bibitem[{Wang et~al.(2023{\natexlab{a}})Wang, Yuan, Chen, Zhang, Wang, and Zhang}]{ModelScopeT2V}
Jiuniu Wang, Hangjie Yuan, Dayou Chen, Yingya Zhang, Xiang Wang, and Shiwei Zhang. 2023{\natexlab{a}}.
\newblock \href {https://doi.org/10.48550/ARXIV.2308.06571} {Modelscope text-to-video technical report}.
\newblock \emph{CoRR}, abs/2308.06571.

\bibitem[{Wang et~al.(2011)Wang, Babenko, and Belongie}]{SVT}
Kai Wang, Boris Babenko, and Serge~J. Belongie. 2011.
\newblock \href {https://doi.org/10.1109/ICCV.2011.6126402} {End-to-end scene text recognition}.
\newblock In \emph{{IEEE} International Conference on Computer Vision, {ICCV} 2011, Barcelona, Spain, November 6-13, 2011}, pages 1457--1464. {IEEE} Computer Society.

\bibitem[{Wang et~al.(2024{\natexlab{b}})Wang, Bai, Tan, Wang, Fan, Bai, Chen, Liu, Wang, Ge, Fan, Dang, Du, Ren, Men, Liu, Zhou, Zhou, and Lin}]{qwen2vl}
Peng Wang, Shuai Bai, Sinan Tan, Shijie Wang, Zhihao Fan, Jinze Bai, Keqin Chen, Xuejing Liu, Jialin Wang, Wenbin Ge, Yang Fan, Kai Dang, Mengfei Du, Xuancheng Ren, Rui Men, Dayiheng Liu, Chang Zhou, Jingren Zhou, and Junyang Lin. 2024{\natexlab{b}}.
\newblock \href {https://doi.org/10.48550/ARXIV.2409.12191} {Qwen2-vl: Enhancing vision-language model's perception of the world at any resolution}.
\newblock \emph{CoRR}, abs/2409.12191.

\bibitem[{Wang et~al.(2023{\natexlab{b}})Wang, Xie, Hu, Zou, Fan, Tong, Wen, Wu, Deng, Li, Tian, Lu, Zhu, Wang, Qiao, and Dai}]{drivemlm}
Wenhai Wang, Jiangwei Xie, Chuanyang Hu, Haoming Zou, Jianan Fan, Wenwen Tong, Yang Wen, Silei Wu, Hanming Deng, Zhiqi Li, Hao Tian, Lewei Lu, Xizhou Zhu, Xiaogang Wang, Yu~Qiao, and Jifeng Dai. 2023{\natexlab{b}}.
\newblock \href {https://doi.org/10.48550/ARXIV.2312.09245} {Drivemlm: Aligning multi-modal large language models with behavioral planning states for autonomous driving}.
\newblock \emph{CoRR}, abs/2312.09245.

\bibitem[{Wang et~al.(2019)Wang, Wu, Chen, Li, Wang, and Wang}]{vatex}
Xin Wang, Jiawei Wu, Junkun Chen, Lei Li, Yuan{-}Fang Wang, and William~Yang Wang. 2019.
\newblock \href {https://doi.org/10.1109/ICCV.2019.00468} {Vatex: {A} large-scale, high-quality multilingual dataset for video-and-language research}.
\newblock In \emph{2019 {IEEE/CVF} International Conference on Computer Vision, {ICCV} 2019, Seoul, Korea (South), October 27 - November 2, 2019}, pages 4580--4590. {IEEE}.

\bibitem[{Wang et~al.(2024{\natexlab{c}})Wang, Zhuang, and Wu}]{modaverse}
Xinyu Wang, Bohan Zhuang, and Qi~Wu. 2024{\natexlab{c}}.
\newblock \href {https://doi.org/10.1109/CVPR52733.2024.02512} {Modaverse: Efficiently transforming modalities with llms}.
\newblock In \emph{{IEEE/CVF} Conference on Computer Vision and Pattern Recognition, {CVPR} 2024, Seattle, WA, USA, June 16-22, 2024}, pages 26596--26606. {IEEE}.

\bibitem[{Wang et~al.(2024{\natexlab{d}})Wang, He, Li, Li, Yu, Ma, Li, Chen, Chen, Wang, Luo, Liu, Wang, Wang, and Qiao}]{internvid}
Yi~Wang, Yinan He, Yizhuo Li, Kunchang Li, Jiashuo Yu, Xin Ma, Xinhao Li, Guo Chen, Xinyuan Chen, Yaohui Wang, Ping Luo, Ziwei Liu, Yali Wang, Limin Wang, and Yu~Qiao. 2024{\natexlab{d}}.
\newblock \href {https://openreview.net/forum?id=MLBdiWu4Fw} {Internvid: {A} large-scale video-text dataset for multimodal understanding and generation}.
\newblock In \emph{The Twelfth International Conference on Learning Representations, {ICLR} 2024, Vienna, Austria, May 7-11, 2024}. OpenReview.net.

\bibitem[{Wang et~al.(2024{\natexlab{e}})Wang, Li, Li, Yu, He, Chen, Pei, Zheng, Wang, Shi, Jiang, Li, Xu, Zhang, Huang, Qiao, Wang, and Wang}]{internvideo2}
Yi~Wang, Kunchang Li, Xinhao Li, Jiashuo Yu, Yinan He, Guo Chen, Baoqi Pei, Rongkun Zheng, Zun Wang, Yansong Shi, Tianxiang Jiang, Songze Li, Jilan Xu, Hongjie Zhang, Yifei Huang, Yu~Qiao, Yali Wang, and Limin Wang. 2024{\natexlab{e}}.
\newblock \href {https://doi.org/10.1007/978-3-031-73013-9\_23} {Internvideo2: Scaling foundation models for multimodal video understanding}.
\newblock In \emph{Computer Vision - {ECCV} 2024 - 18th European Conference, Milan, Italy, September 29-October 4, 2024, Proceedings, Part {LXXXV}}, volume 15143 of \emph{Lecture Notes in Computer Science}, pages 396--416. Springer.

\bibitem[{Wang et~al.(2024{\natexlab{f}})Wang, Zhu, Xu, Zhou, Liu, Zhang, Wang, Shi, Li, Li, Que, Zhang, Zhang, Zhang, Xu, Fu, and Huang}]{MIO}
Zekun Wang, King Zhu, Chunpu Xu, Wangchunshu Zhou, Jiaheng Liu, Yibo Zhang, Jiashuo Wang, Ning Shi, Siyu Li, Yizhi Li, Haoran Que, Zhaoxiang Zhang, Yuanxing Zhang, Ge~Zhang, Ke~Xu, Jie Fu, and Wenhao Huang. 2024{\natexlab{f}}.
\newblock \href {https://arxiv.org/abs/2409.17692} {Mio: A foundation model on multimodal tokens}.
\newblock \emph{Preprint}, arXiv:2409.17692.

\bibitem[{Wei et~al.(2024)Wei, Yuan, Li, Hu, Gan, and Ding}]{occllama}
Julong Wei, Shanshuai Yuan, Pengfei Li, Qingda Hu, Zhongxue Gan, and Wenchao Ding. 2024.
\newblock \href {https://doi.org/10.48550/ARXIV.2409.03272} {Occllama: An occupancy-language-action generative world model for autonomous driving}.
\newblock \emph{CoRR}, abs/2409.03272.

\bibitem[{Wu et~al.(2021)Wu, Yu, Chen, Tenenbaum, and Gan}]{star}
Bo~Wu, Shoubin Yu, Zhenfang Chen, Josh Tenenbaum, and Chuang Gan. 2021.
\newblock \href {https://datasets-benchmarks-proceedings.neurips.cc/paper/2021/hash/5ef059938ba799aaa845e1c2e8a762bd-Abstract-round2.html} {{STAR:} {A} benchmark for situated reasoning in real-world videos}.
\newblock In \emph{Proceedings of the Neural Information Processing Systems Track on Datasets and Benchmarks 1, NeurIPS Datasets and Benchmarks 2021, December 2021, virtual}.

\bibitem[{Wu et~al.(2017)Wu, Zheng, Zhao, Li, Yan, Liang, Wang, Zhou, Lin, Fu, Wang, and Wang}]{ai-Challenger}
Jiahong Wu, He~Zheng, Bo~Zhao, Yixin Li, Baoming Yan, Rui Liang, Wenjia Wang, Shipei Zhou, Guosen Lin, Yanwei Fu, Yizhou Wang, and Yonggang Wang. 2017.
\newblock \href {https://arxiv.org/abs/1711.06475} {{AI} challenger : {A} large-scale dataset for going deeper in image understanding}.
\newblock \emph{CoRR}, abs/1711.06475.

\bibitem[{Wu et~al.(2024{\natexlab{a}})Wu, Wang, Yang, Gan, Liu, Yuan, and Wang}]{GRIT}
Jialian Wu, Jianfeng Wang, Zhengyuan Yang, Zhe Gan, Zicheng Liu, Junsong Yuan, and Lijuan Wang. 2024{\natexlab{a}}.
\newblock \href {https://doi.org/10.1007/978-3-031-72989-8\_12} {Grit: {A} generative region-to-text transformer for object understanding}.
\newblock In \emph{Computer Vision - {ECCV} 2024 - 18th European Conference, Milan, Italy, September 29-October 4, 2024, Proceedings, Part {LXXX}}, volume 15138 of \emph{Lecture Notes in Computer Science}, pages 207--224. Springer.

\bibitem[{Wu et~al.(2023)Wu, Gan, Chen, Wan, and Yu}]{survey:b}
Jiayang Wu, Wensheng Gan, Zefeng Chen, Shicheng Wan, and Philip~S. Yu. 2023.
\newblock \href {https://doi.org/10.1109/BIGDATA59044.2023.10386743} {Multimodal large language models: {A} survey}.
\newblock In \emph{{IEEE} International Conference on Big Data, BigData 2023, Sorrento, Italy, December 15-18, 2023}, pages 2247--2256. {IEEE}.

\bibitem[{Wu et~al.(2024{\natexlab{b}})Wu, Fei, Qu, Ji, and Chua}]{nextgpt}
Shengqiong Wu, Hao Fei, Leigang Qu, Wei Ji, and Tat{-}Seng Chua. 2024{\natexlab{b}}.
\newblock \href {https://openreview.net/forum?id=NZQkumsNlf} {Next-gpt: Any-to-any multimodal {LLM}}.
\newblock In \emph{Forty-first International Conference on Machine Learning, {ICML} 2024, Vienna, Austria, July 21-27, 2024}. OpenReview.net.

\bibitem[{Wu et~al.(2015)Wu, Song, Khosla, Yu, Zhang, Tang, and Xiao}]{ModelNet40}
Zhirong Wu, Shuran Song, Aditya Khosla, Fisher Yu, Linguang Zhang, Xiaoou Tang, and Jianxiong Xiao. 2015.
\newblock \href {https://doi.org/10.1109/CVPR.2015.7298801} {3d shapenets: {A} deep representation for volumetric shapes}.
\newblock In \emph{{IEEE} Conference on Computer Vision and Pattern Recognition, {CVPR} 2015, Boston, MA, USA, June 7-12, 2015}, pages 1912--1920. {IEEE} Computer Society.

\bibitem[{Xiao et~al.(2024)Xiao, Zhou, Liu, Liu, Li, Liu, and Huang}]{survey_medicine_1}
Hanguang Xiao, Feizhong Zhou, Xingyue Liu, Tianqi Liu, Zhipeng Li, Xin Liu, and Xiaoxuan Huang. 2024.
\newblock \href {https://doi.org/10.48550/ARXIV.2405.08603} {A comprehensive survey of large language models and multimodal large language models in medicine}.
\newblock \emph{CoRR}, abs/2405.08603.

\bibitem[{Xiao et~al.(2021)Xiao, Shang, Yao, and Chua}]{nextqa}
Junbin Xiao, Xindi Shang, Angela Yao, and Tat{-}Seng Chua. 2021.
\newblock \href {https://doi.org/10.1109/CVPR46437.2021.00965} {Next-qa: Next phase of question-answering to explaining temporal actions}.
\newblock In \emph{{IEEE} Conference on Computer Vision and Pattern Recognition, {CVPR} 2021, virtual, June 19-25, 2021}, pages 9777--9786. Computer Vision Foundation / {IEEE}.

\bibitem[{Xie and Wu(2024)}]{Mini-Omni2}
Zhifei Xie and Changqiao Wu. 2024.
\newblock \href {https://doi.org/10.48550/ARXIV.2410.11190} {Mini-omni2: Towards open-source gpt-4o with vision, speech and duplex capabilities}.
\newblock \emph{CoRR}, abs/2410.11190.

\bibitem[{Xu et~al.(2017)Xu, Zhao, Xiao, Wu, Zhang, He, and Zhuang}]{msvd}
Dejing Xu, Zhou Zhao, Jun Xiao, Fei Wu, Hanwang Zhang, Xiangnan He, and Yueting Zhuang. 2017.
\newblock \href {https://doi.org/10.1145/3123266.3123427} {Video question answering via gradually refined attention over appearance and motion}.
\newblock In \emph{Proceedings of the 2017 {ACM} on Multimedia Conference, {MM} 2017, Mountain View, CA, USA, October 23-27, 2017}, pages 1645--1653. {ACM}.

\bibitem[{Xu et~al.(2023)Xu, Zhang, Cai, Rezatofighi, Yu, Tao, and Geiger}]{unimatch}
Haofei Xu, Jing Zhang, Jianfei Cai, Hamid Rezatofighi, Fisher Yu, Dacheng Tao, and Andreas Geiger. 2023.
\newblock \href {https://doi.org/10.1109/TPAMI.2023.3298645} {Unifying flow, stereo and depth estimation}.
\newblock \emph{{IEEE} Trans. Pattern Anal. Mach. Intell.}, 45(11):13941--13958.

\bibitem[{Xu et~al.(2024{\natexlab{a}})Xu, Wang, Zhao, Liu, Ma, and Gao}]{Omni-CAD}
Jingwei Xu, Chenyu Wang, Zibo Zhao, Wen Liu, Yi~Ma, and Shenghua Gao. 2024{\natexlab{a}}.
\newblock \href {https://doi.org/10.48550/ARXIV.2411.04954} {{CAD-MLLM:} unifying multimodality-conditioned {CAD} generation with {MLLM}}.
\newblock \emph{CoRR}, abs/2411.04954.

\bibitem[{Xu et~al.(2016)Xu, Mei, Yao, and Rui}]{msrvtts}
Jun Xu, Tao Mei, Ting Yao, and Yong Rui. 2016.
\newblock \href {https://doi.org/10.1109/CVPR.2016.571} {{MSR-VTT:} {A} large video description dataset for bridging video and language}.
\newblock In \emph{2016 {IEEE} Conference on Computer Vision and Pattern Recognition, {CVPR} 2016, Las Vegas, NV, USA, June 27-30, 2016}, pages 5288--5296. {IEEE} Computer Society.

\bibitem[{Xu et~al.(2024{\natexlab{b}})Xu, Wang, Wang, Chen, Pang, and Lin}]{pointllm}
Runsen Xu, Xiaolong Wang, Tai Wang, Yilun Chen, Jiangmiao Pang, and Dahua Lin. 2024{\natexlab{b}}.
\newblock \href {https://doi.org/10.1007/978-3-031-72698-9\_8} {Pointllm: Empowering large language models to understand point clouds}.
\newblock In \emph{Computer Vision - {ECCV} 2024 - 18th European Conference, Milan, Italy, September 29-October 4, 2024, Proceedings, Part {XXV}}, volume 15083 of \emph{Lecture Notes in Computer Science}, pages 131--147. Springer.

\bibitem[{Xue et~al.(2024{\natexlab{a}})Xue, Deng, Gao, and Li}]{auffsion}
Jinlong Xue, Yayue Deng, Yingming Gao, and Ya~Li. 2024{\natexlab{a}}.
\newblock \href {https://doi.org/10.1109/TASLP.2024.3485485} {Auffusion: Leveraging the power of diffusion and large language models for text-to-audio generation}.
\newblock \emph{{IEEE} {ACM} Trans. Audio Speech Lang. Process.}, 32:4700--4712.

\bibitem[{Xue et~al.(2024{\natexlab{b}})Xue, Yu, Zhang, Panagopoulou, Li, Mart{\'{\i}}n{-}Mart{\'{\i}}n, Wu, Xiong, Xu, Niebles, and Savarese}]{ulip2}
Le~Xue, Ning Yu, Shu Zhang, Artemis Panagopoulou, Junnan Li, Roberto Mart{\'{\i}}n{-}Mart{\'{\i}}n, Jiajun Wu, Caiming Xiong, Ran Xu, Juan~Carlos Niebles, and Silvio Savarese. 2024{\natexlab{b}}.
\newblock \href {https://doi.org/10.1109/CVPR52733.2024.02558} {{ULIP-2:} towards scalable multimodal pre-training for 3d understanding}.
\newblock In \emph{{IEEE/CVF} Conference on Computer Vision and Pattern Recognition, {CVPR} 2024, Seattle, WA, USA, June 16-22, 2024}, pages 27081--27091. {IEEE}.

\bibitem[{Yang et~al.(2023{\natexlab{a}})Yang, He, Liu, Chen, Wu, Lin, He, and Ouyang}]{GDMAE}
Honghui Yang, Tong He, Jiaheng Liu, Hua Chen, Boxi Wu, Binbin Lin, Xiaofei He, and Wanli Ouyang. 2023{\natexlab{a}}.
\newblock \href {https://doi.org/10.1109/CVPR52729.2023.00907} {{GD-MAE:} generative decoder for {MAE} pre-training on lidar point clouds}.
\newblock In \emph{{IEEE/CVF} Conference on Computer Vision and Pattern Recognition, {CVPR} 2023, Vancouver, BC, Canada, June 17-24, 2023}, pages 9403--9414. {IEEE}.

\bibitem[{Yang et~al.(2023{\natexlab{b}})Yang, Zhang, Meng, and Zhou}]{teal}
Zhen Yang, Yingxue Zhang, Fandong Meng, and Jie Zhou. 2023{\natexlab{b}}.
\newblock \href {https://doi.org/10.48550/ARXIV.2311.04589} {{TEAL:} tokenize and embed {ALL} for multi-modal large language models}.
\newblock \emph{CoRR}, abs/2311.04589.

\bibitem[{Ye et~al.(2024{\natexlab{a}})Ye, Huang, Lu, Yu, Ping, Tao, Kautz, Han, Xu, Molchanov, and Yin}]{xvila}
Hanrong Ye, De{-}An Huang, Yao Lu, Zhiding Yu, Wei Ping, Andrew Tao, Jan Kautz, Song Han, Dan Xu, Pavlo Molchanov, and Hongxu Yin. 2024{\natexlab{a}}.
\newblock \href {https://doi.org/10.48550/ARXIV.2405.19335} {{X-VILA:} cross-modality alignment for large language model}.
\newblock \emph{CoRR}, abs/2405.19335.

\bibitem[{Ye et~al.(2024{\natexlab{b}})Ye, Yu, Shao, Xie, Torr, and Cao}]{cat}
Qilang Ye, Zitong Yu, Rui Shao, Xinyu Xie, Philip Torr, and Xiaochun Cao. 2024{\natexlab{b}}.
\newblock \href {https://doi.org/10.1007/978-3-031-72684-2\_9} {{CAT:} enhancing multimodal large language model to answer questions in dynamic audio-visual scenarios}.
\newblock In \emph{Computer Vision - {ECCV} 2024 - 18th European Conference, Milan, Italy, September 29-October 4, 2024, Proceedings, Part {X}}, volume 15068 of \emph{Lecture Notes in Computer Science}, pages 146--164. Springer.

\bibitem[{Yin et~al.(2023{\natexlab{a}})Yin, Fu, Zhao, Li, Sun, Xu, and Chen}]{survey:a}
Shukang Yin, Chaoyou Fu, Sirui Zhao, Ke~Li, Xing Sun, Tong Xu, and Enhong Chen. 2023{\natexlab{a}}.
\newblock \href {https://doi.org/10.48550/ARXIV.2306.13549} {A survey on multimodal large language models}.
\newblock \emph{CoRR}, abs/2306.13549.

\bibitem[{Yin et~al.(2023{\natexlab{b}})Yin, Wang, Cao, Shi, Liu, Li, Huang, Wang, Sheng, Bai, Shao, and Ouyang}]{LAMM}
Zhenfei Yin, Jiong Wang, Jianjian Cao, Zhelun Shi, Dingning Liu, Mukai Li, Xiaoshui Huang, Zhiyong Wang, Lu~Sheng, Lei Bai, Jing Shao, and Wanli Ouyang. 2023{\natexlab{b}}.
\newblock \href {http://papers.nips.cc/paper\_files/paper/2023/hash/548a41b9cac6f50dccf7e63e9e1b1b9b-Abstract-Datasets\_and\_Benchmarks.html} {{LAMM:} language-assisted multi-modal instruction-tuning dataset, framework, and benchmark}.
\newblock In \emph{Advances in Neural Information Processing Systems 36: Annual Conference on Neural Information Processing Systems 2023, NeurIPS 2023, New Orleans, LA, USA, December 10 - 16, 2023}.

\bibitem[{Young et~al.(2024)Young, Chen, Li, Huang, Zhang, Zhang, Li, Zhu, Chen, Chang, Yu, Liu, Liu, Yue, Yang, Yang, Yu, Xie, Huang, Hu, Ren, Niu, Nie, Xu, Liu, Wang, Cai, Gu, Liu, and Dai}]{Yi}
Alex Young, Bei Chen, Chao Li, Chengen Huang, Ge~Zhang, Guanwei Zhang, Heng Li, Jiangcheng Zhu, Jianqun Chen, Jing Chang, Kaidong Yu, Peng Liu, Qiang Liu, Shawn Yue, Senbin Yang, Shiming Yang, Tao Yu, Wen Xie, Wenhao Huang, Xiaohui Hu, Xiaoyi Ren, Xinyao Niu, Pengcheng Nie, Yuchi Xu, Yudong Liu, Yue Wang, Yuxuan Cai, Zhenyu Gu, Zhiyuan Liu, and Zonghong Dai. 2024.
\newblock \href {https://doi.org/10.48550/ARXIV.2403.04652} {Yi: Open foundation models by 01.ai}.
\newblock \emph{CoRR}, abs/2403.04652.

\bibitem[{Yu et~al.(2024{\natexlab{a}})Yu, Xiong, Zhang, Diao, Zhuge, Hong, Wang, Lu, He, and Chen}]{PathWeave}
Jiazuo Yu, Haomiao Xiong, Lu~Zhang, Haiwen Diao, Yunzhi Zhuge, Lanqing Hong, Dong Wang, Huchuan Lu, You He, and Long Chen. 2024{\natexlab{a}}.
\newblock \href {http://papers.nips.cc/paper\_files/paper/2024/hash/5942d10ae51b6bd07648e54df07ef9cd-Abstract-Conference.html} {Llms can evolve continually on modality for x-modal reasoning}.
\newblock In \emph{Advances in Neural Information Processing Systems 38: Annual Conference on Neural Information Processing Systems 2024, NeurIPS 2024, Vancouver, BC, Canada, December 10 - 15, 2024}.

\bibitem[{Yu et~al.(2016)Yu, Poirson, Yang, Berg, and Berg}]{refcoco}
Licheng Yu, Patrick Poirson, Shan Yang, Alexander~C. Berg, and Tamara~L. Berg. 2016.
\newblock \href {https://doi.org/10.1007/978-3-319-46475-6\_5} {Modeling context in referring expressions}.
\newblock In \emph{Computer Vision - {ECCV} 2016 - 14th European Conference, Amsterdam, The Netherlands, October 11-14, 2016, Proceedings, Part {II}}, volume 9906 of \emph{Lecture Notes in Computer Science}, pages 69--85. Springer.

\bibitem[{Yu et~al.(2024{\natexlab{b}})Yu, Lezama, Gundavarapu, Versari, Sohn, Minnen, Cheng, Gupta, Gu, Hauptmann, Gong, Yang, Essa, Ross, and Jiang}]{magvit-v2}
Lijun Yu, Jos{\'{e}} Lezama, Nitesh~Bharadwaj Gundavarapu, Luca Versari, Kihyuk Sohn, David Minnen, Yong Cheng, Agrim Gupta, Xiuye Gu, Alexander~G. Hauptmann, Boqing Gong, Ming{-}Hsuan Yang, Irfan Essa, David~A. Ross, and Lu~Jiang. 2024{\natexlab{b}}.
\newblock \href {https://openreview.net/forum?id=gzqrANCF4g} {Language model beats diffusion - tokenizer is key to visual generation}.
\newblock In \emph{The Twelfth International Conference on Learning Representations, {ICLR} 2024, Vienna, Austria, May 7-11, 2024}. OpenReview.net.

\bibitem[{Yu et~al.(2024{\natexlab{c}})Yu, Yoon, and Bansal}]{CREMA}
Shoubin Yu, Jaehong Yoon, and Mohit Bansal. 2024{\natexlab{c}}.
\newblock \href {https://arxiv.org/abs/2402.05889} {Crema: Generalizable and efficient video-language reasoning via multimodal modular fusion}.
\newblock \emph{Preprint}, arXiv:2402.05889.

\bibitem[{Yu et~al.(2024{\natexlab{d}})Yu, Yang, Li, Wang, Lin, Liu, Wang, and Wang}]{mm-vet}
Weihao Yu, Zhengyuan Yang, Linjie Li, Jianfeng Wang, Kevin Lin, Zicheng Liu, Xinchao Wang, and Lijuan Wang. 2024{\natexlab{d}}.
\newblock \href {https://openreview.net/forum?id=KOTutrSR2y} {Mm-vet: Evaluating large multimodal models for integrated capabilities}.
\newblock In \emph{Forty-first International Conference on Machine Learning, {ICML} 2024, Vienna, Austria, July 21-27, 2024}. OpenReview.net.

\bibitem[{Yu et~al.(2019)Yu, Xu, Yu, Yu, Zhao, Zhuang, and Tao}]{activitynet-qa}
Zhou Yu, Dejing Xu, Jun Yu, Ting Yu, Zhou Zhao, Yueting Zhuang, and Dacheng Tao. 2019.
\newblock \href {https://doi.org/10.1609/AAAI.V33I01.33019127} {Activitynet-qa: {A} dataset for understanding complex web videos via question answering}.
\newblock In \emph{The Thirty-Third {AAAI} Conference on Artificial Intelligence, {AAAI} 2019, The Thirty-First Innovative Applications of Artificial Intelligence Conference, {IAAI} 2019, The Ninth {AAAI} Symposium on Educational Advances in Artificial Intelligence, {EAAI} 2019, Honolulu, Hawaii, USA, January 27 - February 1, 2019}, pages 9127--9134. {AAAI} Press.

\bibitem[{Yue et~al.(2024)Yue, Ni, Zheng, Zhang, Liu, Zhang, Stevens, Jiang, Ren, Sun, Wei, Yu, Yuan, Sun, Yin, Zheng, Yang, Liu, Huang, Sun, Su, and Chen}]{MMMU}
Xiang Yue, Yuansheng Ni, Tianyu Zheng, Kai Zhang, Ruoqi Liu, Ge~Zhang, Samuel Stevens, Dongfu Jiang, Weiming Ren, Yuxuan Sun, Cong Wei, Botao Yu, Ruibin Yuan, Renliang Sun, Ming Yin, Boyuan Zheng, Zhenzhu Yang, Yibo Liu, Wenhao Huang, Huan Sun, Yu~Su, and Wenhu Chen. 2024.
\newblock \href {https://doi.org/10.1109/CVPR52733.2024.00913} {{MMMU:} {A} massive multi-discipline multimodal understanding and reasoning benchmark for expert {AGI}}.
\newblock In \emph{{IEEE/CVF} Conference on Computer Vision and Pattern Recognition, {CVPR} 2024, Seattle, WA, USA, June 16-22, 2024}, pages 9556--9567. {IEEE}.

\bibitem[{Zeghidour et~al.(2022)Zeghidour, Luebs, Omran, Skoglund, and Tagliasacchi}]{soundstream}
Neil Zeghidour, Alejandro Luebs, Ahmed Omran, Jan Skoglund, and Marco Tagliasacchi. 2022.
\newblock \href {https://doi.org/10.1109/TASLP.2021.3129994} {Soundstream: An end-to-end neural audio codec}.
\newblock \emph{{IEEE} {ACM} Trans. Audio Speech Lang. Process.}, 30:495--507.

\bibitem[{Zellers et~al.(2022)Zellers, Lu, Lu, Yu, Zhao, Salehi, Kusupati, Hessel, Farhadi, and Choi}]{yt-temporal}
Rowan Zellers, Jiasen Lu, Ximing Lu, Youngjae Yu, Yanpeng Zhao, Mohammadreza Salehi, Aditya Kusupati, Jack Hessel, Ali Farhadi, and Yejin Choi. 2022.
\newblock \href {https://doi.org/10.1109/CVPR52688.2022.01589} {{MERLOT} {RESERVE:} neural script knowledge through vision and language and sound}.
\newblock In \emph{{IEEE/CVF} Conference on Computer Vision and Pattern Recognition, {CVPR} 2022, New Orleans, LA, USA, June 18-24, 2022}, pages 16354--16366. {IEEE}.

\bibitem[{Zeng et~al.(2024)Zeng, Xu, Wang, Zhang, Yin, Rojas, Feng, Zhao, Lai, Yu, Wang, Sun, Zhang, Cheng, Gui, Tang, Zhang, Li, Zhao, Wu, Zhong, Liu, Huang, Zhang, Zheng, Lu, Duan, Zhang, Cao, Yang, Tam, Zhao, Liu, Xia, Zhang, Gu, Lv, Liu, Liu, Yang, Song, Zhang, An, Xu, Niu, Yang, Li, Bai, Dong, Qi, Wang, Yang, Du, Hou, and Wang}]{chatglm}
Aohan Zeng, Bin Xu, Bowen Wang, Chenhui Zhang, Da~Yin, Diego Rojas, Guanyu Feng, Hanlin Zhao, Hanyu Lai, Hao Yu, Hongning Wang, Jiadai Sun, Jiajie Zhang, Jiale Cheng, Jiayi Gui, Jie Tang, Jing Zhang, Juanzi Li, Lei Zhao, Lindong Wu, Lucen Zhong, Mingdao Liu, Minlie Huang, Peng Zhang, Qinkai Zheng, Rui Lu, Shuaiqi Duan, Shudan Zhang, Shulin Cao, Shuxun Yang, Weng~Lam Tam, Wenyi Zhao, Xiao Liu, Xiao Xia, Xiaohan Zhang, Xiaotao Gu, Xin Lv, Xinghan Liu, Xinyi Liu, Xinyue Yang, Xixuan Song, Xunkai Zhang, Yifan An, Yifan Xu, Yilin Niu, Yuantao Yang, Yueyan Li, Yushi Bai, Yuxiao Dong, Zehan Qi, Zhaoyu Wang, Zhen Yang, Zhengxiao Du, Zhenyu Hou, and Zihan Wang. 2024.
\newblock \href {https://doi.org/10.48550/ARXIV.2406.12793} {Chatglm: {A} family of large language models from {GLM-130B} to {GLM-4} all tools}.
\newblock \emph{CoRR}, abs/2406.12793.

\bibitem[{Zhai et~al.(2023)Zhai, Mustafa, Kolesnikov, and Beyer}]{sigclip}
Xiaohua Zhai, Basil Mustafa, Alexander Kolesnikov, and Lucas Beyer. 2023.
\newblock \href {https://doi.org/10.1109/ICCV51070.2023.01100} {Sigmoid loss for language image pre-training}.
\newblock In \emph{{IEEE/CVF} International Conference on Computer Vision, {ICCV} 2023, Paris, France, October 1-6, 2023}, pages 11941--11952. {IEEE}.

\bibitem[{Zhan et~al.(2024)Zhan, Dai, Ye, Zhou, Zhang, Liu, Zhang, Yuan, Zhang, Li, Yan, Fu, Gui, Sun, Jiang, and Qiu}]{anygpt}
Jun Zhan, Junqi Dai, Jiasheng Ye, Yunhua Zhou, Dong Zhang, Zhigeng Liu, Xin Zhang, Ruibin Yuan, Ge~Zhang, Linyang Li, Hang Yan, Jie Fu, Tao Gui, Tianxiang Sun, Yu{-}Gang Jiang, and Xipeng Qiu. 2024.
\newblock \href {https://doi.org/10.18653/V1/2024.ACL-LONG.521} {Anygpt: Unified multimodal {LLM} with discrete sequence modeling}.
\newblock In \emph{Proceedings of the 62nd Annual Meeting of the Association for Computational Linguistics (Volume 1: Long Papers), {ACL} 2024, Bangkok, Thailand, August 11-16, 2024}, pages 9637--9662. Association for Computational Linguistics.

\bibitem[{Zhang et~al.(2022{\natexlab{a}})Zhang, Lv, Guo, Shao, Yang, Xie, Xu, Bu, Chen, Zeng, Wu, and Peng}]{wenetspeech}
Binbin Zhang, Hang Lv, Pengcheng Guo, Qijie Shao, Chao Yang, Lei Xie, Xin Xu, Hui Bu, Xiaoyu Chen, Chenchen Zeng, Di~Wu, and Zhendong Peng. 2022{\natexlab{a}}.
\newblock \href {https://doi.org/10.1109/ICASSP43922.2022.9746682} {{WENETSPEECH:} {A} 10000+ hours multi-domain mandarin corpus for speech recognition}.
\newblock In \emph{{IEEE} International Conference on Acoustics, Speech and Signal Processing, {ICASSP} 2022, Virtual and Singapore, 23-27 May 2022}, pages 6182--6186. {IEEE}.

\bibitem[{Zhang et~al.(2023{\natexlab{a}})Zhang, Li, Zhang, Zhan, Wang, Zhou, and Qiu}]{speechgpt}
Dong Zhang, Shimin Li, Xin Zhang, Jun Zhan, Pengyu Wang, Yaqian Zhou, and Xipeng Qiu. 2023{\natexlab{a}}.
\newblock \href {https://doi.org/10.18653/V1/2023.FINDINGS-EMNLP.1055} {Speechgpt: Empowering large language models with intrinsic cross-modal conversational abilities}.
\newblock In \emph{Findings of the Association for Computational Linguistics: {EMNLP} 2023, Singapore, December 6-10, 2023}, pages 15757--15773. Association for Computational Linguistics.

\bibitem[{Zhang et~al.(2024{\natexlab{a}})Zhang, Yu, Dong, Li, Su, Chu, and Yu}]{survey_i}
Duzhen Zhang, Yahan Yu, Jiahua Dong, Chenxing Li, Dan Su, Chenhui Chu, and Dong Yu. 2024{\natexlab{a}}.
\newblock \href {https://doi.org/10.18653/V1/2024.FINDINGS-ACL.738} {Mm-llms: Recent advances in multimodal large language models}.
\newblock In \emph{Findings of the Association for Computational Linguistics, {ACL} 2024, Bangkok, Thailand and virtual meeting, August 11-16, 2024}, pages 12401--12430. Association for Computational Linguistics.

\bibitem[{Zhang et~al.(2023{\natexlab{b}})Zhang, Li, and Bing}]{videollama}
Hang Zhang, Xin Li, and Lidong Bing. 2023{\natexlab{b}}.
\newblock \href {https://doi.org/10.18653/V1/2023.EMNLP-DEMO.49} {Video-llama: An instruction-tuned audio-visual language model for video understanding}.
\newblock In \emph{Proceedings of the 2023 Conference on Empirical Methods in Natural Language Processing, {EMNLP} 2023 - System Demonstrations, Singapore, December 6-10, 2023}, pages 543--553. Association for Computational Linguistics.

\bibitem[{Zhang et~al.(2024{\natexlab{b}})Zhang, Fei, Wang, Wu, Cao, Li, and Zhang}]{REAMO}
Meishan Zhang, Hao Fei, Bin Wang, Shengqiong Wu, Yixin Cao, Fei Li, and Min Zhang. 2024{\natexlab{b}}.
\newblock \href {https://doi.org/10.18653/V1/2024.FINDINGS-ACL.863} {Recognizing everything from all modalities at once: Grounded multimodal universal information extraction}.
\newblock In \emph{Findings of the Association for Computational Linguistics, {ACL} 2024, Bangkok, Thailand and virtual meeting, August 11-16, 2024}, pages 14498--14511. Association for Computational Linguistics.

\bibitem[{Zhang et~al.(2022{\natexlab{b}})Zhang, Roller, Goyal, Artetxe, Chen, Chen, Dewan, Diab, Li, Lin, Mihaylov, Ott, Shleifer, Shuster, Simig, Koura, Sridhar, Wang, and Zettlemoyer}]{opt}
Susan Zhang, Stephen Roller, Naman Goyal, Mikel Artetxe, Moya Chen, Shuohui Chen, Christopher Dewan, Mona~T. Diab, Xian Li, Xi~Victoria Lin, Todor Mihaylov, Myle Ott, Sam Shleifer, Kurt Shuster, Daniel Simig, Punit~Singh Koura, Anjali Sridhar, Tianlu Wang, and Luke Zettlemoyer. 2022{\natexlab{b}}.
\newblock \href {https://doi.org/10.48550/ARXIV.2205.01068} {{OPT:} open pre-trained transformer language models}.
\newblock \emph{CoRR}, abs/2205.01068.

\bibitem[{Zhang et~al.(2023{\natexlab{c}})Zhang, Zhang, Li, Zhou, and Qiu}]{SpeechTokenizer}
Xin Zhang, Dong Zhang, Shimin Li, Yaqian Zhou, and Xipeng Qiu. 2023{\natexlab{c}}.
\newblock \href {https://doi.org/10.48550/ARXIV.2308.16692} {Speechtokenizer: Unified speech tokenizer for speech large language models}.
\newblock \emph{CoRR}, abs/2308.16692.

\bibitem[{Zhang et~al.(2023{\natexlab{d}})Zhang, Gong, Zhang, Li, Qiao, Ouyang, and Yue}]{metatransformer}
Yiyuan Zhang, Kaixiong Gong, Kaipeng Zhang, Hongsheng Li, Yu~Qiao, Wanli Ouyang, and Xiangyu Yue. 2023{\natexlab{d}}.
\newblock \href {https://doi.org/10.48550/ARXIV.2307.10802} {Meta-transformer: {A} unified framework for multimodal learning}.
\newblock \emph{CoRR}, abs/2307.10802.

\bibitem[{Zhao et~al.(2023{\natexlab{a}})Zhao, Lin, Zhou, Huang, Feng, and Kang}]{bubogpt}
Yang Zhao, Zhijie Lin, Daquan Zhou, Zilong Huang, Jiashi Feng, and Bingyi Kang. 2023{\natexlab{a}}.
\newblock \href {https://doi.org/10.48550/ARXIV.2307.08581} {Bubogpt: Enabling visual grounding in multi-modal llms}.
\newblock \emph{CoRR}, abs/2307.08581.

\bibitem[{Zhao et~al.(2023{\natexlab{b}})Zhao, Guo, Yue, Chen, Shao, Zhu, Yuan, and Liu}]{chatbridge}
Zijia Zhao, Longteng Guo, Tongtian Yue, Sihan Chen, Shuai Shao, Xinxin Zhu, Zehuan Yuan, and Jing Liu. 2023{\natexlab{b}}.
\newblock \href {https://doi.org/10.48550/ARXIV.2305.16103} {Chatbridge: Bridging modalities with large language model as a language catalyst}.
\newblock \emph{CoRR}, abs/2305.16103.

\bibitem[{Zhong et~al.(2024)Zhong, Wang, Liu, Yang, Tang, Zhang, Li, Qu, Li, Chen, Yu, Wu, Lo, Liu, and Jia}]{Lyra}
Zhisheng Zhong, Chengyao Wang, Yuqi Liu, Senqiao Yang, Longxiang Tang, Yuechen Zhang, Jingyao Li, Tianyuan Qu, Yanwei Li, Yukang Chen, Shaozuo Yu, Sitong Wu, Eric Lo, Shu Liu, and Jiaya Jia. 2024.
\newblock \href {https://arxiv.org/abs/2412.09501} {Lyra: An efficient and speech-centric framework for omni-cognition}.
\newblock \emph{Preprint}, arXiv:2412.09501.

\bibitem[{Zhu et~al.(2024{\natexlab{a}})Zhu, Lin, Ning, Yan, Cui, Wang, Pang, Jiang, Zhang, Li, Zhang, Li, Liu, and Yuan}]{languagebind}
Bin Zhu, Bin Lin, Munan Ning, Yang Yan, Jiaxi Cui, Hongfa Wang, Yatian Pang, Wenhao Jiang, Junwu Zhang, Zongwei Li, Caiwan Zhang, Zhifeng Li, Wei Liu, and Li~Yuan. 2024{\natexlab{a}}.
\newblock \href {https://openreview.net/forum?id=QmZKc7UZCy} {Languagebind: Extending video-language pretraining to n-modality by language-based semantic alignment}.
\newblock In \emph{The Twelfth International Conference on Learning Representations, {ICLR} 2024, Vienna, Austria, May 7-11, 2024}. OpenReview.net.

\bibitem[{Zhu et~al.(2024{\natexlab{b}})Zhu, Qin, Su, Lin, Li, and Gao}]{survey_agriculture_1}
Hongyan Zhu, Shuai Qin, Min Su, Chengzhi Lin, Anjie Li, and Junfeng Gao. 2024{\natexlab{b}}.
\newblock \href {https://doi.org/10.48550/ARXIV.2407.19679} {Harnessing large vision and language models in agriculture: {A} review}.
\newblock \emph{CoRR}, abs/2407.19679.

\bibitem[{Zhu et~al.(2023{\natexlab{a}})Zhu, Hessel, Awadalla, Gadre, Dodge, Fang, Yu, Schmidt, Wang, and Choi}]{Multimodal_c4}
Wanrong Zhu, Jack Hessel, Anas Awadalla, Samir~Yitzhak Gadre, Jesse Dodge, Alex Fang, Youngjae Yu, Ludwig Schmidt, William~Yang Wang, and Yejin Choi. 2023{\natexlab{a}}.
\newblock \href {http://papers.nips.cc/paper\_files/paper/2023/hash/1c6bed78d3813886d3d72595dbecb80b-Abstract-Datasets\_and\_Benchmarks.html} {Multimodal {C4:} an open, billion-scale corpus of images interleaved with text}.
\newblock In \emph{Advances in Neural Information Processing Systems 36: Annual Conference on Neural Information Processing Systems 2023, NeurIPS 2023, New Orleans, LA, USA, December 10 - 16, 2023}.

\bibitem[{Zhu et~al.(2023{\natexlab{b}})Zhu, Hessel, Awadalla, Gadre, Dodge, Fang, Yu, Schmidt, Wang, and Choi}]{multimodalc4}
Wanrong Zhu, Jack Hessel, Anas Awadalla, Samir~Yitzhak Gadre, Jesse Dodge, Alex Fang, Youngjae Yu, Ludwig Schmidt, William~Yang Wang, and Yejin Choi. 2023{\natexlab{b}}.
\newblock \href {http://papers.nips.cc/paper\_files/paper/2023/hash/1c6bed78d3813886d3d72595dbecb80b-Abstract-Datasets\_and\_Benchmarks.html} {Multimodal {C4:} an open, billion-scale corpus of images interleaved with text}.
\newblock In \emph{Advances in Neural Information Processing Systems 36: Annual Conference on Neural Information Processing Systems 2023, NeurIPS 2023, New Orleans, LA, USA, December 10 - 16, 2023}.

\end{thebibliography}
\clearpage
\appendix
\section{Related Survey}\label{section:appendix_related}
With the advent of MLLMs, there are several surveys detailing the current progress of MLLMs. \citet{survey:a,survey:b,survey_m} focus on the early Vision-MLLMs, while \citet{survey_audio} and \citet{3d_survey} respectively summarize the Audio-MLLMs and 3D-MLLMs. \citet{survey_i,survey_j} conduct an investigation into various Specific-MLLMs of different modalities. \citet{survey_g,survey_h} discuss the expansion of MLLM's generative capabilities. Some works discuss MLLMs in specific domains, such as medicine~\citep{survey_medicine_1}, agriculture~\citep{survey_agriculture_1}, and autonomous driving~\citep{survey_driving_1}. Some works highlight some specific tasks such as safety~\citep{survey_safety_1}, hallucination~\citep{survey_hallucination_1}, and acceleration~\citep{survey_agriculture_1}. And \citet{survey_evaluation_1,survey_evaluation_2} focus on the evaluation of MLLM performance.

Distinct from the above-mentioned surveys, this paper focuses on MLLMs that align multiple non-linguistic modalities\footnote{Since MLLMs capable of comprehending both video and imagery generally process video as multiple frames and employ a single vision encoder, we categorize them as Specific-MLLMs, i.e. Vision-MLLMs.} with LLMs (Omni-MLLMs), enabling cross-modal understanding or cross-modal generation. As the first systematic survey on Omni-MLLMs, we hope our work will serve as an overview of this emerging direction, fostering future research in the field.

\section{Details about Omni-MLLMs architures}\label{section:omnimllm_architures_details}
Table~\ref{table:architecture} presents the details of the structure of mainstream Omni-MLLMs. We will list some of the pre-trained models used.
\subsection{Modality Encoder}
\paragraph{Visual Specific-Encoder} Vit~\citep{vit}, SigCLIP Vit~\citep{sigclip}, CLIP Vit~\citep{clip}, EVA CLIP Vit~\citep{evaclip}, InternVit~\citep{internvl}, DINOv2 Vit~\citep{dinov2}, DFNCLIP Vit~\citep{dfnclip}, and OpenCLIP ConvNext~\citep{convnext} encode images to obtain continuous features. TimeSformer~\citep{TimesFormer}, VideoMAE~\citep{videomae}, MAE-DFer~\citep{mae-dfer}, Omni-VL~\citep{omnivl}, Video-Swin~\citep{videoswin}, and Vivit~\citep{vivit} encode videos to obtain continuous features.  
\paragraph{Audio Specific-Encoder} AST~\citep{AST}, Beats~\citep{Beats}, Whisper~\citep{Whisper}, HuBert~\citep{hubert}, CLAP~\citep{CLAP}, Conformer~\citep{Conformer}, MERT~\citep{MERT}, and PANN~\citep{pann} encode the audio modality to obtain continuous features.   
\paragraph{3D Specific-Encoder} ULIP2~\citep{ulip2}, GD-MAE~\citep{GDMAE}, PointEncoder~\citep{pointllm}, FrozenCLIP~\citep{frozenclip}, and M3D-CLIP~\citep{M3DCLIP} encode the 3D modality to obtain continuous features.
\paragraph{Pre-align Uni-Encoder} LanguageBind~\citep{languagebind}, ImageBind~\citep{imagebind}, Meta-Transformers~\citep{metatransformer}, TVL~\citep{TVL}, and SSVTP~\citep{SSVTP} encode multiple non-linguistic modalities into a unified feature space and obtain continuous features. TVL, LanguageBind, ImageBind, and SSVTP construct modality-specific encoders for different modalities and achieve multi-modalities alignment through indirect alignment. Meta-Transformers design distinct modality-specific patch embeddings and use a shared encoder to encode multiple modalities.
\paragraph{Other Specific-Encoder} IMU2CLIP~\citep{IMU2CLIP} encodes the IMU modality to obtain continuous features. Individual modality-specific encoders from LanguageBind or ImageBind are often used independently as specific encoders.
\subsection{Modality Tokenizer}
\paragraph{Visual Tokenizer} 
VQ-GAN~\citep{vqgan}, DALL-E~\citep{Dalle1}, BEiT-V2~\citep{Beitv2}, MAGVIT-v2~\citep{magvit-v2}, and SEED~\citep{SEED} encode the visual modality into discrete visual tokens, which can be decoded back into the original image using the de-tokenizer.
\paragraph{Audio Tokenizer} 
Jukebox~\citep{jukebox}, SoundStream~\citep{soundstream}, SpeechTokenizer~\citep{SpeechTokenizer}, Encodec~\citep{encodec}, and S2U~\citep{emova} encode the audio modality into discrete audio tokens, which can be decoded back into the audio using the corresponding de-tokenizer.
\paragraph{Other Tokenizer}Scene Tokenizer~\citep{occllama} encodes the 3D modality into discrete 3D tokens. LEO~\citep{LEO}, Ground-Action~\citep{groundaction}, OccLLaMA~\citep{occllama}, and GMA~\citep{GMA} perform discrete encoding of the action modality to obtain corresponding action tokens, which can be decoded back into the original action using the corresponding de-tokenizer. M3GPT~\citep{M3GPT}, Gesticulator~\citep{Gesticulator}, and SOLAMI~\citep{solami} perform discrete encoding of the motion modality to obtain corresponding motion tokens, which can be decoded back into the original motion using the corresponding de-tokenizer.

\subsection{Modality Generation Model}
For image generation, Stable Diffusion~\citep{stablediffusion} and Instruct-Pix2Pix~\citep{InstructPix2Pix} are used. Video generation models include Zeroscope~\citep{zeroscope}, VideoFusion~\citep{VideoFusion}, VideoCrafter~\citep{VideoCrafter}, and ModelScope~\citep{ModelScopeT2V}. For audio generation, models such as AudioLDM~\citep{audioldm}, SNAC~\citep{SNAC}, LLaMA-Omni’s audio decoder~\citep{llama-omni}, MusicGen~\citep{musicgen}, and TiCodec~\citep{ticodec} are utilized. Meanwhile, StyleTTS~\citep{styletts} and GPT-SoVITS~\citep{gpt-sovits} are employed for speech generation.

\subsection{LLM Backbone}
Commonly used LLMs include the T5 series~\citep{T5}, LLaMA series~\citep{llama}, Qwen series~\citep{qwen}, Internlm series~\citep{internlm2}, Chatglm series~\citep{chatglm}, OPT series~\citep{opt}, Mixtral series~\citep{mixtral}, Mistral series~\citep{Mistral}, Phi series~\citep{phi}, and Yi series~\citep{Yi}.
\section{Details of Training and evaluation}\label{section:omnimllm_traing_training_evaluation_details}
\subsection{Details of Training Data}\label{section:omnimllm_traing_data_details}
The statistical results of some commonly used alignment datasets and the instruction data of mainstream Omni-MLLMs are shown in Table~\ref{table:alignment_dataset} and Table~\ref{table:sft_dataset}. There is still a lack of alignment data for data-scarcity modalities and cross-modal instruction data.  

\begin{table*}[t]
        \centering
    \resizebox{1.0\linewidth}{!}{\begin{tabular}{l|l|ccc| ccc|cc|ccc}
    \toprule
       \multirow{2}{*}{Model} & \multirow{2}{*}{Capabilities}   & \multicolumn{3}{c|}{Multi-Modalities Encoding}  & \multicolumn{3}{c|}{Multi-Modalities Alignment} & \multicolumn{2}{c|}{Multi-Modalities Interaction} & \multicolumn{3}{c}{Multi-Modalities Generation} \\
 &  &Modalities &   Method 
 & Encoding Model
 & Method & Projector & Vocabulary & Method & LLM & Modalities & Method  & Generation model \\ 
 \midrule
eP-ALM 
& \makecell{Corss-modal Understanding} 
& Visual/Audio & Continuous Encoding & Vit/TimeSformer/AST
& multi-branch & linear & -- 
& injection & OPT 
& -- & --  & -- \\
VALOR 
& \makecell{Corss-modal Understanding} 
& Visual/Audio & Continuous Encoding & Vit/VideoSwin/AST
& multi-branch & MLP & -- 
& injection & Bert 
& -- & --  & -- \\
X-LLM 
& \makecell{Corss-modal Understanding} 
& Visual/Audio & Continuous Encoding & Vit/Conformer
& multi-branch & Q-former+Linear & -- 
& concatenate & ChatGLM 
& -- & --  & -- \\
ChatBridge
& \makecell{Corss-modal Understanding} 
& Visual/Audio & Continuous Encoding & EVAL CLIP Vit/Beats
& multi-branch & Preceiver & -- 
& concatenate & Vicuna 
& -- & --  & -- \\
PandaGPT
& \makecell{Corss-modal Understanding} 
& \makecell{Visual/Audio/3D/ \\IMU/thermal} & Continuous Encoding & ImageBind
& uni-branch & linear & -- 
& concatenate & Vicuna 
& -- & --  & -- \\
VideoLLama
& \makecell{Corss-modal Understanding} 
& \makecell{Visual/Audio} & Continuous Encoding & \makecell{EVA CLIP Vit/ \\ImageBind-Audio}
& multi-branch & Q-former+Linear & -- 
& concatenate & Vicuna 
& -- & --  & -- \\
LAMM
& \makecell{Corss-modal Understanding} 
& \makecell{Visual/Audio} & Continuous Encoding & \makecell{CLIP Vit/ \\FrozenCLIP}
& multi-branch & MLP & -- 
& concatenate & Vicuna 
& -- & --  & -- \\
Macaw-LLM
& \makecell{Corss-modal Understanding} 
& \makecell{Visual/Audio} & Continuous Encoding & Vit/Whisper
& multi-branch & Cross-Attention & -- 
& concatenate & LLaMA 
& -- & --  & -- \\
BuboGPT
& \makecell{Corss-modal Understanding} 
& \makecell{Visual/Audio} & Continuous Encoding & CLIP Vit/ImageBind-Audio
& multi-branch & Q-former+Linear & -- 
& concatenate & LLaMA 
& -- & --  & -- \\
Teal
& \makecell{Corss-modal Understanding\\ \&\& Generation} 
& \makecell{Visual/Audio} & Discrete Encoding & \makecell{VQ-GAN/\\Whisper+K-means}
& embedding & -- & Added Vocabulary
& concatenate & LLaMA 
& Image & Modality-Token-based  & VQGAN detokenizer \\
ImageBind-LLM
& \makecell{Corss-modal Understanding} 
& \makecell{Visual/Audio/3D} & Continuous Encoding & \makecell{ImageBind}
& uni-branch & MLP & --
& injection & LLaMA 
& -- & --  & -- \\
Next-GPT
& \makecell{Corss-modal Understanding\\ \&\& Generation}
& \makecell{Visual/Audio} & Continuous Encoding & \makecell{ImageBind}
& multi-branch & Linear & --
& concatenate & Vicuna 
& Image/Video/Audio & Representation-based & \makecell{StableDiffusion1.5/ \\Zeroscope/AudioLDM} \\
Any-MAL
& \makecell{Corss-modal Understanding}
& \makecell{Visual/Audio/IMU} & Continuous Encoding & \makecell{CLIP Vit/CLAP/\\IMU2CLIP}
& multi-branch & Preceiver & --
& injection & LLaMA-2 
& -- & -- & -- \\
FAVOR
& \makecell{Corss-modal Understanding}
& \makecell{Visual/Audio} & Continuous Encoding & \makecell{EVA CLIP Vit/Whisper}
& multi-branch & Q-former+Linear & --
& concatenate & Vicuna
& -- & -- & -- \\
Octavius
& \makecell{Corss-modal Understanding}
& \makecell{Visual/3D} & Continuous Encoding & \makecell{CLIP Vit/Object-As-Scene}
& multi-branch & MLP & --
& concatenate & Vicuna
& -- & -- & -- \\
LEO
& \makecell{Corss-modal Understanding\\ \&\& Generation}
& \makecell{Visual/3D/Action} & Hybrid Encoding & \makecell{OpenCLIP Convnext/PointNet++/\\LEO's Action tokenizer}
& multi-branch & \makecell{ MLP/\\Spatial Transformer} & Overwrite Vocabulary
& concatenate & Vicuna
& Action & Modality-Token-based & LEO's Action detokenizer \\

CoDi-2
& \makecell{Corss-modal Understanding\\ \&\& Generation}
& \makecell{Visual/Audio} & Continuous Encoding & \makecell{ImageBind}
& uni-branch & \makecell{ MLP} & --
& concatenate & LLaMA-2
& Image/Video/Audio & Representation-based & \makecell{StableDiffusion2.1/\\Zeroscope/AudioLDM2} \\

X-InstructBLIP
& \makecell{Corss-modal Understanding}
& \makecell{Visual/Audio/3D} & Continuous Encoding & \makecell{EVA CLIP Vit/Beats/\\ULIP2}
& multi-branch & \makecell{ Q-former+Linear} & --
& concatenate & Vicuna
& -- & -- & \makecell{--} \\

One-LLM
& \makecell{Corss-modal Understanding}
& \makecell{Visual/Audio/3D/IMU/fMRI/\\Normal Map} & Continuous Encoding & \makecell{Meta-transformer}
& uni-branch & \makecell{UPM(self-attention)} & --
& concatenate & LLaMA-2
& -- & -- & \makecell{--} \\

AV-LLM
& \makecell{Corss-modal Understanding}
& \makecell{Visual/Audio} & Continuous Encoding & \makecell{CLIP Vit/CLAP}
& multi-branch & \makecell{Linear} & --
& concatenate & Vicuna
& -- & -- & \makecell{--} \\

DriveMLM
& \makecell{Corss-modal Understanding}
& \makecell{Visual/3D} & Continuous Encoding & \makecell{EVA CLIP Vit/GD-MAE}
& multi-branch & \makecell{Q-former+Linear} & --
& concatenate & LLaMA
& -- & -- & \makecell{--} \\

Omni-3D 
& \makecell{Corss-modal Understanding}
& \makecell{Visual/3D} & Continuous Encoding & \makecell{CLIP Vit/PointNet++}
& multi-branch & \makecell{MLP} & --
& concatenate & LLaMA-2
& -- & -- & \makecell{--} \\

ModaVerse
& \makecell{Corss-modal Understanding\\ \&\& Generation}
& \makecell{Visual/Audio} & Continuous Encoding & \makecell{ImageBind}
& multi-branch & \makecell{Linear} & --
& concatenate & Vicuna
 & Image/Video/Audio & Text-based & \makecell{StabelDiffusion\\/AudioLDM/VideoFusion} \\

MultiPLY
& \makecell{Corss-modal Understanding}
& \makecell{Visual/Audio/3D\\/thermal/touch} & Continuous Encoding & \makecell{CLIP Vit/CLAP\\/ConceptGraph}
& multi-branch & \makecell{MLP/Linear} & --
& concatenate & Vicuna
& -- & -- & \makecell{--} \\

CREMA
& \makecell{Corss-modal Understanding}
& \makecell{Visual/Audio/3D\\/thermal/touch/optical} & Continuous Encoding & \makecell{EVA CLIP Vit+Linear/\\Beats+Linear/ConceptFusion+Linear}
& uni-branch & \makecell{Q-former+Linear} & --
& concatenate & Flan-T5
& -- & -- & \makecell{--} \\

GroundingGPT
& \makecell{Corss-modal Understanding}
& \makecell{Visual/Audio} & Continuous Encoding & \makecell{CLIP Vit/ImageBind-Audio}
& multi-branch & \makecell{Q-former+Linear/MLP} & --
& concatenate & Vicuna
& -- & -- & \makecell{--} \\

DAMC
& \makecell{Corss-modal Understanding}
& \makecell{Visual/Audio} & Continuous Encoding & \makecell{CLIP Vit/Beats/PointBert(PointLLM)}
& multi-branch & \makecell{Q-former+Linear/MLP} & --
& concatenate & Vicuna
& -- & -- & \makecell{--} \\

AnyGPT
& \makecell{Corss-modal Understanding\\ \&\& Generation}
& \makecell{Visual/Audio} & Diserect Encoding & \makecell{SEED tokenizer/\\Speech tokenizer/Encodec}
& embedding & \makecell{--} & \makecell{Extend Vocabulary}
& concatenate & LLaMA-2
& Image/Speech/Music & -- & \makecell{SEED de-tokenizer/Speech \\de-tokenizer/Encodec de-tokenizer} \\

TVL-LLaMA
& \makecell{Corss-modal Understanding}
& \makecell{Visual/Touch} & Continuous Encoding & \makecell{TVL Encoders}
& uni-branch & \makecell{MLP} & \makecell{--}
& injection & LLaMA
& -- & -- & -- \\

SSVTP-LLaMA
& \makecell{Corss-modal Understanding}
& \makecell{Visual/Touch} & Continuous Encoding & \makecell{SSVTP Encoders}
& uni-branch & \makecell{MLP} & \makecell{--}
& injection & LLaMA
& -- & -- & -- \\

CAT
& \makecell{Corss-modal Understanding}
& \makecell{Visual/Audio} & Continuous Encoding & \makecell{ImageBind}
& multi-branch & \makecell{Linear} & \makecell{--}
& concatenate & LLaMA-2
& -- & -- & -- \\

AVicuna
& \makecell{Corss-modal Understanding}
& \makecell{Visual/Audio} & Continuous Encoding & \makecell{CLIP Vit/CLAP}
& multi-branch & \makecell{MLP} & \makecell{--}
& concatenate & Vicuna
& -- & -- & -- \\

WorldGPT
& \makecell{Corss-modal Understanding\\ \&\& Generation}
& \makecell{Visual/Audio} & Continuous Encoding & \makecell{LanguageBind}
& uni-branch & \makecell{Linear} & \makecell{--}
& concatenate & Vicuna
& Image/Video/Audio & Representation-based  & \makecell{Stable Diffusion\\/AudioLDM/Zeroscope} \\

QaP
& \makecell{Corss-modal Understanding}
& \makecell{Visual/Audio} & Continuous Encoding & \makecell{CLIP Vit/CLAP}
& multi-branch & \makecell{dot attention+Linear} & \makecell{--}
& injection & DeBERTa-V2-XLarge
& -- & --  & \makecell{--} \\

Uni-Moe
& \makecell{Corss-modal Understanding}
& \makecell{Visual/Audio} & Continuous Encoding & \makecell{CLIP Vit/Beats}
& multi-branch & \makecell{Q-former+Linear/MLP} & \makecell{--}
& concatenate & LLaMA
& -- & --  & \makecell{--} \\

M3GPT
& \makecell{Corss-modal Understanding}
& \makecell{Audio/Motion} & Diserect Encoding & \makecell{Jukebox tokenizer\\/M3GPT's Motion tokenier}
& embedding & \makecell{Extend Vocabulary} & \makecell{--}
& concatenate & T5
& Music/Motion & Modality-Token-based   & \makecell{Jukebox de-tokenizer\\/M3GPT's Motion de-tokenizer} \\

X-VILA
& \makecell{Corss-modal Understanding}
& \makecell{Visual/Audio} & Continuous Encoding & \makecell{ImageBind}
& multi-branch & \makecell{MLP} & \makecell{--}
& concatenate & Vicuna
& Image/Video/Audio & Representation-based    & \makecell{Stable Diffusion\\/AudioLDM/VideoCrafter} \\

REAMO
& \makecell{Corss-modal Understanding}
& \makecell{Visual/Audio} & Continuous Encoding & \makecell{ImageBind}
& multi-branch & \makecell{Linear} & \makecell{--}
& concatenate & Vicuna
& -- & --    & \makecell{--} \\

VideoLLaMA2
& \makecell{Corss-modal Understanding}
& \makecell{Visual/Audio} & Continuous Encoding & \makecell{CLIP Vit/Beats}
& multi-branch & \makecell{MLP} & \makecell{--}
& concatenate & Mistral
& -- & --    & \makecell{--} \\

Ground-Action
& \makecell{Corss-modal Understanding}
& \makecell{Visual/Action} & Hybrid Encoding & \makecell{CLIP Vit\\/Ground-Action action tokenizer}
& multi-branch & \makecell{Preceiver} & \makecell{Overwrite Vocabulary}
& concatenate & Vicuna
& Action & Modality-Token-based    & \makecell{Ground-Action action de-tokenizer} \\

Emotion-LLaMA
& \makecell{Corss-modal Understanding\\ \&\& Generation}
& \makecell{Visual/Audio} & Continuous Encoding & \makecell{CLIP Vit\\/HuBert-Chinese}
& multi-branch & \makecell{MLP} & \makecell{--}
& concatenate & LLaMA-2
& -- & --    & -- \\

EmpathyEar
& \makecell{Corss-modal Understanding\\ \&\& Generation}
& \makecell{Visual/Audio} & Continuous Encoding & \makecell{ImageBind}
& uni-branch & \makecell{Linear} & \makecell{--}
& concatenate & ChatGLM
& Video/Audio  & Text-based    & StyleTTS2/EAT \\

video-SALMONN
& \makecell{Corss-modal Understanding}
& \makecell{Visual/Audio} & Continuous Encoding & \makecell{Vit/Beats}
& multi-branch & \makecell{Q-former+Linear} & \makecell{--}
& concatenate & Vicuna
& --  & --   & -- \\

Meerkat
& \makecell{Corss-modal Understanding}
& \makecell{Visual/Audio} & Continuous Encoding & \makecell{CLIP Vit/CLAP}
& multi-branch & \makecell{MLP} & \makecell{--}
& concatenate & LLaMA-2
& --  & --   & -- \\

InternOmni
& \makecell{Corss-modal Understanding}
& \makecell{Visual/Audio} & Continuous Encoding & \makecell{Intern Vit/Whisper}
& multi-branch & \makecell{MLP} & \makecell{--}
& concatenate & InternLM-2.5
& --  & --   & -- \\

SynesLM
& \makecell{Corss-modal Understanding}
& \makecell{Visual/Audio} & Hybird Encoding & \makecell{SigCLIP Vit\\/XLSR+K-means}
& multi-branch & \makecell{MLP} & \makecell{Extend Vocabulary}
& concatenate & OPT
& --  & --   & -- \\

UnifiedMLLM
& \makecell{Corss-modal Understanding\\ \&\& Generation}
& \makecell{Visual/Audio} & Continuous Encoding & \makecell{CLIP Vit\\/ImageBind-Audio}
& multi-branch & \makecell{Q-former+Linear} & \makecell{--}
& concatenate & OPT
& Image/Video/Audio  & Text-based   & \makecell{Instruct-pix2pix/\\Auffusion/ModelScope} \\

VITA
& \makecell{Corss-modal Understanding\\ \&\& Generation}
& \makecell{Visual/Audio} & Continuous Encoding & \makecell{CLIP Vit\\/VITA's Audio Encoder}
& multi-branch & \makecell{MLP} & \makecell{--}
& concatenate & Mixtral
& Speech  & Text-based   & \makecell{GPT-SoVITS} \\

OccLLaMA
& \makecell{Corss-modal Understanding\\ \&\& Generation}
& \makecell{3D/Action} & Discrete Encoding & \makecell{OccLLaMA's 3D tokenizer\\/OccLLaMA's Action tokenizer}
& embedding & \makecell{--} & \makecell{Extend Vocabulary}
& concatenate & LLaMA-3.1
& Action/3D  & Modality-Token-based    & \makecell{OccLLaMA's 3D de-tokenizer\\/OccLLaMA's Action de-tokenizer} \\

Llama-AVSR
& \makecell{Corss-modal Understanding\\ \&\& Generation}
& \makecell{Visual/Audio} & Continuous Encoding & \makecell{AV-HuBert\\/Whisper}
& multi-branch & \makecell{MLP} & \makecell{--}
& concatenate & LLaMA-3.1
& --  & --   & -- \\

MIO
& \makecell{Corss-modal Understanding\\ \&\& Generation}
& \makecell{Visual/Audio} & Discrete Encoding & \makecell{SEED tokenizer\\/Speech tokenizer}
& embedding & \makecell{--} & \makecell{Extend Vocabulary}
& concatenate & Yi
& Image/Speech  & Modality-Token-based   & \makecell{SEED de-tokenizer\\/Speech
de-tokenizer} \\

EMOVA
& \makecell{Corss-modal Understanding\\ \&\& Generation}
& \makecell{Visual/Audio} & Hybird Encoding & \makecell{Intern Vit\\/EMOVA's S2U tokenizer}
& multi-branch & \makecell{C-Abstractor} & \makecell{Extend Vocabulary}
& concatenate & LLaMA-3.1
& Speech  & Modality-Token-based   & \makecell{EMOVA's S2U de-tokenizer} \\

LLM Gesticulator
& \makecell{Corss-modal Understanding\\ \&\& Generation}
& \makecell{Audio/Motion} & Diserect Encoding & \makecell{MotionRVQ tokenizer\\/Encodec tokenizer}
& embedding & \makecell{--} & \makecell{Extend Vocabulary}
& concatenate & Qwen-1.5
& Audio/Motion  & Modality-Token-based   & \makecell{MotionRVQ de-tokenizer\\/Encodec de-tokenizer} \\

Baichuan-Omni
& \makecell{Corss-modal Understanding}
& \makecell{Audio/Motion} & Continuous Encoding & \makecell{SigCLIP \\/Whisper}
& multi-branch & \makecell{CNN+MLP/Conv-GMLP} & \makecell{--}
& concatenate & --
& --  & --   & --  \\

EGMI
& \makecell{Corss-modal Understanding\\ \&\& Generation}
& \makecell{Audio/Motion} & Continuous Encoding & \makecell{ImageBind}
& uni-branch & \makecell{Linear} & \makecell{--}
& concatenate & Vicuna
& Image/Audio  & Representation-based   & \makecell{StableDiffusion\\/AudioLDM}  \\

Dolphin
& \makecell{Corss-modal Understanding\\ \&\& Generation}
& \makecell{Visual/Audio} & Continuous Encoding & \makecell{CLIP Vit/ImageBind-Audio}
& multi-branch & \makecell{MLP} & \makecell{--}
& concatenate & Vicuna
& --  & --   & --  \\

Mini-Omni2
& \makecell{Corss-modal Understanding\\ \&\& Generation}
& \makecell{Visual/Audio} & Continuous Encoding & \makecell{CLIP Vit/Whisper}
& multi-branch & \makecell{MLP} & \makecell{--}
& concatenate & Qwen2
& Audio  & Modality-Token-based   & SNAC de-tokenizer  \\

OMCAT
& \makecell{Corss-modal Understanding}
& \makecell{Visual/Audio} & Continuous Encoding & \makecell{CLIP Vit/ImageBind-Audio}
& multi-branch & \makecell{Q-former+transformer} & \makecell{--}
& concatenate & Vicuna
& --  & --   & --  \\

PathWeave
& \makecell{Corss-modal Understanding}
& \makecell{Visual/Whisper} & Continuous Encoding & \makecell{EVA CLIP Vit/Beats\\/ULIP2}
& uni-branch & \makecell{Q-former+Linear} & \makecell{--}
& concatenate & Vicuna
& --  & --   & --  \\

CAD-MLLM
& \makecell{Corss-modal Understanding}
& \makecell{Visual/3D} & Continuous Encoding & \makecell{DINO v2\\/Michelangelo}
& multi-branch & \makecell{Preceiver+Linear/Linear} & \makecell{--}
& concatenate & Vicuna
& --  & --   & --  \\

EAGLE
& \makecell{Corss-modal Understanding}
& \makecell{Visual/Audio} & Continuous Encoding & \makecell{ImageBind}
& multi-branch & \makecell{MLP} & \makecell{--}
& concatenate & LLaMA-2
& Image/Video/Audio  & Text-based   & \makecell{StableDiffusion\\/AudioLDM/Zeroscope}   \\

Spider
& \makecell{Corss-modal Understanding\\ \&\& Generation}
& \makecell{Visual/Audio} & Continuous Encoding & \makecell{ImageBind}
& multi-branch & \makecell{MLP} & \makecell{--}
& concatenate & LLaMA-2
& Image/Video/Audio  & Text-based   & \makecell{StableDiffusion\\/AudioLDM/Zeroscope}   \\

Med-2E3
& \makecell{Corss-modal Understanding}& \makecell{Visual/3D}
 & Continuous Encoding & \makecell{SigCLIP Vit/M3D-CLIP}
& multi-branch & \makecell{Q-former+Linear/MLP} & \makecell{--}
& concatenate & Phi
& --  & --   & --  \\

LongVALE-LLM
& \makecell{Corss-modal Understanding}& \makecell{Visual/Audio}
 & Continuous Encoding & \makecell{CLIP Vit/Beats/Whisper}
& multi-branch & \makecell{MLP} & \makecell{--}
& concatenate & Vicuna
& --  & --   & --  \\

SOLAMI
& \makecell{Corss-modal Understanding}& \makecell{Audio/Motion}
 & Diserect Encoding & \makecell{SpeechTokenizer\\/SOLAMI's MotionTokenizer}
& embedding & \makecell{--} & \makecell{Extend Vocabulary}
& concatenate & Vicuna
& Speech/Motion  & Modality-Token-based    & \makecell{Speech de-tokenizer\\/SOLAMI's Motion de-tokenizer}  \\

MuMu-LLaMA
& \makecell{Corss-modal Understanding\\ \&\& Generation}
& \makecell{Visual/Audio}
 & Continuous Encoding & \makecell{Vit/ViViT/MERT}
& multi-branch & \makecell{Conv+MLP/Conv+Rnn+MLP} & \makecell{--}
& injection & LLaMA-2
& Music  & Representation-based    & \makecell{MusicGen}  \\

GMA
& \makecell{Corss-modal Understanding\\ \&\& Generation}
& \makecell{Visual/Action}
 & Hybird Encoding & \makecell{SigCLIP Vit\\/GMA's Action tokenizer}
& multi-branch & \makecell{MLP} & \makecell{Overwrite Vocabulary}
& concatenate & Qwen-2
& Action  & Modality-Token-based   & \makecell{GMA's Action de-tokenizer}  \\

Lyra
& \makecell{Corss-modal Understanding\\ \&\& Generation}
& \makecell{Visual/Audio}
 & Continuous Encoding & \makecell{DFNCLIP Vit/Whipser}
& multi-branch & \makecell{MLP} & \makecell{--}
& concatenate & Qwen-2
& Audio  & Representation-based    & \makecell{LLaMA-Omni's audio decoder}  \\

VITA-1.5
& \makecell{Corss-modal Understanding\\ \&\& Generation}
& \makecell{Visual/Audio}
 & Continuous Encoding & \makecell{InternVit/VITA's Audio Encoder}
& multi-branch & \makecell{MLP/CNN+MLP} & \makecell{--}
& concatenate & Mixtral
& Speech  & Representation-based    & \makecell{TiCodec decoder}  \\

 EmpatheticLLM
& \makecell{Corss-modal Understanding}
& \makecell{Visual/Audio}
 & Continuous Encoding & \makecell{CLIP Vit/Whisper}
& multi-branch & \makecell{Q-former+Linear} & \makecell{--}
& concatenate & Qwen2.5
& --  & --   & --  \\
        \bottomrule
    \end{tabular}}\caption{ \textbf{The architectures of mainstream OmniMLLMs.} The architectures of 70 Omni-MLLMs are displayed by encoding, alignment, interaction, and generation.}\label{table:architecture}   
\end{table*}
\begin{table*}[t]
    \centering
     \resizebox{0.8\linewidth}{!}{
    \begin{tabular}{lllc}
        
    \toprule
        Name & Type & Modality & \#Sample \\ 
 \midrule
        MSCOCO~\citep{MSCOCO} & X-Text & Image,Text & 620K \\ 
        Visual Genome~\citep{visualgenome} & X-Text & Image,Text & 4.5M\\ 
        Flickr30k~\citep{flicker30k} & X-Text & Image,Text & 158K \\ 
        SBU~\citep{SBU} & X-Text & Image,Text & 1M \\ 
        DCI~\citep{DCI} & X-Text & Image,Text & 7.8K \\ 
        BLIP-Capfilt~\citep{BLIP}  & X-Text & Image,Text & 129M \\ 
        AI Challenger captions~\citep{ai-Challenger} & X-Text & Image,Text & 1.5M \\ 
        Wukong Captions~\citep{wukong} & X-Text & Image,Text & 101M \\ 
        CC12M~\citep{cc12m} & X-Text & Image,Text & 12.4M \\ 
        CC3M~\citep{cc3m} & X-Text & Image,Text & 3.3M \\ 
        LAION-5B~\citep{LAION-5B} & X-Text & Image,Text & 5.9B \\ 
        Redcaps~\citep{RedCaps} & X-Text & Image,Text & 12M \\ 
        LAION-COCO~\citep{laioncoco} & X-Text & Image,Text & 600M \\ 
        LAION-CAT~\citep{LAION-Cat} & X-Text & Image,Text & 440M \\ 
        LAION-AESTHETICS~\citep{laion-aesthetics} & X-Text & Image,Text & 120M \\ 
        ShareGPT4V~\citep{sharegpt4v} & X-Text & Image,Text & 1.2M \\ 
        LAION-115M~\citep{laion-400m} & X-Text & Image,Text & 115M \\ 
        Journeydb~\citep{Journeydb} & X-Text & Image,Text & 4.4M \\ 
        Multimodal c4~\citep{multimodalc4} & X-Text-X & Image,Text & 43.3M \\ 
        OBELICS~\citep{obelics}  & X-Text-X & Image,Text & 141M \\ 
         Panda-70M~\citep{panda-70m} & X-Text & Video,Text & 70M \\ 
        Webvid2M~\citep{webvid} & X-Text & Video,Text & 2M \\ 
        Valley-Pretrain-703k~\citep{valley} & X-Text & Video,Text & 703K \\ 
        Webvid10M~\citep{webvid} & X-Text & Video,Text & 10M \\ 
        YT-Temporal~\citep{yt-temporal} & X-Text & Video,Text & 180M \\ 
         ActivityNet Captions~\citep{activitynet-caption} & X-Text & Video,Text & 100K \\ 
        InterVid~\citep{internvid}  & X-Text & Video,Text & 10M \\ 
        MSRVTT~\citep{msrvtts}  & X-Text & Video,Text & 200K \\ 
        ShareGemini~\citep{sharegemini} & X-Text & Video,Text & 530K \\ 
        AudioSet~\cite{audioset} & X-Text & Audio,Text & 2.1M \\ 
        Clotho~\citep{clotho} & X-Text & Audio,Text & 5k \\ 
        Auto-ACD~\citep{autocad} & X-Text & Audio,Text & 1.5M \\ 
        AudioCap~\citep{audiocap} & X-Text & Audio,Text & 46k \\ 
        WavCaps~\citep{wavcaps} & X-Text & Audio,Text & 403K \\ 
        AISHELL-1~\citep{AISHELL-1} & X-Text & Audio,Text & 128K \\ 
        AISHELL-2~\citep{aishell-2} & X-Text & Audio,Text & 1M \\ 
         Gigaspeech~\citep{gigaspeech} & X-Text & Speech,Text & -- \\ 
        Common Voice~\citep{commonvoice} & X-Text & Speech,Text & -- \\ 
        MLS~\citep{MLS} & X-Text & Speech,Text & -- \\ 
        Music caption~\citep{anygpt} & X-Text & Music,Text & 100M \\ 
        Cap3D~\citep{cap3d}  & X-Text & 3D,Text & 1M \\ 
        Objaverse~\citep{objaverse} & X-Text & 3D,Text & 800K \\  
        ScanRefer~\citep{ScanRefer} & X-Text & 3D,Text & 51.5K \\ 
         Normal Caption~\citep{onellm} & X-Text & Normal,Text & 0.5M \\ 
         Depth Caption~\citep{onellm}& X-Text & Depth,Text & 0.5M \\ 
           NSD~\citep{nsd}  & X-Text & fMRI,Text & 9K \\ 
        Ego4d~\citep{ego4d} & X-Text & Video,IMU,Text & 528k \\ 
        PU-VALOR~\citep{avicuna} & X-Y-Text & Video,Audio,Text & 114K \\ 
        VALOR~\citep{VALOR}  & X-Y-Text & Video,Audio,Text & 16k \\ 
        VAST~\citep{vast}  & X-Y-Text & Video,Audio,Text & 414k \\

        VIDAL~\citep{languagebind} & X-Y-Text &  Video, Thermal, Depth, Audio & 10M \\ 
        
        TVL~\citep{TVL} & X-Y-Text & Image,Touch,Text & 44K \\ 
        M3D-Cap~\citep{M3DCLIP} & X-Y-Text & Image,3D,Text & 115K \\

        \bottomrule
    \end{tabular}
     
     }\caption{ \textbf{The statistics for alignment datasets in Omni-MLLMs}, including single non-linguistic modality text pairing data (X-Text), multiple non-linguistic modalities text pairing data (X-Text-Y), and single non-linguistic modality text interleaved data (X-Text-X).}\label{table:alignment_dataset}
\end{table*}
    \begin{table*}[t]
    \centering
    \resizebox{1.0\linewidth}{!}{
     \begin{tabular}{lcllll}
        \toprule
        Name & Source & Task & Modality & Construction Method & \#Sample \\ 
        \midrule
                XLLM's SFT~\cite{xllm2023chen} & \makecell{MiniGPT-4, AISHELL-2,\\VSDial-CN, ActivityNet Caps} & Uni-Modal Understanding,Cross-Modal Understanding & Image,Video,Audio,Text & \makecell{Template Instructionalization,\\T2X generation} & 10k \\ 
        ChatBridge's SFT~\citep{chatbridge} & \makecell{MSRVTT, AudioCaps,\\VQAv2, VG-QA...} & Uni-Modal Understanding,Cross-Modal Understanding & Image,Video,Audio,Text & \makecell{Template Instructionalization,\\GPT generation} & 4.4M+209k \\ 
        
        Macaw-LLM~\citep{macawllm} & \makecell{MSCOCO, Charades,\\AVSD, VG-QA...} & Uni-Modal Understanding,Cross-Modal Understanding & Image,Video,Audio,Text & \makecell{GPT generation} & 69K+50K \\ 
        BuboGPT's SFT & \makecell{LLaVA, Clotho,\\VGGSS} & Uni-Modal Understanding,Cross-Modal Understanding & Image,Audio,Text & \makecell{Template Instructionalization,\\GPT generation} & 196K \\ 
        
        NextGPT's SFT~\citep{nextgpt} & \makecell{WebVid, CC3M,\\AudioCap, Youtube...} & \makecell{Uni-Modal Understanding,Cross-Modal Understanding,\\Uni-Modal Generation,Cross-Modal Generation} & Image,Video,Audio,Text &  \makecell{Template Instructionalization,\\GPT generation+retrieval,\\T2X generation} & 20K \\ 
        
        AnyMal's SFT~\citep{anymal} & -- & Uni-Modal Understanding & Image,Video,Audio,Text & \makecell{Manual Annotation,\\GPT generation} & 210K \\ 
        
        FAVOR's SFT~\citep{favor} & \makecell{LLaVA, MSCOCO,\\Ego4D, LibriSpeech...}  & Uni-Modal Understanding,Cross-Modal Understanding & Image,Video,Audio,Text & \makecell{Template Instructionalization,\\GPT generation} &  -- \\ 
        
        LEO's SFT~\citep{LEO} & \makecell{ScanQA, SQA3D,\\3RScan, CLIPort...} & Uni-Modal Understanding,Cross-Modal Understanding & Image,3D,Text,Action & \makecell{Template Instructionalization,\\GPT generation} & 220k \\ 
        
        CoDi-2's SFT~\citep{codi2} & \makecell{MIMIC-IT, LAION-400M,\\AudioSet, Webvid...} & \makecell{Uni-Modal Understanding,Uni-Modal Generation} & Image,Audio,Text & \makecell{Template Instructionalization} & -- \\ 
        
        X-InstructBLIP's SFT~\citep{xinstruclblip} & \makecell{MSCOCO, Clotho,\\MSVD, Cap3D...} & Uni-Modal Understanding & Image,Video,Audio,3D,Text & \makecell{Template Instructionalization,\\GPT generation} & 1.6M \\ 
        
        OneLLM's SFT~\citep{onellm} & \makecell{LLaVA-150K, Clotho,\\Ego4D, NSD...} & Uni-Modal Understanding & \makecell{Image,Video,Audio,3D,ImU,\\Depth,fMRI,Normal,Text} & \makecell{Template Instructionalization,\\T2X generation} & 2M \\

        AVLLM's SFT~\citep{avllm} & \makecell{ACAV100M, VGGSound,\\WebVid2M, WavCaps...} & Uni-Modal Understanding,Cross-Modal Understanding & Video,Audio,Text & \makecell{GPT generation} & 1.4M \\

        Uni-IO2's SFT~\citep{unifiedio2} & \makecell{CC3M, AudioSet,\\Webvid3m, Omni3D...} & \makecell{Uni-Modal Understanding,Cross-Modal Understanding,\\Uni-Modal Generation,Cross-Modal Generation} & Image,Video,Audio,Text & \makecell{Template Instructionalization,\\GPT generation} & 775m \\

        ModaVerse's SFT~\citep{modaverse} & -- & \makecell{Uni-Modal Understanding,Cross-Modal Understanding,\\Uni-Modal Generation} & Image,Video,Audio,Text & \makecell{GPT generation} & 2M \\ 
        
        REAMO's SFT~\citep{REAMO} & -- & Uni-Modal Understanding,Cross-Modal Understanding & Image,Video,Audio,Text & \makecell{Template Instructionalization} & 10K \\ 

        GroundingGPT's SFT~\citep{groundinggpt} & \makecell{Flickr30K, VCR,\\Activitynet Captions, Clotho...} & Uni-Modal Understanding & Image,Video,Audio,Text & \makecell{GPT Instructionalization} & 1M \\ 
        
        AnyGPT's SFT~\citep{aishell-2} & -- & \makecell{Uni-Modal Understanding,Cross-Modal Understanding,\\Uni-Modal Generation,Cross-Modal Generation} & Image,Audio,Text & \makecell{GPT Instructionalization,\\T2X generation} & 208K \\ 
        
        CAT's SFT~\citep{cat} & \makecell{VGGSound, AVQA,\\VideoInstruct100K...}  & Cross-Modal Understanding & Video,Audio,Text & \makecell{GPT Instructionalization} & 100K \\ 
        
        AVicuna's SFT~\citep{avicuna} & \makecell{UnAV-100, VideoInstruct100K,\\ActivityNet Captions, DiDeMo} & Uni-Modal Understanding,Cross-Modal Understanding & Video,Audio,Text & \makecell{Template Instructionalization} & 49K \\

        M3DBench'SFT~\citep{Omni-3D} & \makecell{Scannet, ScanRefer,\\ShapeNet...} & Uni-Modal Understanding,Cross-Modal Understanding & Image,3D,Text & \makecell{Template Instructionalization,//GPT Instructionalization} & 320k \\  
        
        Uni-Moe's SFT~\citep{unimoe} & \makecell{LLaVA-Instruct-150K, LibriSpeech,\\VideoInstruct100K...} & Uni-Modal Understanding,Cross-Modal Understanding & Image,Video,Audio,Text & \makecell{Template Instructionalization,\\T2X generation} & 874K \\ 
        
        X-VILA's SFT~\citep{xvila} & \makecell{WebVid, ActivityNetCaption,\\LLaVA-Instruct-150K...} & \makecell{Uni-Modal Understanding,Cross-Modal Understanding,\\Uni-Modal Generation,Cross-Modal Generation} & Image,Video,Audio,Text & \makecell{Template Instructionalization} & -- \\ 
    
        EMOVA's SFT~\citep{emova} & \makecell{ShareGPT-4o, MSCOCO,\\LLaVA-Instruct-150K...} & \makecell{Uni-Modal Understanding,Cross-Modal Understanding,\\Uni-Modal Generation,Cross-Modal Generation} & Image,Audio,Text & \makecell{Template Instructionalization,\\GPT Instructionalization, T2X generation} & 4.4M \\ 
    
        VideoLLaMA2's SFT~\citep{videollama2} & \makecell{AVQA, AVSD,\\MusicCaps...} & Uni-Modal Understanding,Cross-Modal Understanding & Image,Video,Text & \makecell{Template Instructionalization} & 1.5M \\ 
        
        PathWeave's SFT~\citep{PathWeave} & \makecell{VQAV2, MSRVTT,\\Cap3D...} & Uni-Modal Understanding & Image,Video,Audio,3D,Depth & \makecell{Template Instructionalization,\\T2X generation} & 23.2M \\ 
        
        Spider's SFT~\citep{Spider} & \makecell{AudioCap, CC3M,\\Webvid...} & Cross-Modal Understanding,Cross-Modal Generation & Image,Video,Audio,Text & \makecell{Template Instructionalization,\\GPT Generation} & -- \\ 

        GMA's SFT~\citep{GMA} & \makecell{Meta-World,CALVIN,\\Maniskill...} & \makecell{Uni-Modal Understanding,Cross-Modal Understanding,\\Cross-Modal Generation} & Image,Text,Action & \makecell{Template Instructionalization} & 2.2M \\ 
        
        OCTAVIUS's SFT~\citep{Octavius} & \makecell{MSCOCO,Bamboo,\\ScanNet...} & Uni-Modal Understanding & Image,3D,Text & \makecell{Template Instructionalization,\\GPT Generation} & ~ \\ 
        
        Lyra's SFT~\citep{Lyra} & \makecell{Mini-Gemini,\\Collected Youtube's Audio} & \makecell{Uni-Modal Understanding,Cross-Modal Understanding,\\Uni-Modal Generation} & Image,Audio,Text & \makecell{GPT Generation,\\T2X Generation} & 1.5M \\ 
        
        video-SALMONN's SFT~\citep{video-SALMONN} & \makecell{LibriSpeech,AudioCaps,\\LLaVA-Instruct-150K...} & Uni-Modal Understanding,Cross-Modal Understanding & Video,Audio,Text & \makecell{Template Instructionalization,\\T2X Generation} & -- \\ 
        
        Meerkat's SFT~\citep{meerkat} & \makecell{VGG-SS,AVSBench,\\AVQA,MUSIC-AVQA...} & Cross-Modal Understanding & Video,Audio,Text & \makecell{Template Instructionalization,\\GPT Generation} & 3M \\ 
        
        VITA's SFT~\citep{vita} & \makecell{ShareGPT4V,LLaVA-Instruct-150K,\\ShareGTP4o,ShareGemini...} & Uni-Modal Understanding,Cross-Modal Understanding & Image,Video,Audio,Text & \makecell{T2X Generation} & -- \\ 
        
        Baichuan-omni's SFT~\citep{baichuan-omni} & \makecell{vFLAN,VideoInstruct100K...} & Uni-Modal Understanding,Cross-Modal Understanding & Image,Video,Audio,Text & \makecell{T2X Generation} & -- \\ 
        
        LongVALE-LLM's SFT~\citep{LongVALE-LLM} & \makecell{LongVALE} & Uni-Modal Understanding,Cross-Modal Understanding & Video,Audio,Text & \makecell{GPT Generation} & 25.4K \\ 
         
        UnifiedMLLM's SFT~\citep{unifiedmllm} & \makecell{LISA,SmartEdit...} & \makecell{Uni-Modal Understanding,Cross-Modal Understanding,\\Uni-Modal Generation,Cross-Modal Generation} & Image,Video,Audio,Text & \makecell{Template Instructionalization,\\GPT Generation} & 100K \\ 
        
        Dolphin's SFT~\citep{Dolphin} & \makecell{AVQA,Flickr-SoundNet,\\VGGSound,LLP...} & Uni-Modal Understanding,Cross-Modal Understanding & Video,Audio,Text & \makecell{Template Instructionalization,\\GPT Generation} & -- \\ 
        \bottomrule
    \end{tabular}  
        
    }\caption{ \textbf{The statistics for OmniMLLM's Instruction Data,} including the data sources, interaction forms, involved modalities, and construction methods.}\label{table:sft_dataset}   
\end{table*}

\subsection{Details of Benchmark}\label{section:omnimllm_benchmark_details}
The statistical data of some commonly used benchmarks are shown in Table~\ref{table:benchmark_stastic}. Existing benchmarks still require improvements in terms of the number of modalities and the forms of modality interaction.
\begin{table*}[t]
    \centering
     \resizebox{0.7\linewidth}{!}{
    \begin{tabular}{lllll}
        
    \toprule
               Name & Capability Category & Modality & Specific-Task & Metrics \\ 
    \midrule
        
        VQA v2~\citep{VQAv2} & Unimodal Understanding & Image,Text & QA & Acc \\ 
        GQA~\citep{GQA} & Unimodal Understanding & Image,Text & QA & Acc \\ 
       
        DocVQA~\citep{DocVQA} & Unimodal Understanding & Image,Text & QA & Acc \\ 
        IconQA~\citep{IconQA}  & Unimodal Understanding & Image,Text & QA & Acc \\ 
        OCR-VQA~\citep{OCR-VQA} & Unimodal Understanding & Image,Text & QA & Acc \\ 
        STVQA~\citep{stvqa} & Unimodal Understanding & Image,Text & QA & Acc \\ 
        VSR~\citep{VSR} & Unimodal Understanding & Image,Text & QA & Acc \\ 
        Hateful Meme~\citep{HatefulMeme} & Unimodal Understanding & Image,Text & QA & AUC \\ 
        OKVQA~\citep{OKVQA} & Unimodal Understanding & Image,Text & QA & Acc \\ 
        VizWiz~\citep{vizwiz}   & Unimodal Understanding & Image,Text & QA & Acc \\ 
        TextVQA~\citep{TextVQA} & Unimodal Understanding & Image,Text & QA & Acc \\ 
        nocap~\citep{nocaps} & Unimodal Understanding & Image,Text & Caption & CIDER \\ 
        ScienceQA~\citep{ScienceQA}  & Unimodal Understanding & Image,Text & QA & Acc \\ 
        MSCOCO Caption~\citep{MSCOCO} & Unimodal Understanding & Image,Text & Caption & CIDER,BLEU \\ 
        Flickr Caption~\citep{flicker30k} & Unimodal Understanding & Image,Text & Caption & CIDER \\ 
        Visual Dialog~\citep{VisualDialog} & Unimodal Understanding & Image,Text & Dialogue & MRR \\ 
        RefCOCO~\citep{refcoco} & Unimodal Understanding & Image,Text & Grounding & Acc \\ 
        RefCOCO+~\citep{refcoco} & Unimodal Understanding & Image,Text & Grounding & Acc \\ 
        RefCOCOg~\citep{refcocog} & Unimodal Understanding & Image,Text & Grounding & Acc \\ 
        A-okvqa~\citep{A-OKVQA} & Unimodal Understanding & Image,Text & QA & Acc \\ 
        POPE~\citep{POPE} & Unimodal Understanding & Image,Text & Hallucination & Acc \\ 
        IIIT5K~\citep{IIIT5K} & Unimodal Understanding & Image,Text & OCR & WAC(word ACC) \\ 
        IC13~\citep{IC13} & Unimodal Understanding & Image,Text & OCR & WAC(word ACC) \\ 
        IC15~\citep{IC15} & Unimodal Understanding & Image,Text & OCR & WAC(word ACC) \\ 
        Total-Text~\citep{TotalText} & Unimodal Understanding & Image,Text & OCR & WAC(word ACC) \\ 
        CUTE80~\citep{CUTE80} & Unimodal Understanding & Image,Text & OCR & WAC(word ACC) \\ 
        SVT~\citep{SVT} & Unimodal Understanding & Image,Text & OCR & WAC(word ACC) \\ 
        SVTP~\citep{SVTP} & Unimodal Understanding & Image,Text & OCR & WAC(word ACC) \\ 
        COCO-Text~\citep{COCO-Text} & Unimodal Understanding & Image,Text & OCR & WAC(word ACC) \\ 
        MMB~\citep{mmb} & Unimodal Understanding & Image,Text & Comprehensive Benchmark & GPT ACC \\ 
        MME~\citep{MME}  & Unimodal Understanding & Image,Text & Comprehensive Benchmark & GPT ACC \\ 
         LLaVA-Bench~\citep{llava} & Unimodal Understanding & Image,Text & Comprehensive Benchmark & GPT ACC \\ 
        Mmmu~\citep{MMMU} & Unimodal Understanding & Image,Text & Comprehensive Benchmark & GPT ACC \\ 
        SEED~\citep{SEED} & Unimodal Understanding & Image,Text & Comprehensive Benchmark & GPT ACC \\ 
        MM-Vet~\citep{mm-vet} & Unimodal Understanding & Image,Text & Comprehensive Benchmark & GPT ACC \\ 

        ActivityNet-QA~\citep{activitynet-qa} & Unimodal Understanding & Video,Text & QA & Acc \\ 
        MSRVTT-QA~\citep{msrvtts}  & Unimodal Understanding & Video,Text & QA & Acc \\ 
        MSVD-QA~\citep{msvd}  & Unimodal Understanding & Video,Text & QA & Acc \\ 
        
        How2QA~\citep{how2qa} & Unimodal Understanding & Video,Text & QA & Acc \\ 
        NExTQA~\citep{nextqa}  & Unimodal Understanding & Video,Text & QA & ACC \\
        
        STAR~\citep{star} & Unimodal Understanding & Video,Text & QA & Acc \\ 
        MSVD-Caption~\citep{msvd}  & Unimodal Understanding & Video,Text & QA & CIDER \\ 
        VATEX ~\citep{vatex} & Unimodal Understanding & Video,Text & Caption & CIDER \\ 
        MSRVTT-Caption~\citep{msrvtts} & Unimodal Understanding & Video,Text & Caption & CIDER,BLEU \\ 
        Video-ChatGPT Benchmark~\citep{videochatgpt} & Unimodal Understanding & Video,Text & Comprehensive Benchmark & GPT ACC,GPT Score \\ 
        Kinetics-400 & Unimodal Understanding & Video,Text & Classification & Acc \\ 
        Perception test~\citep{preception-test} & Unimodal Understanding & Video,Text & Comprehensive Benchmark & GPT ACC \\ 
        EgoSchema~\citep{egoschema}  & Unimodal Understanding & Video,Text & Comprehensive Benchmark & GPT ACC \\ 
        Mvbench~\citep{mvbench} & Unimodal Understanding & Video,Text & Comprehensive Benchmark & GPT ACC \\ 
        VideoMME~\citep{videomme} & Unimodal Understanding & Video,Text & Comprehensive Benchmark & GPT ACC \\ 
        Charades-STA~\citep{charades-sta} & Unimodal Understanding & Video,Text & Grounding & IoU \\        
        AudioCaps~\citep{audiocap}  & Unimodal Understanding & Audio,Text & Caption & CIDER,SPICE,METEOR,BLEU,SPIDER \\ 
        
        ClothoAQA~\citep{clothoaqa}   & Unimodal Understanding & Audio,Text & QA & Acc \\ 
        Vocalsound~\citep{vocalsound} & Unimodal Understanding & Audio,Text & QA & Acc \\ 
        Clotho v1~\citep{clotho}  & Unimodal Understanding & Audio,Text & Caption & CIDER \\ 
        Clotho v2~\citep{clotho} & Unimodal Understanding & Audio,Text & Caption & CIDER \\ 
        ESC50~\citep{ecs50} & Unimodal Understanding & Audio,Text & Classification & Acc \\  
        LibriSpeech~\citep{librispeech}  & Unimodal Understanding & Audio,Text & ASR & WER \\ 
        AISHELL-2~\citep{aishell-2}  & Unimodal Understanding & Audio,Text & ASR & WER \\ 
        Wenetspeech~\citep{wenetspeech}  & Unimodal Understanding & Audio,Text & ASR & WER \\ 
         MusicCap~\citep{musiccap} & Unimodal Understanding & Audio,Text & Caption & CLAP Score \\ 
        TUT2017~\citep{tut2017} & Unimodal Understanding & Audio,Text & Classification & Acc \\ 
        EHSL~\citep{unimoe} & Unimodal Understanding & Audio,Text & QA & Acc \\ 
        
        Cap3D Caption~\citep{cap3d} & Unimodal Understanding & 3D,Text & Caption & CIDER \\ 
        Objaverse  Caption~\citep{objaverse} & Unimodal Understanding & 3D,Text & Caption & METEOR,ROUGE,BLEU \\ 
        Cap3D QA~\citep{cap3d} & Unimodal Understanding & 3D,Text & QA & Acc \\ 
        Objaverse  Classification~\citep{objaverse} & Unimodal Understanding & 3D,Text & Classification & GPT ACC \\ 
        Modelnet40~\citep{ModelNet40} & Unimodal Understanding & 3D,Text & Classification & Acc \\ 
        ScanRefer~\citep{ScanRefer} & Unimodal Understanding & 3D,Text & Grounding & mAP \\ 
        Nr3D~\citep{NR3D} & Unimodal Understanding & 3D,Text & Caption & BLEU,CIDER,METEOR,ROUGE-L \\ 
        SQA3D~\citep{SQA3D} & Unimodal Understanding & 3D,Text & QA & Acc \\ 
        ScanQA~\citep{scanqa} & Unimodal Understanding & 3D,Text & QA & Acc \\ 
        SUN RGB-D~\citep{sun-rgb-d} & Unimodal Understanding & Depth,Text & Classification & Acc \\ 
        NYUv2~\citep{nyuv2} & Unimodal Understanding & Depth,Text & Classification & Acc \\ 
        SUN RGB-D\_generated Nomral~\citep{onellm} & Unimodal Understanding & Normal,Text & Classification & Acc \\ 
        NYUv2\_generated Nomral~\citep{onellm} & Unimodal Understanding & Normal,Text & Classification & Acc \\ 
        ThermalQA~\citep{CREMA}  & Unimodal Understanding & Thermal, Text & QA & Acc \\ 
        TochQA~\citep{CREMA} & Unimodal Understanding & Touch, Text & QA & Acc \\ 
        Ego4D~\citep{ego4d}  & Unimodal Understanding & IMU,Text & Caption & CIDER,ROUGE \\ 
        NSD~\citep{nsd} & Unimodal Understanding & fMRI,Depth Map,Text & Caption & CIDER,ROUGE \\ 
        MSCOCO~\citep{MSCOCO} & Unimodal Generation & Image,Text & TX2X Edit & FID,CLIPSIM \\ 
        MSRVTT~\citep{msrvtts} & Unimodal Generation & Video,Text & T2X Generate & CLIPSIM \\ 
        AudioCaps~\citep{audiocap} & Unimodal Generation & Audio,Text & T2X Generate,TX2X Edit & FAD \\ 
        DAVIS~\citep{DAVIS}  & Unimodal Generation & Video,Text & TX2X Edit & CLIPSIM \\ 
        UCF-101~\citep{ucf-101} & Unimodal Generation & Video,Text & T2X Generate & FID,FVD,IS,CLIPSIM \\ 
        Evalcrafter~\citep{evalcrafter} & Unimodal Generation & Video,Text & T2X Generate & FVD,CLIPSIM \\ 
        VCTK~\citep{VCTK} & Unimodal Generation & Audio,Text & T2X Generate,TX2X Edit & WER,MCD \\ 
        MusicCap~\citep{musiccap} & Unimodal Generation & Audio,Text & T2X Generate & FAD \\ 
        Dreambench~\citep{dreambench} & Unimodal Generation & Image,Text & T2X Generate & CLIP-I,CLIP-T,DINO \\ 
        
        MUSIC-AVQA~\citep{musicavqa} & Crossmodal Understanding & Video,Audio,Text & QA & Acc \\ 
        AVSD~\citep{AVSD} & Crossmodal Understanding & Video,Audio,Text & Dialogue & CIDER,BLEU \\ 
        RACE-Audio~\citep{unimoe} & Crossmodal Understanding & Image,Audio,Text & Comprehensive Benchmark & Acc \\ 
        VALOR Caption~\citep{VALOR} & Crossmodal Understanding & Video,Audio,Text & Caption & CIDER,BLEU \\ 
        MMBench-Audio~\citep{unimoe} & Crossmodal Understanding & Image,Audio,Text & Comprehensive Benchmark & Acc \\ 
        AVQA~\citep{AVQA} & Crossmodal Understanding & Video,Audio,Text & QA & Acc \\ 
        MCUB~\citep{ModelComposition} & Crossmodal Understanding & Image,Video,Audio,3D,Text & Comprehensive Benchmark & Acc \\ 
        DisCRn~\citep{xinstruclblip} & Crossmodal Understanding & Image,Video,Audio,3D & Comprehensive Benchmark & Acc \\ 
        OmniXR~\citep{omnixr} & Crossmodal Understanding & Image,Video,Audio,Text & Comprehensive Benchmark & Acc \\ 
        Curse~\citep{curse} & Crossmodal Understanding & Image,Video,Audio,Text & Hallucination & Acc \\ 
        ISQA~\citep{favor} & Crossmodal Understanding & Image,Audio,Text & QA & Acc \\ 
        VGGSound~\citep{vggsound} & Crossmodal Understanding & Video,Audio,Text & QA & Acc \\ 
        VATEX~\citep{vatex}  & Crossmodal Understanding & Video,Audio,Text & Caption & CIDER \\ 
        UnAV-100~\citep{unav-100}  & Crossmodal Understanding & Video,Audio,Text & Ground & IoU \\ 
        LLP~\citep{LLP}  & Crossmodal Understanding & Video,Audio,Text & Ground & IoU \\ 
        Presentation-QA~\citep{video-SALMONN}  & Crossmodal Understanding & Video,Audio,Text & QA & ACC \\
        LongVALE Caption & Crossmodal Understanding & Video,Audio,Text & Caption & CIDER \\ 
        TVL Benchmark~\citep{TVL} & Cross-modal Understanding & Touch,Image,Text & QA & ACC \\ 
        AVEB~\citep{favor} & Crossmodal Understanding,Unimodal Understanding & Image,Video,Audio,Text & Comprehensive Benchmark & ACC,METEOR,SPIDER,WER \\ 
        XtoX Benchmark~\citep{xvila} & Crossmodal Understanding,Crossmodal Generation & Image,Video,Audio,Text & Comprehensive Benchmark & X-to-X Alignment Score \\ 
        \bottomrule
    \end{tabular}
     
     }\caption{ \textbf{An overview of benchmarks and tasks of Omni-MLLMs}, including the abilities being evaluated, the involved modalities, specific tasks, and evaluation metrics.}\label{table:benchmark_stastic}
\end{table*}


\subsection{Performance of Omni-MLLMs}
We statistic the performance of various mainstream Omni-MLLMs in uni-modal understanding, cross-modal understanding, and cross-modal, as shown in Table~\ref{table:performance}. We also show the performance of several Specific-MLLMs~\citep{vila,mvbench,qwenaudio,pointllm,emu,Video-lavit} on selected tasks for comparison. The results are mainly from corresponding papers (some results are used as baselines in other papers). It is worth noting that due to differences in the size and performance of the pre-trained models, Omni-MLLMs with the same backbone LLM may still not be fairly comparable. Therefore, this table only provides a rough trend of performance.

It can be seen from the table that most Omni-MLLMs still exhibit a significant performance gap in uni-modal understanding tasks compared to Specific-MLLMs. Meanwhile, in uni-modal generation tasks, models like AnyGPT and CoDi-2 have achieved performance close to or even surpassing Specific-MLLMs. Additionally, Omni-MLLMs are capable of performing cross-modal tasks that Specific-MLLMs cannot handle.


\begin{table*}[t]
        \centering
    \resizebox{1.0\linewidth}{!}{\begin{tabular}{l|l| l lllllll|lll|llll}
    \toprule
       \multirow{2}{*}{Model} & \multirow{2}{*}{LLM}   & \multicolumn{8}{c|}{Uni-Modal Understanding}  & \multicolumn{3}{c|}{Uni-Modal Generation} & \multicolumn{4}{c}{Cross-Modal Understanding} \\
 &  &  MSVD-QA 
 & MSRVTT-QA
 & VQA$^\text{v2 test}$  & Flickr & MMB$^\text{en}$ & AudioCaps$^\text{cap test}$  & ClothoAQA  & Objaverse  & COCO$^\text{gen}$  &   AudioCaps$^\text{gen}$  
 & MSRVTT$^\text{gen}$ 
 & VGGSS & AVSD 
 & MUSIC-AVQA & AVQA \\ 
 \midrule
 \multicolumn{16}{c}{\Large{Omni-MLLMs}} \\
 \midrule
        eP-ALM  & OPT-2.7B & 38.4 & 38.51 & 54.47 & -- & -- & 61.86 & -- & -- & -- & -- & -- & -- & -- & -- & -- \\ 
        ChatBridge 13B & Vicuna-13B  & 45.3 & -- & -- & 82.5 & -- & -- & -- & -- & -- & -- & -- & -- &  & 43 & -- \\ 
        PandaGPT & Vicuna-13B & 46.7 & 23.7 & -- & -- & -- & -- & -- & -- & -- & -- & -- & 32.7 & 26.1 & 33.7 & 79.8 \\ 
        Video-LLaMA & Vicuna-7B & 51.6 & 29.6 & -- & -- & -- & -- & -- & -- & -- & -- & -- & 40.8 & 36.7 & 36.6 & 81 \\ 
        Macaw-LLM & LLAMA-7B & 42.1 & 25.5 & -- & -- & 3.84 & 33.3 & -- & -- & -- & -- & -- & 36.1 & 34.3 & 31.8 & 78.7 \\ 
        ImageBind-LLM & LLaMA-7B & -- & -- & -- & 23.49 & -- & -- & 10.3 & 31 & -- & -- & -- & -- & -- & 39.72 & 54.26 \\ 
        NExT-GPT & Vicuna-7B & 64.5 & 58.4 & 66.7 & 84.5 & 58 & 81.3 & -- & -- & 10.07 & 8.67 & 31.97 & -- & -- & -- & -- \\ 
        AnyMAL 13B & LLaMA2-13B & -- & -- & 59.6 & -- & -- & -- & -- & -- & -- & -- & -- & -- & -- & -- & -- \\ 
        AnyMAL 70B & LLaMA2-70B & -- & -- & 64.2 & 95.9 & -- & 77.8 & -- & -- & -- & -- & -- & -- & -- & -- & -- \\ 
        X-InstructBLIP 7B & Vicuna-7B & 51.7 & 41.3 & 30.61 & 82.1 & 8.96 & 67.9 & 15.4 & 50 & -- & -- & -- & -- & -- & 28.1 & -- \\ 
        X-InstructBLIP 13B & Vicuna-13B & 49.2 & -- & -- & 74.7 & -- & 53.7 & 21.7 & -- & -- & -- & -- & 20.3 & 52.1 & 44.5 & 44.23 \\ 
        OneLLM 7B & LLaMA2-7B & 56.5 & -- & 71.6 & 78.6 & 60 & -- & 57.9 & 44.5 & -- & -- & -- & -- & -- & 47.6 & -- \\ 
        AV-LLM & Vicuna-7B & 67.3 & 53.7 & -- & -- & -- & 35.5 & -- & -- & -- & -- & -- & 47.6 &52.6 & 45.2 & -- \\ 
        UIO-2xxl 6.8B & -- & 52.2 & 41.5 & 79.4 & -- & 71.5 & 48.9 & -- & -- & 13.39 & 5.89 & -- & -- & -- & -- & -- \\ 
        ModaVerse & Vicuna-7b & -- & 56.5 & -- & -- & -- & 79.2 & -- & -- & 11.24 & 8.22 & 30.14 & -- & -- & -- & -- \\ 
        CREMA 7B & Mistral-7B & -- & -- & -- & -- & -- & -- & -- & -- & -- & -- & -- & -- & -- & 52.6 & -- \\ 
        GroundingGPT & Vicuna-7B & 67.8 & 51.6 & 78.7 & -- & 63.8 & -- & -- & -- & -- & -- & -- & -- & -- & -- & -- \\ 
        NaiveMC & Vicuna-7B & -- & -- & -- & -- & -- & -- & -- & 55 & -- & -- & -- & -- & -- & 53.63 & 80.7 \\ 
        DAMC & Vicuna-7B & -- & -- & -- & -- & -- & -- & -- & 60.5 & -- & -- & -- & -- & -- & 57.32 & 81.31 \\ 
        AnyGPT & LLaMA2-7B & -- & -- & -- & -- & -- & -- & -- & -- & -- & -- & -- & -- & -- & -- & -- \\ 
        CAT & LLaMA2-7B & -- & 62.7 & -- & -- & -- & -- & -- & -- & -- & -- & -- & -- & 48.6 & 92 \\ 
        AVicuna & Vicuna-7B & 70.2 & 59.7 & -- & -- & -- & -- & -- & -- & -- & -- & -- & -- & 53.1 & 49.6 & -- \\ 
        Uni-MoE & LLaMA-7B & 55.6 & -- & 66.2 & -- & 69.82 & -- & 32.6 & -- & -- & -- & -- & -- & -- & -- & -- \\ 
        X-VILA 7B & Vicuna-7B  & -- & -- & 72.9 & -- & -- & -- & -- & -- & -- & -- & -- & -- & -- & -- & -- \\ 
        VideoLLaMA2-7B & Mistral-7B & 71.7 & -- & -- & -- & -- & -- & -- & -- & -- & -- & -- & 71.4 & 57.2 & 80.9 & -- \\ 
        Meerkat & Llama-2-7B-Chat & -- & -- & -- & -- & -- & -- & -- & -- & -- & -- & -- & -- & -- & -- & 87.14 \\ 
        InternOmni & InternLM-2-Chat-7B & -- & -- & -- & -- & 81.7 & -- & -- & -- & -- & -- & -- & -- & -- & -- & -- \\ 
        UnifiedMLLM & Vicuna-7B & -- & -- & -- & -- & -- & -- & -- & -- & -- & -- & -- & -- & -- & -- & -- \\ 
        VITA & Mixtral-8x7B & -- & -- & -- & -- & 71.8 & -- & -- & -- & -- & -- & -- & -- & -- & -- & -- \\ 
        EMOVA & LLaMA-3.1-8B & -- & -- & -- & -- & 82.8 & -- & -- & -- & -- & -- & -- & -- & -- & -- & -- \\ 
        BaiChuan-omni-7B &  -- & 72.2 & -- & -- & -- & 76.2 & -- & -- & -- & -- & -- & -- & -- & -- & -- & -- \\ 
        OMCAT & Vicuna-7B & -- & -- & -- & -- & -- & -- & -- & -- & -- & -- & -- & -- & 49.4 & 73.8 & 90.2 \\ 
        PathWeave-7B & Vicuna-7B & 47.8 & 37.4 & -- & -- & -- & 64 & 33.5 & -- & -- & -- & -- & -- & -- & -- & --  \\ 
        Spider & Llama-2-7B & -- & -- & -- & -- & -- & 81.7 & -- & -- & 11.23  & 8.18 & 30.97 & -- & -- & -- & --  \\ 
         \midrule
 \multicolumn{16}{c}{\Large{Specifc-MLLMs}} \\
 \midrule
VILA-7B & LLaMA-2-7B & -- & -- & 79.9 & 74.7 & 68.9 & -- & -- & -- & -- & -- & -- & -- & -- & -- & --  \\ 
VideoChat2 & Vicuna-7B & 70 & 54.1 & -- & -- & -- & -- & -- & -- & -- & -- & -- & -- & -- & -- & --  \\ 
Qwen-Audio & Qwen-7B & -- & -- & -- & -- & -- & -- & 57.9 & -- & -- & -- & -- & -- & -- & -- & --  \\ 
PointLLM & Vicuna-7B & -- & -- & -- & -- & -- & -- & -- & 47.5 & -- & -- & -- & -- & -- & -- & --  \\ 
Emu-13B & -- & -- & -- & 52 & -- & -- & -- & -- & -- & 11.66 & -- & -- & -- & -- & -- & --  \\ 
Video-LaVIT & Llama2-7B & 73.2 & -- & 80.3 & -- & 67.3 & -- & -- & -- & -- & -- & 30.12 & -- & -- & -- & -- \\

        \bottomrule
    \end{tabular}}\caption{ \textbf{The performance of Omni-MLLMs on different benchmarks. }The selected uni-modal understanding benchmarks include Video-Text2Text~\citep{msvd,msrvtts}, Image-Text2Text~\citep{VQAv2,flicker30k,mmb}, Audio-Text2Text~\citep{audiocap,clothoaqa}, and 3D-Text2Text~\citep{objaverse}. The chosen uni-modal generation benchmarks include Text2Image~\citep{MSCOCO}, Text2Video~\citep{msrvtts}, and Text2Audio~\citep{audiocap}. The selected cross-modal understanding benchmarks are Image-Audio-Text2Text~\citep{vggsound,AVSD} and Video-Audio-Text2Text~\citep{musicavqa,AVQA}. }\label{table:performance}   
\end{table*}

\end{document}